\documentclass[lettersize,journal]{IEEEtran}
\usepackage{amsmath,amsfonts}
\usepackage{algorithmic}
\usepackage{algorithm}
\usepackage{array}
\usepackage[caption=false,font=normalsize,labelfont=sf,textfont=sf]{subfig}
\usepackage{textcomp}
\usepackage{stfloats}
\usepackage{url}
\usepackage{verbatim}
\usepackage{graphicx}
\usepackage{cite}
\usepackage{bm}

\usepackage{multirow}

\usepackage{amsmath}
\usepackage{amssymb}
\usepackage{amsthm}

\usepackage{subfig}

\usepackage{color}
\usepackage{bbding} 

\usepackage[table]{xcolor} 

\newtheorem{thm}{Theorem}

\theoremstyle{definition}
\newtheorem{defn}{Definition}

\usepackage{lineno,hyperref}

\begin{document}

\title{Homophily Heterogeneity Matters in Graph Federated Learning: A Spectrum Sharing and Complementing Perspective}

\author{Wentao Yu,~\IEEEmembership{Student Member,~IEEE}
}

\markboth{Journal of \LaTeX\ Class Files,~Vol.~14, No.~8, August~2024}%
{Shell \MakeLowercase{\textit{et al.}}: A Sample Article Using IEEEtran.cls for IEEE Journals}


\maketitle

\begin{abstract}
Since heterogeneity presents a fundamental challenge in graph federated learning, many existing methods are proposed to deal with node feature heterogeneity and structure heterogeneity. However, they overlook the critical homophily heterogeneity, which refers to the substantial variation in homophily levels across graph data from different clients. The homophily level represents the proportion of edges connecting nodes that belong to the same class. Due to adapting to their local homophily, local models capture inconsistent spectral properties across different clients, significantly reducing the effectiveness of collaboration. Specifically, local models trained on graphs with high homophily tend to capture low-frequency information, whereas local models trained on graphs with low homophily tend to capture high-frequency information. To effectively deal with homophily heterophily, we introduce the spectral Graph Neural Network (GNN) and propose a novel \underline{Fed}erated learning method by mining \underline{G}raph \underline{S}pectral \underline{P}roperties (FedGSP). On one hand, our proposed FedGSP enables clients to share generic spectral properties (\textit{i.e.}, low-frequency information), allowing all clients to benefit through collaboration. On the other hand, inspired by our theoretical findings, our proposed FedGSP allows clients to complement non-generic spectral properties by acquiring the spectral properties they lack (\textit{i.e.}, high-frequency information), thereby obtaining additional information gain. Extensive experiments conducted on six homophilic and five heterophilic graph datasets, across both non-overlapping and overlapping settings, validate the superiority of our method over eleven state-of-the-art methods. Notably, our FedGSP outperforms the second-best method by an average margin of 3.28\% on all heterophilic datasets.
\end{abstract}

\begin{IEEEkeywords}
Graph federated learning, Personalized federated learning, Spectral graph neural network.
\end{IEEEkeywords}

\section{Introduction}
Graph is a type of fundamental data structure across various domains, including social networks, transportation systems, and molecular chemistry~\cite{10505798, 10496248, yu2023atom, 10032180, bai2021two}. In real-world applications, large-scale graphs are often partitioned into lots of subgraphs and distributed across multiple clients. However, these distributed subgraphs cannot be combined together to train a centralized model, which is restricted by privacy regulations and data protection protocols. Graph Federated Learning (GFL)~\cite{baek2023personalized, ijcai2023426, zhu2024fedtad, wentao2025fediih} has emerged as a promising approach for collaboratively training models on distributed subgraphs while preserving data privacy.

However, the distributions of subgraphs on different clients are inconsistent, leading to heterogeneity and significantly reducing the effectiveness of federated collaboration~\cite{huang2024federated, 10295990, 10697408}. To deal with the heterogeneity issue, lots of GFL methods have been proposed. For example, FedGTA~\cite{li2023fedgta} employs a topology-aware optimization method to solve the structure heterogeneity in GFL. Meanwhile, FED-PUB~\cite{baek2023personalized} solves the node feature heterogeneity. Specifically, FED-PUB utilizes the node embeddings of local models to compute the similarities among clients and then performs the weighted averaging of local model parameters. In contrast, FGSSL~\cite{ijcai2023426} and FedTAD~\cite{zhu2024fedtad} deal with the node feature heterogeneity and structure heterogeneity, simultaneously.

\begin{figure}[!t]
    \centering
    \subfloat[\footnotesize{Subgraphs partitioned from \textit{CiteSeer} dataset}]{\includegraphics[width=0.92\columnwidth]{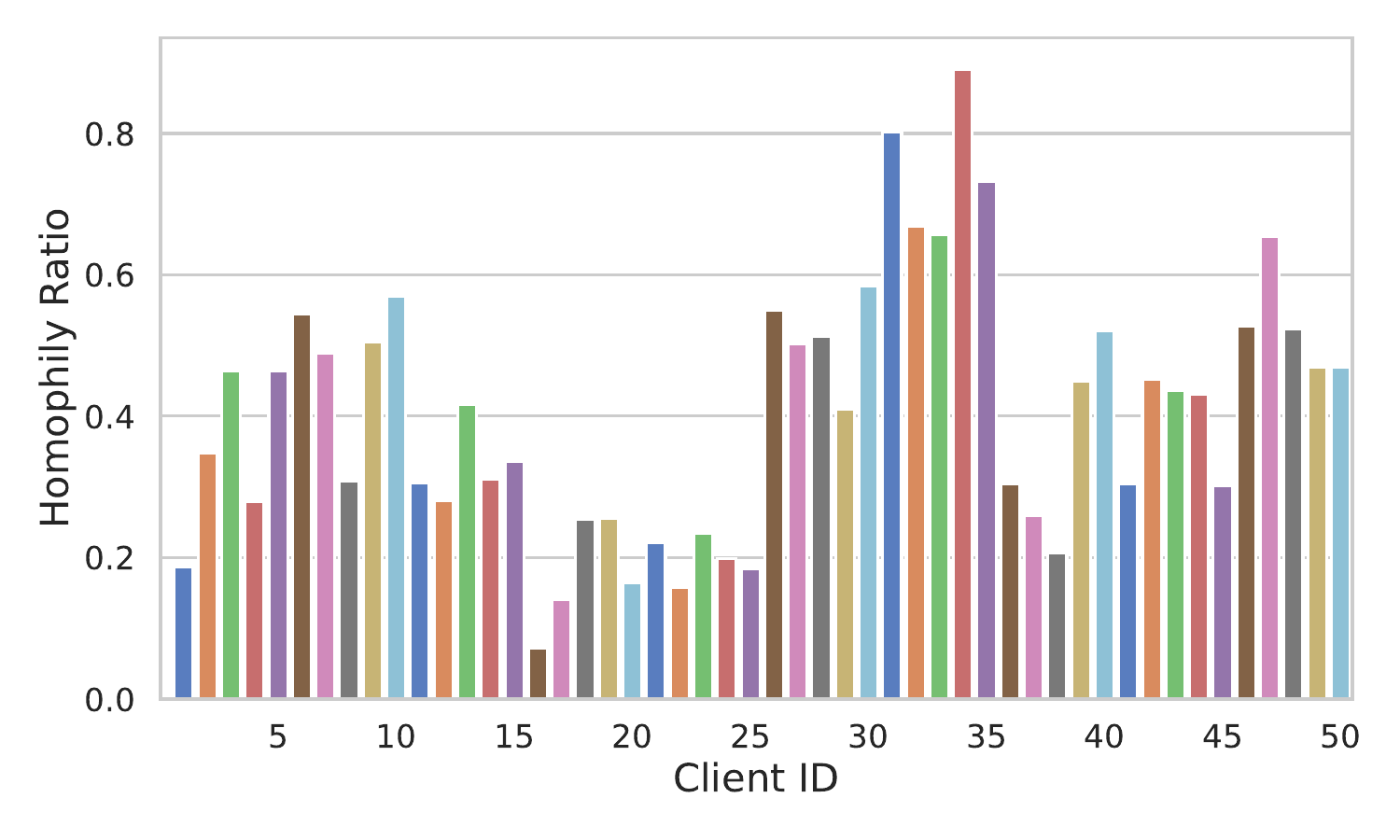}\label{fig1_1}}
    \hfill
    \subfloat[\footnotesize{Subgraphs partitioned from \textit{Questions} dataset}]{\includegraphics[width=0.92\columnwidth]{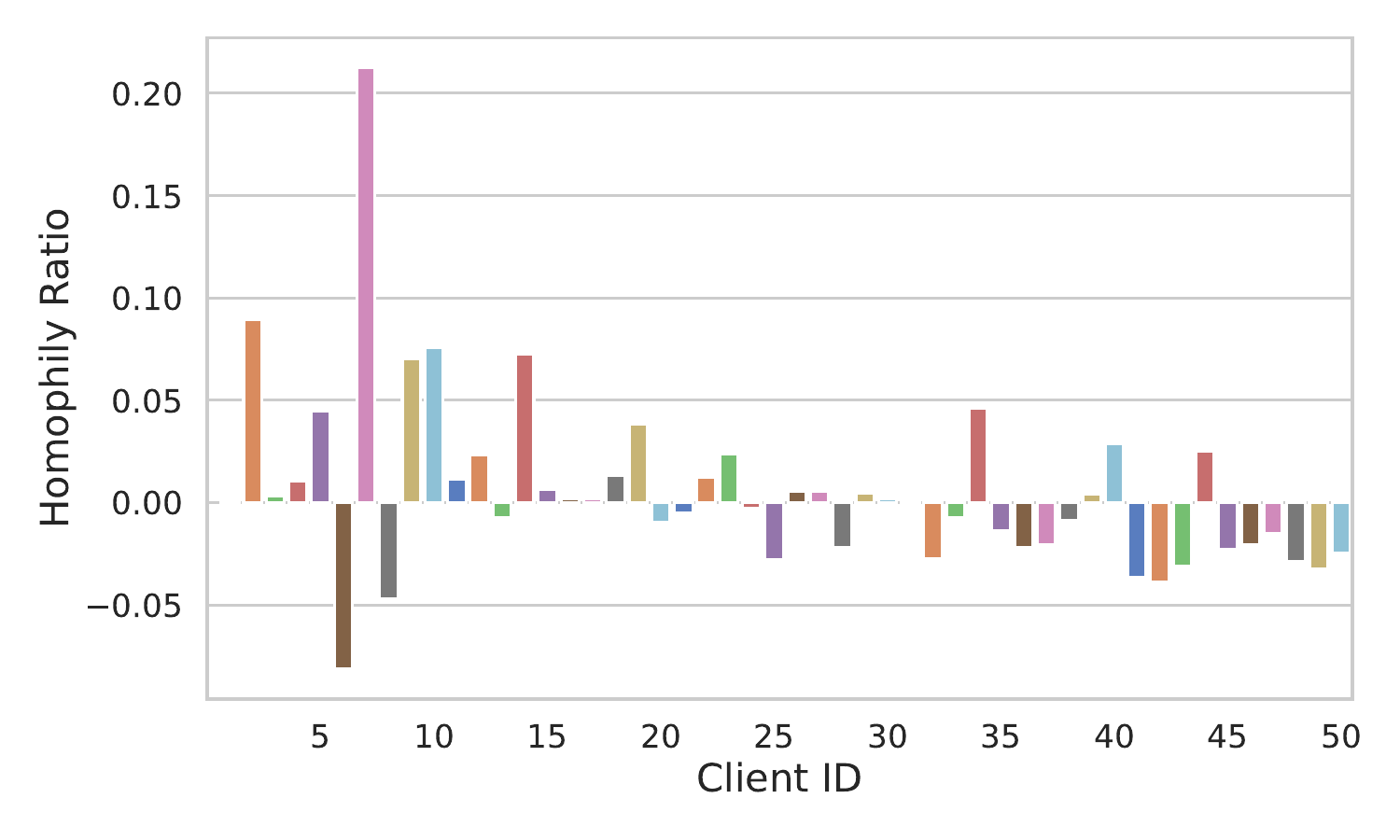}\label{fig1_2}}
    \caption{Homophily levels across clients vary significantly in subgraphs from the \textit{CiteSeer} and \textit{Questions} datasets.}  
    \label{fig1}
\end{figure}

\begin{figure}[!t]
  \centering
  \subfloat[\footnotesize{Client 34 of \textit{CiteSeer} dataset\\(homophily level=0.891)}]{\includegraphics[width=0.5\columnwidth]{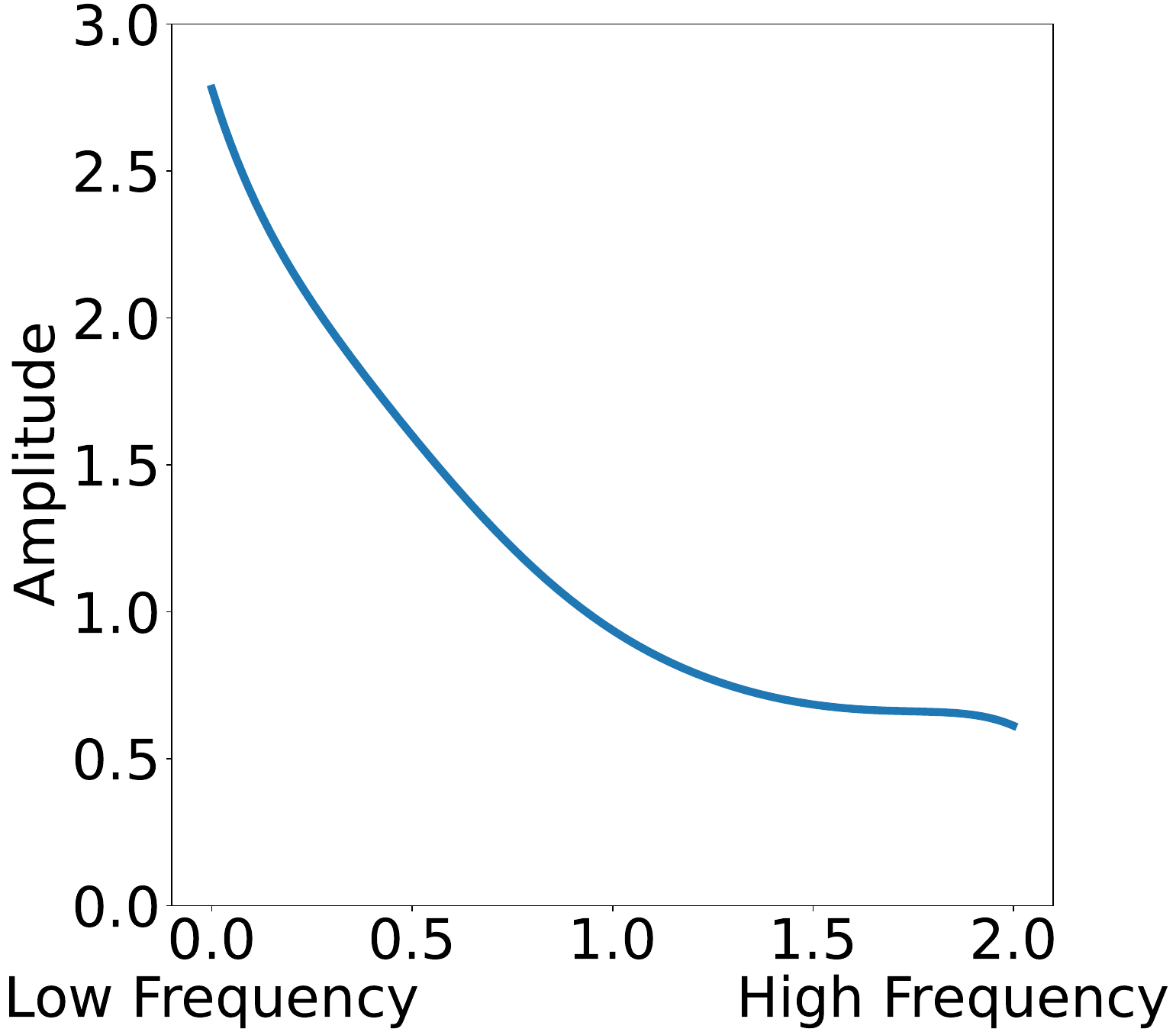}\label{fig2_1}}
  \hfill
  \subfloat[\footnotesize{Client 16 of \textit{CiteSeer} dataset\\(homophily level=0.071)}]{\includegraphics[width=0.5\columnwidth]{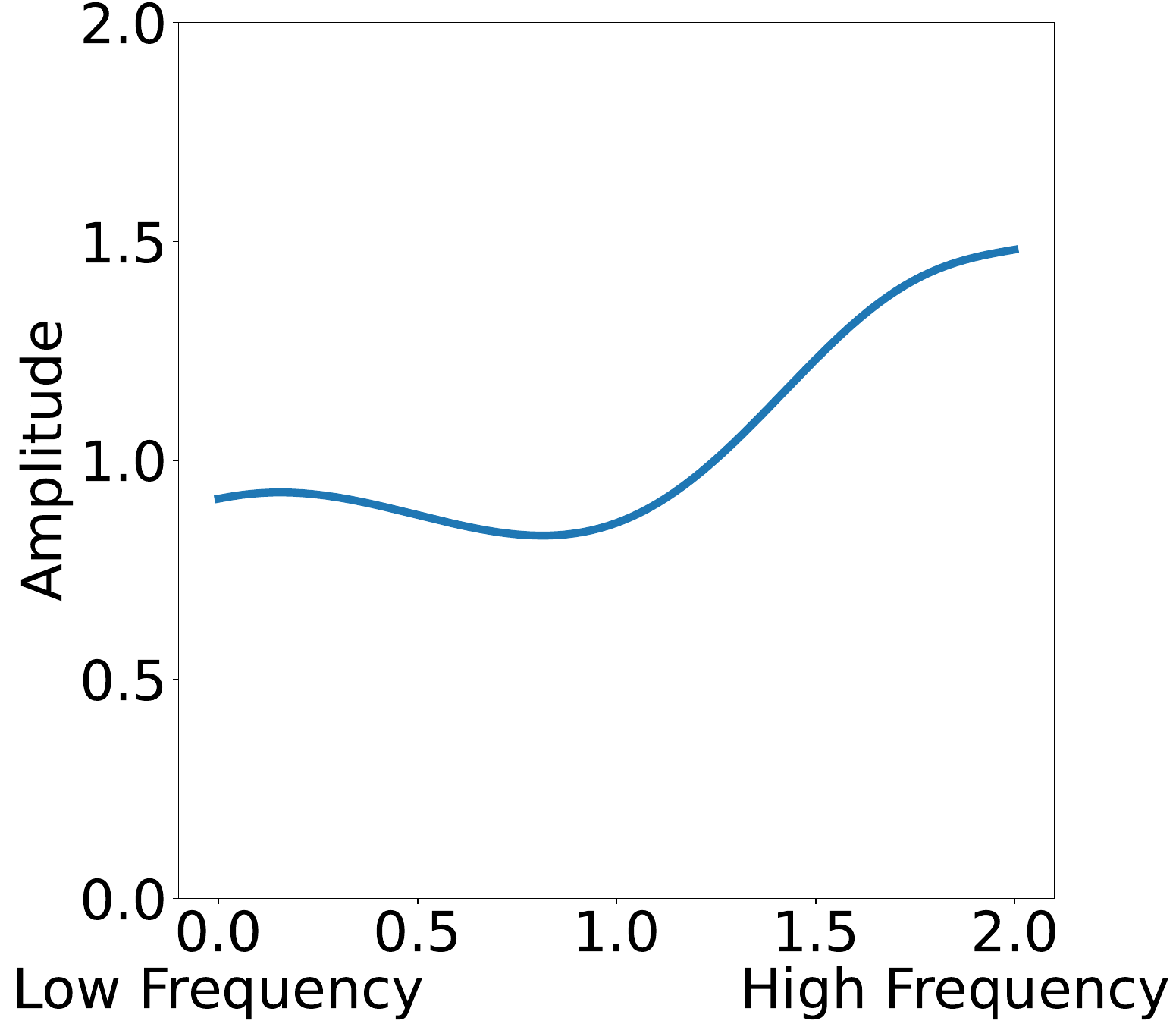}\label{fig2_2}}
  \hfill
  \subfloat[\footnotesize{Client 7 of \textit{Questions} dataset\\(homophily level=0.212)}]{\includegraphics[width=0.5\columnwidth]{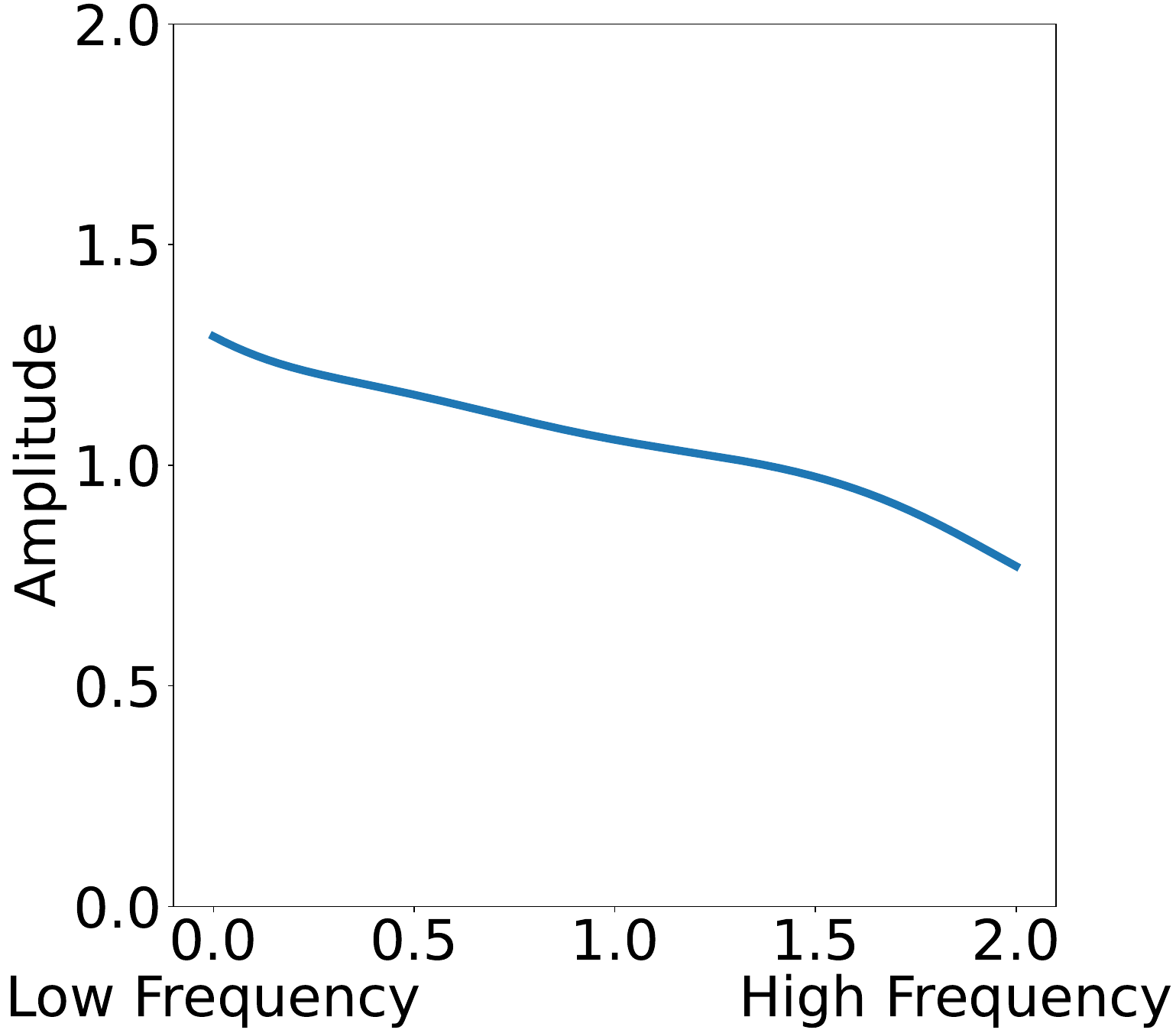}\label{fig2_3}}
  \hfill
  \subfloat[\footnotesize{Client 6 of \textit{Questions} dataset\\(homophily level=-0.081)}]{\includegraphics[width=0.5\columnwidth]{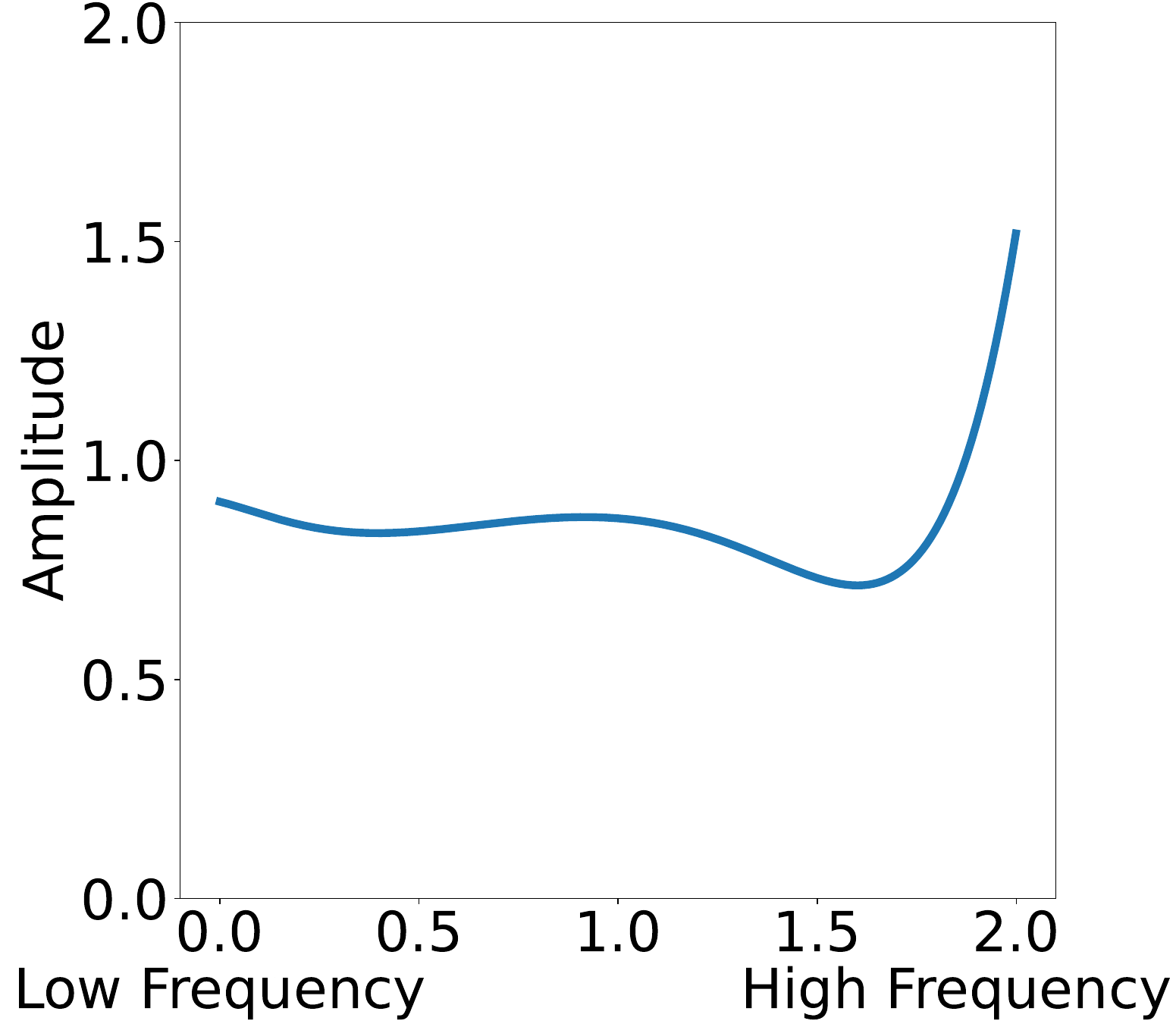}\label{fig2_4}}
  \caption{Spectral properties captured by local models vary across clients: (a) and (c) emphasize low-frequency information; (b) and (d) highlight high-frequency information.}
  \label{fig2}
\end{figure}

Despite the above-mentioned methods exploring the node feature heterogeneity and structure heterogeneity, they fail to consider the critical homophily heterogeneity. The homophily heterogeneity refers to the significant differences in homophily levels exhibited by graph data across different clients. Homophily refers to the propoperty that edges tend to connect similar nodes~\cite{NEURIPS2023_01b68102}. For example, people in social networks tend to connect to others with similar hobbies. To measure the level of homophily, we employ the \textit{adjusted homophily}~\cite{NEURIPS2023_01b68102}, which solves the issues that existing measures (\textit{e.g.}, edge homophily~\cite{zhu2020beyond}) are sensitive to the number of classes and their balance (see Appendix~VI). Generally, the higher the homophily level, the stronger the homophily. As shown in Fig.~\ref{fig1_1}, client 34 has graph data with the highest homophily level (\textit{i.e.}, 0.891), while client 16 has graph data with the lowest homophily level (\textit{i.e.}, 0.071). Similarly, as shown in Fig.~\ref{fig1_2}, client 7 has graph data with the highest homophily level (\textit{i.e.}, 0.212), while client 6 has graph data with the lowest homophily level (\textit{i.e.}, -0.081). Since local models on clients adapt to their corresponding homophily levels~\cite{balcilar2021analyzing, he2021bernnet, wang2022powerful, guo2023graph}, local models capture inconsistent spectral properties across different clients. Specifically, local models trained on graphs with high homophily tend to capture low-frequency information (see Fig.~\ref{fig2_1} and Fig.~\ref{fig2_3}), whereas local models trained on graphs with low homophily tend to capture high-frequency information (see Fig.~\ref{fig2_2} and Fig.~\ref{fig2_4}). Inconsistent spectral properties captured by different clients pose the challenge to federated aggregation, leading to sub-optimal collaborative effectiveness. Therefore, we raise the following research question:
\begin{center}
\textit{Since homophily heterogeneity matters in graph federated learning, how can we deal with it?}
\end{center}

To answer this question and deal with the homophily heterogeneity issue, we aim to introduce the spectral Graph Neural Network (GNN) as the local model and further theoretically analyze the heterogeneity in GFL. Since spectral GNNs are made up of polynomial bases, they can be easily split into low-frequency and high-frequency parts for subsequent analysis. Moreover, we propose a novel \underline{\textbf{Fed}}erated learning method by mining \underline{\textbf{G}}raph \underline{\textbf{S}}pectral \underline{\textbf{P}}roperties (FedGSP). Specifically, on one hand, since low-frequency information contains generic spectral properties~\cite{tan2024fedssp, zhou2024traffic}, we enable clients to share generic spectral properties (\textit{i.e.}, low-frequency information), thus allowing all clients to benefit through collaboration. Meanwhile, we theoretically find that the heterogeneity in GFL is proportional to the complementarity ratio. Our theoretical findings inspires us that to deal with the heterogeneity in GFL, we should pursue not only similarities between different clients but also complementarities between different clients. Therefore, on the other hand, we allow clients to complement non-generic spectral properties by acquiring the spectral properties they lack (\textit{i.e.}, high-frequency information), thereby obtaining additional information gain~\cite{chen2024polygcl, yan2024balancing}. These strategies allow the local model on each client not only to preserve its local graph spectral properties, but also to benefit from collaborations.

Moreover, since we introduce the spectral GNN as the local model, we propose a novel federated aggregation method from the perspective of polynomial bases, which are fundamental components in the spectral GNNs. By doing so, our proposed FedGSP not only effectively alleviates the homophily heterogeneity, but also significantly improves the model performances on graphs with different homophily levels. Notably, we compare our proposed FedGSP with eleven state-of-the-art methods on six homophilic and five heterophilic graph datasets, and the experimental results validate the effectiveness of our proposed FedGSP. The key contributions of our work are as follows:
\begin{itemize}
  \item This is the \textit{first} work to deal with the homophily heterogeneity in GFL, which leads to inconsistent spectral properties across different clients. Therefore, we introduce the spectral GNN as the local model, so that we can mine graph spectral properties for GFL.
  
  \item We are the \textit{first} to prove that the heterogeneity in GFL is proportional to the complementarity ratio. Inspired by our theoretical findings, we propose a novel FedGSP method. It not only shares generic spectral properties but also complements non-generic spectral properties among different clients, so that we can successfully alleviate the homophily heterogeneity and improve the effectiveness of GFL.
 
  \item Extensive experiments on eleven datasets demonstrate the superiority of our proposed FedGSP over eleven state-of-the-art methods, where our FedGSP outperforms the second-best method by an average margin of 3.28\% on all heterophilic datasets. Moreover, case studies validate that our proposed FedGSP allows the local model on each client not only to preserve its local graph spectral properties, but also to obtain the additional spectral properties from collaborations.
 
\end{itemize}

\section{Related Work}
In this section, we review the typical work related to this paper, including GFL, PFL, and Spectral GNNs.

\subsection{Graph Federated Learning}
GFL aims to utilize the distributed learning framework to collaboratively train GNNs while maintaining the privacy and security of the local graphs. Most of the existing GFL methods can be categorized into two types: optimization-based methods and model-based methods. On one hand, optimization-based methods focus on optimizing the collaboration strengths between clients, often leveraging conventional GNNs (\textit{e.g.}, Graph Convolutional Networks (GCN)~\cite{kipf2017semisupervised}). For instance, FED-PUB~\cite{baek2023personalized} determines client collaboration strengths according to the estimated similarities between subgraphs based on the outputs of local GCN-based models. Similarly, FedGTA~\cite{li2023fedgta} introduces the optimization-driven federated learning algorithm that adjusts the weight of each client's contribution based on the mixed moments of processed neighbor features. On the other hand, model-based methods focus on designing specific local models while using simple federation strategies (\textit{e.g.}, FedAvg~\cite{mcmahan2017communication}). For example, FedSage+~\cite{NEURIPS2021_34adeb8e} builds on FedSage\cite{NEURIPS2021_34adeb8e} by introducing a missing neighbor generator to address missing links across local subgraphs, while FedSage itself combines GraphSAGE~\cite{NIPS2017_5dd9db5e} with FedAvg. To deal with the structure heterogeneity, AdaFGL~\cite{li2024adafgl} introduces homophilous and heterophilous propagation modules for each client so that each client's model is adapted to the local graph structures. Besides, FedTAD~\cite{zhu2024fedtad} enhances the knowledge transfer from the local models to the global model, alleviating the negative impact of unreliable knowledge caused by node feature heterogeneity. However, both optimization-based methods and model-based methods have performance limitations because they overlook the critical homophily heterogeneity, which leads to inconsistent spectral properties across different clients. Therefore, we naturally introduce the spectral GNN as the local model to mine graph spectral properties for GFL.

\subsection{Personalized Federated Learning}
Heterogeneity presents a fundamental challenge in GFL~\cite{ye2023heterogeneous}. To deal with the heterogeneity, PFL methods~\cite{MLSYS2020_1f5fe839, 9743558, Arivazhagan2019} have obtained increasing attention. Unlike FedAvg, which aims to train a global model collaboratively, PFL methods aim to train a personalized model for each client. Most of existing PFL methods can be categorized as similarity-based methods~\cite{baek2023personalized, li2023fedgta, wentao2025fediih}, local customization-based methods~\cite{MLSYS2020_1f5fe839, Arivazhagan2019, NEURIPS2020_f4f1f13c}, and meta-learning-based methods~\cite{chen2018federated, NEURIPS2020_24389bfe, 10485381}. Similarity-based methods first compute the inter-client similarities and then perform the weighted federation of local models for each client based on these similarities. Specifically, clients with larger similarity scores are assigned larger weights for federated aggregation. In contrast, local customization-based methods, such as FedProx~\cite{MLSYS2020_1f5fe839}, incorporate a proximal term to customize personalized models for each client. Similarly, FedALA~\cite{zhang2023fedala} captures the desired information from the global model in an element-wise manner and then aggregates the local models. As an alternative, FedPer~\cite{Arivazhagan2019} focuses on federating the backbone weights while training a personalized classification layer locally on each client. Additionally, meta-learning-based methods, such as~\cite{NEURIPS2020_24389bfe}, focus on discovering an initial shared model that can be efficiently adapted to each client, enabling the creation of personalized local models. Due to the simplicity and effectiveness of similarity-based methods, we concentrate on similarity-based PFL methods in this paper. However, similarity-based PFL methods only consider the similarities between pairwise clients, while ignoring the critical complementarities between clients. Therefore, according to our theoretical findings, we propose to pursue not only similarities between different clients but also complementarities between different clients. 

\subsection{Spectral Graph Neural Networks}
Spectral GNNs represent a class of GNNs that operate by designing graph signal filters in the spectral domain~\cite{wang2022powerful}. Specifically, they approximate filtering operations using polynomial bases of Laplacian eigenvalues~\cite{guo2023graph}. For example, ChebNet~\cite{defferrard2016convolutional} employs a $K$-order truncated Chebyshev polynomial basis to implement a $K$-hop localized filtering. However, Chien~\textit{et al.}~\cite{chien2021adaptive} argue that the depth of ChebNet is limited in practice due to the over-smoothing phenomenon. To address this issue, they propose a Generalized PageRank (GPR) GNN (\textit{a.k.a.} GPR-GNN~\cite{chien2021adaptive}) that adaptively learns the GPR weights to control the contribution of each propagation step. However, the above-mentioned spectral GNNs learn the graph signal filters without a clear constraint, which may lead to oversimplified or ill-posed filters. To overcome this problem, BernNet~\cite{he2021bernnet} estimates the normalized Laplacian spectrum by a $K$-order Bernstein polynomial basis and learns the polynomial coefficients based on the observed graphs. Meanwhile, Wang \textit{et al.}~\cite{wang2022powerful} theoretically analyze the expressive power of spectral GNNs and find the advantage of orthogonal polynomial bases. Inspired by their theoretical findings, they propose JacobiConv~\cite{wang2022powerful}, which is based on the Jacobi polynomial basis due to its orthogonality and flexibility. Furthermore, OptBasisGNN~\cite{guo2023graph} learns an orthogonal polynomial basis directly from the graph data. Nevertheless, the above-mentioned spectral GNNs can not deal with the varying homophily levels. To solve this problem, Huang \textit{et al.}~\cite{huanguniversal} propose the universal polynomial bases. Inspired by this work, we not only introduce the spectral GNN as the local model to tackle the issue of homophily heterogeneity across different clients, but also propose a novel federated aggregation method from the perspective of polynomial bases.

\section{Preliminaries}
In this section, we introduce some mathematical notations related to the setting of GFL. In this paper, we concentrate on the task of node classification under the GFL scenario. In particular, we aim at the collaborative training of node classifiers with local subgraphs on different clients. Suppose we have $M$ clients, each of which possesses a local subgraph $\mathcal{G}_m=\langle\mathcal{V}_m, \mathcal{E}_m\rangle$, where $\mathcal{V}_m$ denotes the node set, $\mathcal{E}_m$ represents the edge set, and $m = 1, \dots, M$. We utilize $\mathbf{X}_m \in \mathbb{R}^{n_m \times d}$ and $\mathbf{A}_m \in \mathbb{R}^{n_m \times n_m}$ to represent the node feature matrix and the adjacency matrix of $\mathcal{G}_m$, respectively. Here the number of nodes in $\mathcal{G}_m$ and the feature dimension are denoted as $n_m$ and $d$, respectively. Similarly, $e_m$ and $c_m$ denote the number of edges and classes, respectively. In addition, let $\mathbf{L}_m= \mathbf{D}_m - \mathbf{A}_m=\mathbf{I}-\mathbf{D}_m^{-\frac{1}{2}}\mathbf{A}_m\mathbf{D}_m^{-\frac{1}{2}}$ denote the symmetric normalized Laplacian matrix of $\mathcal{G}_m$ without self-loops, where $\mathbf{D}_m$ is the degree matrix of  $\mathcal{G}_m$, and  $\mathbf{I}$ is the identity matrix. Then, we can have the propagation matrix $\mathbf{P}_m = \mathbf{I} - \mathbf{L}_m$. Here we also introduce some norms used in this paper. First, we denote the $l_2$-norm of a vector $\mathbf{v}$ as $\Vert \mathbf{v} \Vert_2$, where $\Vert \mathbf{v} \Vert_2 = \sqrt{\sum\nolimits_{i} v_i^2}$, and $v_i$ is the $i$-th element of $\mathbf{v}$. Second, $\Vert \mathbf{X} \Vert_F$ denotes the Frobenius norm of a matrix $\mathbf{X}$, where $\Vert \mathbf{X} \Vert_F = \sqrt{\sum\nolimits_{i,j} |x_{ij}|^2}$, and $x_{ij}$ is the element of $\mathbf{X}$.

\section{Theoretical Foundations}
In this section, we theoretically analyze the heterogeneity in GFL and present our theoretical findings, which construct the theoretical foundation of our proposed method. First, we describe the construction of the federated collaboration graph, which is used to derive our theoretical findings. Second, we provide some definitions to facilitate the subsequent theorems. Third, we establish the relationship between the heterogeneity in GFL and the client's complementarity.

\subsection{Federated Collaboration Graph}
Inspired by pFedGraph~\cite{ye2023personalized}, we construct a federated collaboration graph $\mathcal{G}_\mathrm{c}=\langle\mathcal{V}_\mathrm{c}, \mathcal{E}_\mathrm{c}\rangle$ to model the collaboration strength between clients, where $\mathcal{V}_\mathrm{c}=\{\mathrm{c}_1, \mathrm{c}_2, \cdots, \mathrm{c}_M\}$ represents the node set, $\mathcal{E}_\mathrm{c}$ is the edge set, and $M$ is the number of clients. In $\mathcal{G}_\mathrm{c}$, each node represents a client. Therefore, the adjacency matrix of $\mathcal{G}_\mathrm{c}$ is denoted as $\mathbf{W}_\mathrm{c} \in \mathbb{R}^{M \times M}$, where $\mathbf{W}_\mathrm{c}^{ij}$ reflects the collaboration strength between the $i$-th and the $j$-th client. The neighbor set of node $u \in \mathcal{V}_\mathrm{c}$ is denoted as $\mathcal{N}_\mathrm{c}$. Here the degree of node $u$ is denoted as $d_u = |\mathcal{N}_\mathrm{c}|$. In addition, the degree matrix of $\mathcal{G}_\mathrm{c}$ is denoted as $\mathbf{D}_\mathrm{c} \in \mathbb{R}^{M \times M}$, where $\mathbf{D}_\mathrm{c}^{uu} = d_u$. Similarly, let $\mathbf{L}_\mathrm{c}= \mathbf{D}_\mathrm{c} - \mathbf{W}_\mathrm{c}=\mathbf{I}-\mathbf{D}_\mathrm{c}^{-\frac{1}{2}}\mathbf{W}_\mathrm{c}\mathbf{D}_\mathrm{c}^{-\frac{1}{2}}$ denote the symmetric normalized Laplacian matrix of $\mathcal{G}_\mathrm{c}$ without self-loops. We treat the model parameters on the $i$-th client (\textit{i.e.}, $\bm{\theta}_i$) as the $i$-th node's features of $\mathcal{G}_\mathrm{c}$. Therefore, the node feature matrix of $\mathcal{G}_\mathrm{c}$ can be represented as $\bm{\Theta}$, where $\bm{\Theta} = [\bm{\theta}_1, \bm{\theta}_2, \cdots, \bm{\theta}_M]^{\top}$, and $[\cdot, \cdot]$ denotes the concatenation operation. Consequently, the model aggregation of PFL (\textit{i.e.}, weighted combination of model parameters among different clients) can be treated as the message passing of node features in $\mathcal{G}_\mathrm{c}$.

\subsection{Definitions}
Here we first present some definitions to describe the similarities and complementarities between clients, respectively.

\begin{defn}
\textbf{Similar Client Pair}: Given any pair of clients $i$ and $j$, the normalized similarity of their data distributions is defined as $S(i,j) \in [0, 1]$. If $S(i,j) \geq 0.5$, clients $i$ and $j$ are considered similar.
\end{defn}
    
\begin{defn}
\textbf{Complementary Client Pair}: Given any pair of clients $i$ and $j$, if $S(i,j) < 0.5$, clients $i$ and $j$ are considered complementary.
\end{defn}
    
\begin{defn}
\label{defn_3}
\textbf{Similarity Ratio of Clients $r_\mathrm{s}$}: Given $M$ clients performing federated learning, the similarity ratio of $M$ clients $r_s$ is the fraction of similar client pairs, \textit{i.e.}, 
\begin{equation}\label{eq_r_s}
r_\mathrm{s} = \frac{|\{\langle i, j\rangle \in \mathcal{E}_\mathrm{c}: S(i,j) \geq 0.5\}|}{|\mathcal{E}_\mathrm{c}|}.
\end{equation}
\end{defn}
    
\begin{defn}
\label{defn_4}
\textbf{Complementarity Ratio of Clients $r_\mathrm{c}$}: Given $M$ clients performing federated learning, the complementarity ratio of $M$ clients $r_\mathrm{c}$ is the fraction of complementary client pairs, \textit{i.e.},
\begin{equation}\label{eq_r_c}
r_\mathrm{c} = \frac{|\{\langle i, j\rangle \in \mathcal{E}_\mathrm{c}: S(i,j) < 0.5\}|}{|\mathcal{E}_\mathrm{c}|}.
\end{equation}
Obviously, we can find that $r_\mathrm{c} = 1 - r_\mathrm{s}$. Based on the above definitions, we can find that a higher complementarity ratio means a higher level of heterophily.
\end{defn}

\begin{figure}[!t]
  \centering
  \subfloat[\footnotesize{\textit{ogbn-arxiv} non-overlapping ($r_\mathrm{c}=0.935$)}]{\includegraphics[width=0.45\columnwidth]{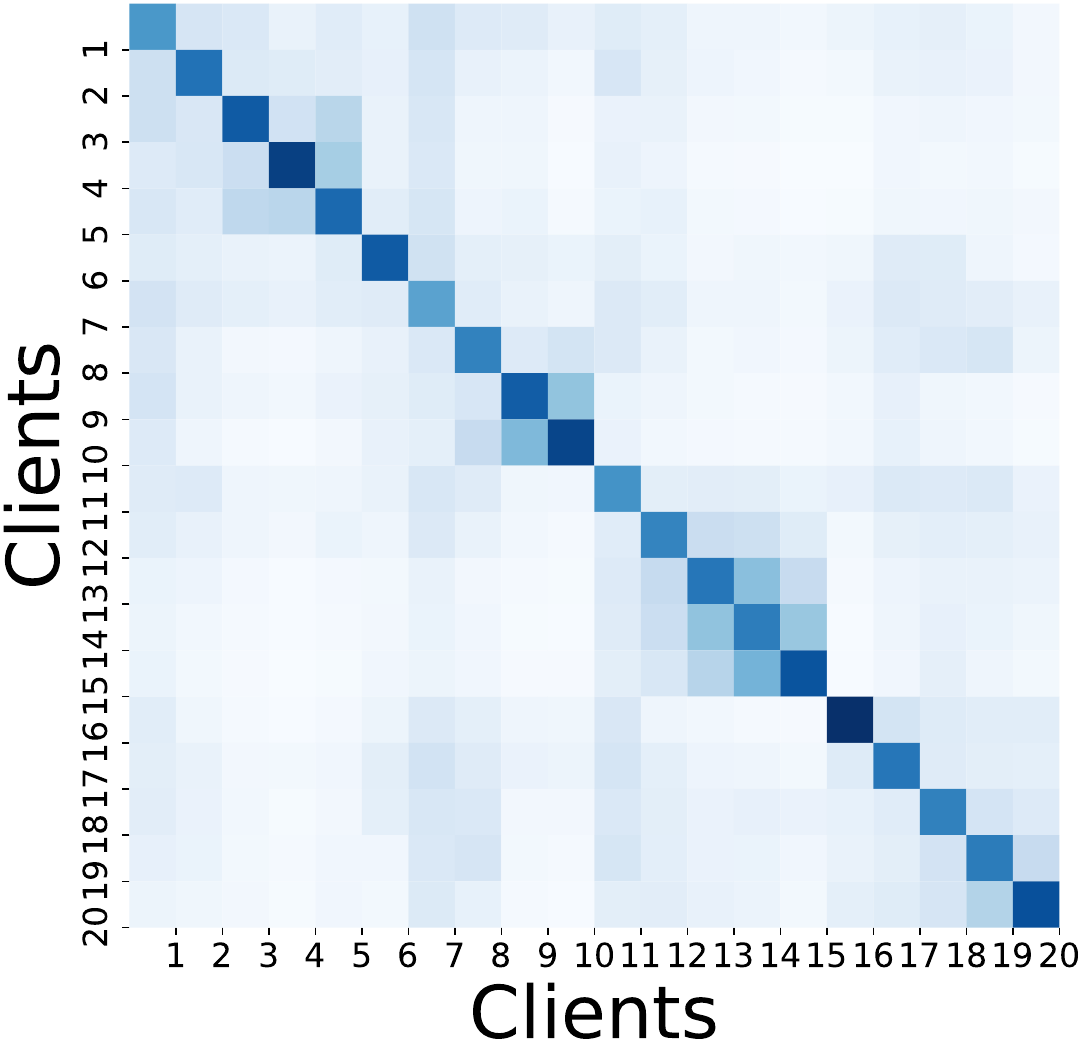}\label{fig_ex_1}}
  \hfill
  \subfloat[\footnotesize{\textit{Amazon-ratings} non-overlapping ($r_\mathrm{c}=0.95$)}]{\includegraphics[width=0.45\columnwidth]{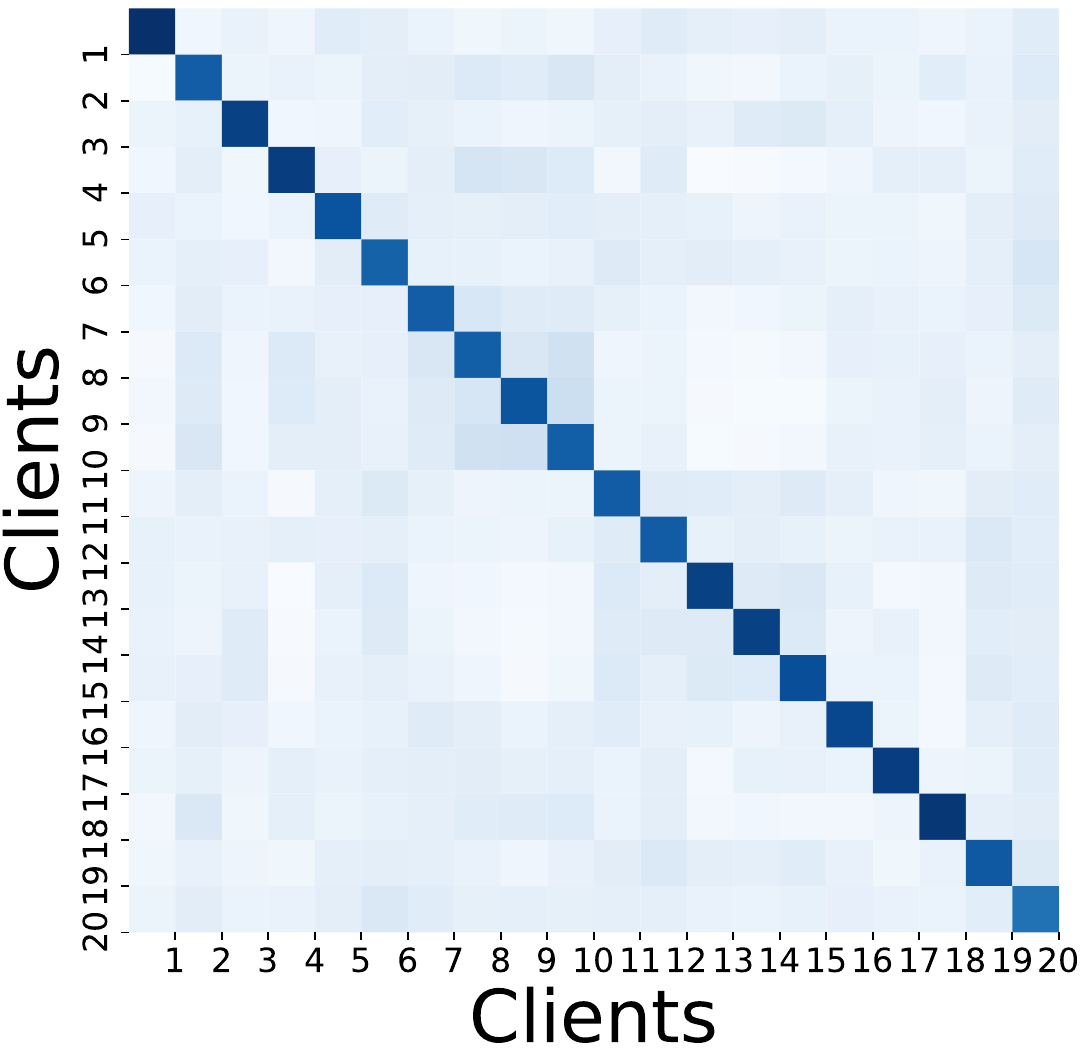}\label{fig_ex_2}}
  \hfill
  \subfloat[\footnotesize{\textit{ogbn-arxiv} overlapping ($r_\mathrm{c}=0.833$)}]{\includegraphics[width=0.45\columnwidth]{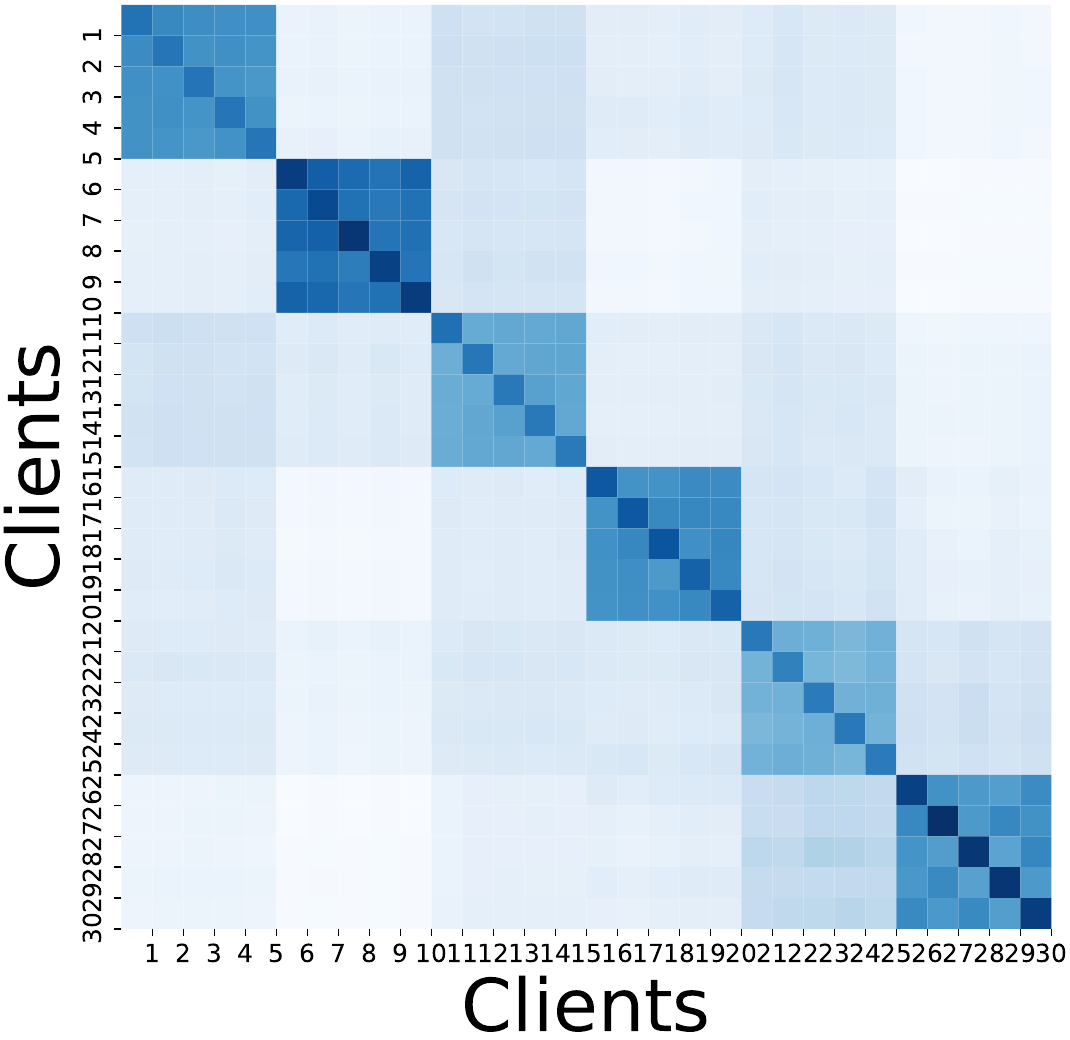}\label{fig_ex_3}}  
  \hfill
  \subfloat[\footnotesize{\textit{Amazon-ratings} overlapping ($r_\mathrm{c}=0.944$)}]{\includegraphics[width=0.45\columnwidth]{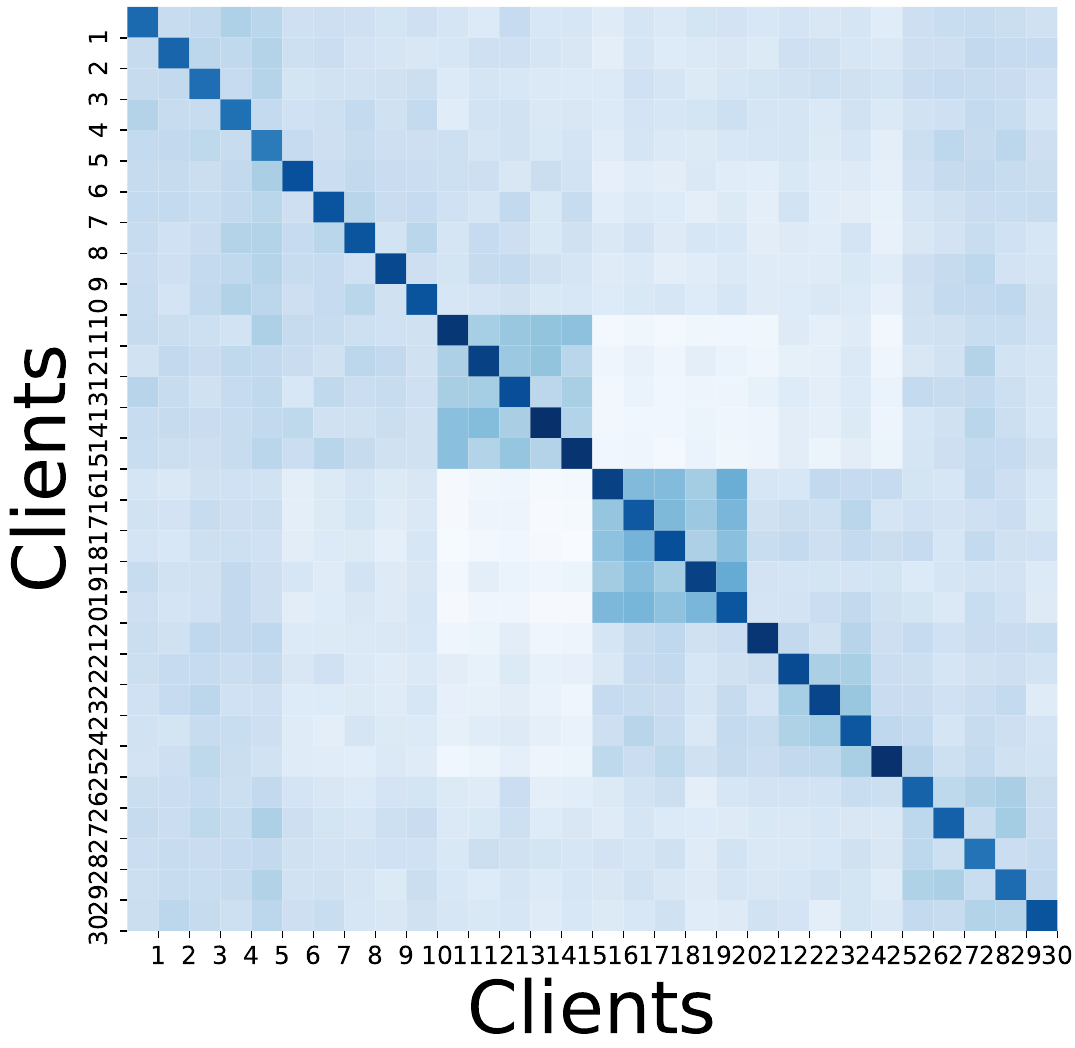}\label{fig_ex_4}}
  \caption{The normalized similarity matrices under the non-overlapping and overlapping partitioning settings.}
  \label{fig_example}
\end{figure}

To validate the practical meaning of the above definitions, we provide four examples on the \textit{ogbn-arxiv} and \textit{Amazon-ratings} datasets, respectively. As shown in Fig.~\ref{fig_example}, we plot the normalized similarity matrices under the non-overlapping and overlapping partitioning settings. According to our definition~\ref{defn_3} and definition~\ref{defn_4}, we can easily obtain their similarity and complementarity ratios, respectively. The computed complementarity ratios are consistent with the practical situations. For example, the \textit{Amazon-ratings} dataset has higher complementarity ratios than the \textit{ogbn-arxiv} dataset, which is consistent with the observation that heterophilic datasets tend to have stronger heterophily than homophilic datasets~\cite{wentao2025fediih}. Furthermore, since each five subgraphs are overlapped under the overlapping partitioning setting~\cite{baek2023personalized}, datasets under the overlapping setting have smaller complementarity ratios than under the non-overlapping setting.

Second, we define the Laplacian frequency component from the spectral perspective.
\begin{defn}
\textbf{Laplacian Frequency Component $f$}: Consider the federated collaboration graph $\mathcal{G}_\mathrm{c}=\langle\mathcal{V}_\mathrm{c}, \mathcal{E}_\mathrm{c}\rangle$ with the symmetric normalized Laplacian matrix $\mathbf{L}_\mathrm{c}$. Given the node feature matrix $\bm{\Theta}$ of $\mathcal{G}_\mathrm{c}$, the Laplacian Frequency Component on $\mathcal{G}_\mathrm{c}$ is defined as $f(\bm{\Theta}) = \mathrm{Tr}(\frac{\bm{\Theta}^{\top} \mathbf{L}_\mathrm{c} \bm{\Theta}}{2})$, where ``$\mathrm{Tr}(\cdot)$'' computes the trace of the corresponding matrix.
\end{defn}

Third, since the heterogeneity in GFL can be reflected in the Euclidean distance between any pair of client models~\cite{luo2021no, vahidian2023rethinking, 10286439}, here we define a typical measure of heterogeneity in GFL as follows:
\begin{defn}
\textbf{Measure of Heterogeneity in GFL $H(\mathcal{G}_\mathrm{c})$}: Consider the federated collaboration graph $\mathcal{G}_\mathrm{c}$ with the adjacency matrix $\mathbf{W}_\mathrm{c}$, the Measure of Heterogeneity $H(\mathcal{G}_\mathrm{c})$ is defined as $H(\mathcal{G}_\mathrm{c})=\sum_{\langle i, j\rangle \in \mathcal{E}_\mathrm{c}} \mathbf{W}_\mathrm{c}^{ij} \Vert \bm{\theta}_i - \bm{\theta}_j\Vert^2_2 $.
\end{defn}

\subsection{Relationship between Heterogeneity and Complementarity}
Due to the characteristic of Laplacian matrix~\cite{guo2023graph}, the Laplacian frequency component $f(\bm{\Theta})$ actually captures the spectral frequency of the federated collaboration graph $\mathcal{G}_\mathrm{c}$. Therefore, we present the following theorem (proved in the Appendix~I-A) to formally describe the relationship between the Laplacian frequency component and the complementarity ratio among clients.
\begin{thm}
\label{theorem_proportional}
The Laplacian frequency component of the federated collaboration graph $f(\bm{\Theta})= \frac{1}{2} r_\mathrm{c} + o(r^2(\bm{\theta}_i, \bm{\theta}_j))$, where ``$o(\cdot)$'' denotes the little-o notation.
\end{thm}
Theorem~\ref{theorem_proportional} means that the Laplacian frequency component of the federated collaboration graph $\mathcal{G}_\mathrm{c}$ is proportional to $r_\mathrm{c}$. Then, we prove in the Appendix~I-B that
\begin{thm}
\label{theorem_equivalent}
The Laplacian frequency component of the federated collaboration graph $f(\bm{\Theta})=\frac{1}{4} \sum_{\langle i, j\rangle \in \mathcal{E}_\mathrm{c}} \mathbf{W}_\mathrm{c}^{ij} \Vert \bm{\theta}_i - \bm{\theta}_j\Vert^2_2$.
\end{thm}
Theorem~\ref{theorem_equivalent} demonstrates that the Laplacian frequency component of the federated collaboration graph $\mathcal{G}_\mathrm{c}$ is equivalent to the measure of heterogeneity in GFL. Consequently, combining theorem~\ref{theorem_proportional} and theorem~\ref{theorem_equivalent}, we can naturally have
\begin{thm}
\label{theorem_proportional_equivalent}
The measure of heterogeneity in GFL $H(\mathcal{G}_\mathrm{c})= 2r_\mathrm{c} + o(r^2(\bm{\theta}_i, \bm{\theta}_j))$.
\end{thm}
Theorem~\ref{theorem_proportional_equivalent} indicates that the heterogeneity in GFL is proportional to the complementary ratio. That is, the higher the complementary ratio of clients, the higher the heterogeneity in GFL. This theoretical result is quite consistent with the practical situation. Furthermore, theorem~\ref{theorem_proportional_equivalent} inspires us that to deal with the heterogeneity in GFL, we should pursue not only similarities between different clients but also complementarities between different clients.

\section{Our Proposed Method}
\begin{figure}[t]
	\centering
	\includegraphics[width=9cm]{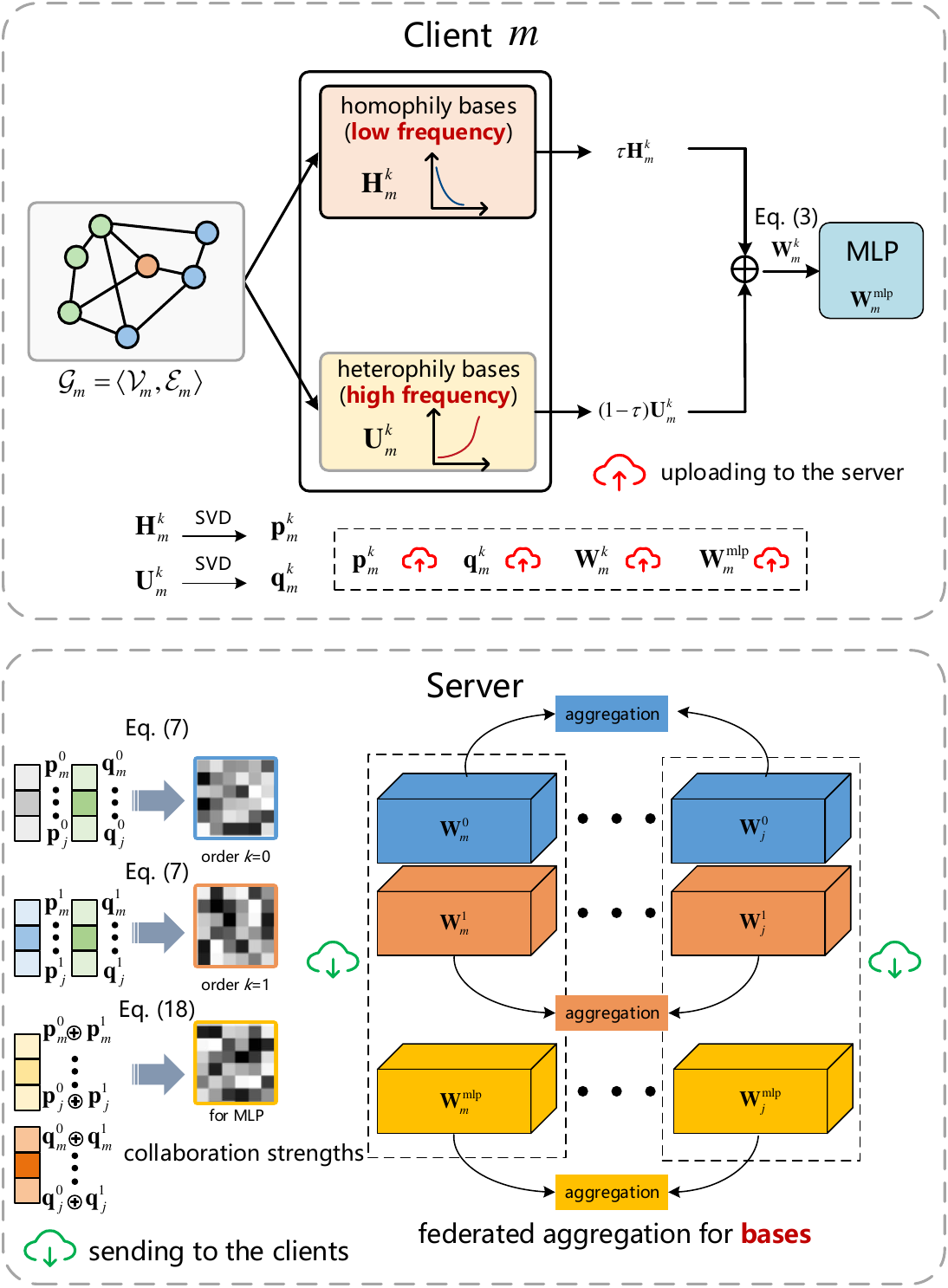}
	\caption{The framework of our proposed FedGSP. On the client side, low-frequency information and high-frequency information are captured by the homophily bases and heterophily bases, respectively. On the server side, we optimize the collaboration strengths and then perform the federated aggregation for polynomial bases.}
	\label{fig3}
\end{figure}

In this section, we provide the details of our proposed \underline{\textbf{Fed}}erated learning method by mining \underline{\textbf{G}}raph \underline{\textbf{S}}pectral \underline{\textbf{P}}roperties (FedGSP). First, we describe our introduced spectral GNN, including the construction of polynomial bases. Second, we formulate and solve an optimization problem to obtain the trade-off between sharing generic spectral properties and complementing non-generic spectral properties. Third, we present the detailed process of our proposed federated aggregation method for polynomial bases. The framework of our proposed FedGSP is shown in Fig.~\ref{fig3}.

\subsection{Spectral GNN}
\label{SpectralGNN}
Since there are significant differences in the homophily levels of graphs on different clients, local models usually adapt to their local homophily levels, leading to inconsistent spectral properties across different clients. To deal with this problem, we aim to introduce the spectral GNN as the local model. Since spectral GNNs are made up of polynomial bases, they can be easily split into low-frequency and high-frequency parts for subsequent analysis. In this paper, we employ the polynomial filter-based spectral GNN (\textit{i.e.}, UniFilter~\cite{huanguniversal}) for its simplicity. Here, we use the UniFilter on the $m$-th client as an example to illustrate UniFilter's bases, including homophily bases and heterophily bases. First, based on the propagation matrix, the homophily bases $\mathbf{H}_m$ can be defined as $\mathbf{H}_m=[\mathbf{H}_m^0,\mathbf{H}_m^1,\cdots,\mathbf{H}_m^K]$, where $\mathbf{H}_m^K=\mathbf{P}_m^K\mathbf{X}_m$, $\mathbf{P}_m^K=(\mathbf{I}-\mathbf{L}_m)^K$, $K$ is the order of homophily bases, and $[\cdot, \cdot]$ denotes the concatenation operation. Based on the definition of $\mathbf{H}_m$, we can find its physical meaning that the graph signal is propagated to $K$-hop neighbors via the propagation matrix $\mathbf{P}_m$. Therefore, this characteristic enables the homophily bases to effectively capture low-frequency information in the graph. Second, according to~\cite{huanguniversal}, the heterophily bases $\mathbf{U}_m$ can be defined as $\mathbf{U}_m=[\mathbf{U}_m^0,\mathbf{U}_m^1,\cdots,\mathbf{U}_m^K]$. Due to space limitations, details of the construction of heterophily bases are provided in the Appendix~II. According to~\cite{huanguniversal}, the heterophily bases are prone to effectively capture high-frequency information in the graph. Given the constructed homophily and heterophily bases, the UniFilter can be defined as
\begin{equation}\label{eq1}
\mathbf{Z}_{m}=\sum_{k=0}^{K}\mathbf{W}_{m}^{k}\left(\tau\mathbf{H}_{m}^{k}+(1-\tau)\mathbf{U}_{m}^{k}\right),
\end{equation}
where $\mathbf{W}_{m}^{k}$ represents the learnable coefficients, $\tau \in [0, 1]$ is the hyperparameter adjusting the impact of homophily and heterophily bases, and $\mathbf{Z}_{m}$ denotes the filtered spectral signal. Then, $\mathbf{Z}_{m}$ is fed into a Multi-Layer Perceptron (MLP) to obtain the node classification results, namely
\begin{equation}\label{eq_mlp}
\hat{\mathbf{Y}}_{m}=\sigma(\mathbf{Z}_{m} \mathbf{W}_m^{\mathrm{mlp}} ),
\end{equation}
where $\mathbf{W}_m^{\mathrm{mlp}}$ represents the parameters of MLP, $\sigma$ denotes the nonlinear activation function (\textit{i.e.}, rectified linear unit (ReLU)~\cite{nair2010rectified} in our FedGSP), and $\hat{\mathbf{Y}}_{m}$ denotes the predicted results. Therefore, the learnable parameters on the $m$-th client consist of $\mathbf{W}_m^0, \mathbf{W}_m^1, \cdots, \mathbf{W}_m^K$, and $\mathbf{W}_m^{\mathrm{mlp}}$.

\subsection{Optimization of the Collaboration Strengths}
\label{Optimization_of_Collaboration}
\begin{figure}[t]
	\centering
	\includegraphics[width=8cm]{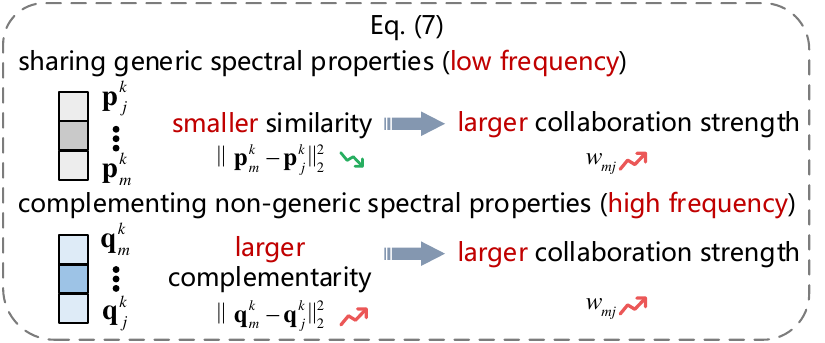}
	\caption{To obtain the trade-off between sharing generic spectral properties and complementing non-generic spectral properties, an optimization problem with respect to the homophily bases and heterophily bases is formulated.}
	\label{fig4}
\end{figure}

To obtain the trade-off between sharing generic spectral properties (\textit{i.e.}, low-frequency information) and complementing non-generic spectral properties (\textit{i.e.}, high-frequency information), we aim to formulate an optimization problem to determine the collaboration strengths (\textit{i.e.}, $\mathbf{W}_\mathrm{c}$) among clients. However, generic spectral properties (\textit{i.e.}, low-frequency information) and non-generic spectral properties (\textit{i.e.}, high-frequency information) can not be directly obtained. Fortunately, according to~\cite{huanguniversal}, low-frequency information and high-frequency information are captured by the homophily bases and heterophily bases, respectively. Therefore, we can formulate a multi-objective optimization problem with respect to the homophily bases and heterophily bases, \textit{i.e.}, pursuing the similarity of homophily bases while promoting the complementarity of heterophily bases. Fig.~\ref{fig4} is the schematic diagram to illustrate this optimization objective. However, due to privacy concerns, it is not allowed to upload the homophily and heterophily bases directly to the server. Therefore, inspired by~\cite{yan2024balancing}, we design a pre-processing procedure to extract the principal components of the homophily and heterophily bases. Specifically, we first apply the Singular Value Decomposition (SVD) on the homophily and heterophily bases, respectively.
For example, the SVD of $\mathbf{H}_m^k$ and $\mathbf{U}_m^k$ can be represented as
\begin{equation}\label{eq2_1}
\mathbf{H}_m^k=\mathbf{Q}_m^{k,p}\mathbf{\Sigma}_m^{k,p}(\mathbf{V}_m^{k,p})^\mathrm{T},
\end{equation}
\begin{equation}\label{eq2_2}
\mathbf{U}_m^k=\mathbf{Q}_m^{k,q}\mathbf{\Sigma}_m^{k,q}(\mathbf{V}_m^{k,q})^\mathrm{T},
\end{equation}
where $\mathbf{Q}_m^{k,p}$ and $\mathbf{Q}_m^{k,q}$ contain the singular vectors of $\mathbf{H}_m^k$ and $\mathbf{U}_m^k$, respectively. Second, we select the first $t$ columns of $\mathbf{Q}_m^{k,p}$ and $\mathbf{Q}_m^{k,q}$ and then flatten them to the vectors, denoted as $\mathbf{p}_m^{k}$ and $\mathbf{q}_m^{k}$, respectively. By doing so, we not only eliminate the risk of data leakage but also greatly reduce the subsequent computation overhead.

Since each client has $K + 1$ homophily and heterophily bases, we can therefore construct $K + 1$ adjacency matrixes to separately determine the collaboration strength concerning the $k$-th order of bases ($k=0,1, \cdots, K$). In other words, there are $K+1$ optimization problems related to the bases (\textit{i.e.}, optimizations of $\mathbf{W}_\mathrm{c_0}, \mathbf{W}_\mathrm{c_1}, \cdots, \mathbf{W}_\mathrm{c_K}$). We take the $0$-th order of bases as the example to present the optimization process of $\mathbf{W}_\mathrm{c_0}$. For ease of expression, here we abbreviate the $\mathbf{W}_\mathrm{c_0}$, $\mathbf{p}_m^{0}$, and $\mathbf{q}_m^{0}$ as $\mathbf{W}_\mathrm{c}$, $\mathbf{p}_m$, and $\mathbf{q}_m$, respectively. Now we describe the objectives of our optimization problem, respectively. First, to pursue the similarity of homophily bases, a smaller Euclidean distance between homophily bases (\textit{i.e.}, $\Vert \mathbf{p}_i - \mathbf{p}_j \Vert^2_2$) means a larger collaboration strength between the $i$-th and the $j$-th client (\textit{i.e.}, $w_{ij}$). Second, to pursue the complementarity of heterophily bases, a larger Euclidean distance between heterophily bases (\textit{i.e.}, $\Vert \mathbf{q}_i - \mathbf{q}_j \Vert^2_2$) means a larger collaboration strength between the $i$-th and the $j$-th client (\textit{i.e.}, $w_{ij}$). Third, to adaptively adjust the importance of different dimensions of bases~\cite{nie2020self}, we assign learnable self-attention weights (\textit{i.e.}, $\mathbf{R}$ and $\mathbf{S}$) to homophily and heterophily bases, respectively. In summary, the optimization objective can be written as
\begin{equation}\label{eq_opt_objective}
\begin{aligned}
&\operatorname*{min}_{\mathbf{W}_c, \mathbf{R}, \mathbf{S}} \sum_{i=1}^{M} \sum_{j=1}^{M}(\Vert \mathbf{R} \mathbf{p}_i - \mathbf{R} \mathbf{p}_j \Vert_2^{2}w_{ij}-\Vert \mathbf{S} \mathbf{q}_i - \mathbf{S} \mathbf{q}_j \Vert_2^{2}w_{ij} \\
&\quad \quad \quad \quad \quad \quad \quad + \gamma w_{ij}^2),\\
&\mathrm{s.t.} \quad  \forall i, \quad \mathbf{w}_i^{\top} \mathbf{1} = 1, \quad w_{ij} \geq 0,\\
& \quad \quad \mathbf{r}^{\top} \mathbf{1} = 1, r_l \geq 0, l=1,2, \cdots, d\\
& \quad \quad \mathbf{s}^{\top} \mathbf{1} = 1, s_l \geq 0, l=1,2, \cdots, d\\
& \quad \quad \mathbf{R} = \mathrm{diag}(\mathbf{r}), \mathbf{S} = \mathrm{diag}(\mathbf{s}),
\end{aligned}
\end{equation}
where $\mathbf{r} \in \mathbb{R}^d$ and $\mathbf{s} \in \mathbb{R}^d$ are learnable self-attention vectors, $\mathbf{R} \in \mathbb{R}^{d \times d}$ and $\mathbf{S} \in \mathbb{R}^{d \times d}$ are diagonal matrices, $\mathbf{w}_i \in \mathbb{R}^{M \times 1}$ is the $i$-th column of $\mathbf{W}_c$, and $w_{ij}$ is the element of $\mathbf{W}_c$. Inspired by~\cite{nie2014clustering}, we employ a regularization term, \textit{i.e.}, $\gamma w_{ij}^2$, where $\gamma$ is the regularization parameter. Without this regularization term, the optimization problem~\ref{eq_opt_objective} would result in a trivial solution. Moreover, to further avoid the trivial solution, the sum of each column of $\mathbf{W}_c$ is constrained to be 1 (\textit{i.e.}, $\mathbf{w}_i^{\top}\mathbf{1}=1$, where $\mathbf{1}$ denotes the all-one column vector). Similarly, $\mathbf{r}^{\top} \mathbf{1} = 1$ and $\mathbf{s}^{\top} \mathbf{1} = 1$ are used to avoid trivial solutions of $\mathbf{r}$ and $\mathbf{s}$, respectively. To solve the optimization objective in Eq.~\eqref{eq_opt_objective}, inspired by~\cite{nie2020self}, we update three variables (\textit{i.e.}, $\mathbf{W}_c$, $\mathbf{R}$, and $\mathbf{S}$) one by one in an iterative manner.

\subsubsection{Update \texorpdfstring{$\mathbf{R}$}{} with \texorpdfstring{$\mathbf{W}_c$}{} and \texorpdfstring{$\mathbf{S}$}{} Fixed}
\label{subsection_update1}
Since $\mathbf{W}_c$ and $\mathbf{S}$ are fixed, the optimization objective in Eq.~\eqref{eq_opt_objective} can be written as
\begin{equation}\label{eq_update1}
\begin{aligned}
&\operatorname*{min}_{\mathbf{R}} \sum_{i=1}^{M} \sum_{j=1}^{M}\Vert \mathbf{R} \mathbf{p}_i - \mathbf{R} \mathbf{p}_j \Vert_2^{2}w_{ij},\\
&\mathrm{s.t.} \quad \mathbf{r}^{\top} \mathbf{1} = 1, r_l \geq 0, l=1,2, \cdots, d\\
& \quad \quad \mathbf{R} = \mathrm{diag}(\mathbf{r}).
\end{aligned}
\end{equation}
Here we introduce an elementary and very important theorem in spectral analysis~\cite{8047308, luo2021bi} to solve the above problem.
\begin{thm}
\label{theorem_introduce}
Given a matrix $\mathbf{O} \in \mathbb{R}^{M \times d}$, the $i$-th row of $\mathbf{O}$ can be denoted as $\mathbf{o}_i^{\top} \in \mathbb{R}^{1 \times d}$. Consider the Laplacian matrix $\mathcal{L}_\mathrm{c}$ of the adjacency matrix $\mathbf{W}_\mathrm{c}$, we can have
\begin{equation}
2\mathrm{Tr}(\mathbf{O}^{\top} \mathcal{L}_\mathrm{c} \mathbf{O})  = \sum_{i=1}^M \sum_{j=1}^M \Vert \mathbf{o}_i - \mathbf{o}_j\Vert^2_2 \mathbf{W}_\mathrm{c}^{ij}. 
\end{equation}
\end{thm}
According to the theorem~\ref{theorem_introduce}, we set the $\mathbf{o}_i = \mathbf{R} \mathbf{p}_i \in \mathbb{R}^{d \times 1} \Rightarrow \mathbf{O} = \mathbf{P}\mathbf{R} $. Therefore, Eq.~\eqref{eq_update1} becomes
\begin{equation}
\begin{aligned}
&\operatorname*{min}_{\mathbf{O}} \mathrm{Tr}(\mathbf{O}^{\top} \mathcal{L}_\mathrm{c} \mathbf{O}) \Rightarrow \operatorname*{min}_{\mathbf{R}} \mathrm{Tr}(\mathbf{R}\mathbf{P}^{\top} \mathcal{L}_\mathrm{c} \mathbf{P}\mathbf{R}),\\
&\mathrm{s.t.} \quad \mathbf{r}^{\top} \mathbf{1} = 1, r_l \geq 0, l=1,2, \cdots, d\\
& \quad \quad \mathbf{R} = \mathrm{diag}(\mathbf{r}).
\end{aligned}
\end{equation}
Then, we employ the Lagrange multiplier method to solve the above problem.
Due to space limitations, the detailed optimization process can be found in the Appendix~III-A. Finally, we can have
\begin{equation}\label{ep_update2_0}
r_i = \frac{1}{e_i^* \sum_{i=1}^d \frac{1}{e_i^*}},
\end{equation}
where $e_i^*=\mathbf{p}^{\top}_{[:,i]} \mathcal{L}_\mathrm{c} \mathbf{p}_{[:,i]}$, $\mathbf{p}_{[:,i]} \in \mathbb{R}^{M \times 1}$ is the $i$-th column of $\mathbf{P}$, and $\mathbf{P} = [\mathbf{p}_1, \mathbf{p}_2, \cdots, \mathbf{p}_M]^{\top}$.

\subsubsection{Update \texorpdfstring{$\mathbf{S}$}{} with \texorpdfstring{$\mathbf{W}_c$}{} and \texorpdfstring{$\mathbf{R}$}{} Fixed}
The update problem of $\mathbf{S}$ when $\mathbf{W}_c$ and $\mathbf{R}$ are fixed is quite similar to the above problem in the Section~\ref{subsection_update1}. For
simplicity, we leave the details in the Appendix~III-B and directly give the result here. Specifically, we can have
\begin{equation}\label{eq_update2_0}
s_i = \frac{1}{n_i^* \sum_{i=1}^d \frac{1}{n_i^*}},
\end{equation}
where $n_i^*=\mathbf{q}^{\top}_{[:,i]} \mathcal{L}_\mathrm{c} \mathbf{q}_{[:,i]}$, $\mathbf{q}_{[:,i]} \in \mathbb{R}^{M \times 1}$ is the $i$-th column of $\mathbf{Q}$, and $\mathbf{Q} = [\mathbf{q}_1, \mathbf{q}_2, \cdots, \mathbf{q}_M]^{\top}$.

\subsubsection{Update \texorpdfstring{$\mathbf{W}_c$}{} with \texorpdfstring{$\mathbf{S}$}{} and \texorpdfstring{$\mathbf{R}$}{} Fixed}
Since $\mathbf{S}$ and $\mathbf{R}$ are fixed, the optimization objective in Eq.~\eqref{eq_opt_objective} can be written as
\begin{equation}\label{eq_update3}
\begin{aligned}
&\operatorname*{min}_{\mathbf{W}_c} \sum_{i=1}^{M} \sum_{j=1}^{M}(\Vert \mathbf{R} \mathbf{p}_i - \mathbf{R} \mathbf{p}_j \Vert_2^{2}w_{ij}-\Vert \mathbf{S} \mathbf{q}_i - \mathbf{S} \mathbf{q}_j \Vert_2^{2}w_{ij} \\
&\quad \quad \quad \quad \quad \quad + \gamma w_{ij}^2),\\
&\mathrm{s.t.} \quad  \forall i, \quad \mathbf{w}_i^{\top} \mathbf{1} = 1, \quad w_{ij} \geq 0.
\end{aligned}
\end{equation}
Since the optimization problem in Eq.~\eqref{eq_update3} is independent across different clients for a specific client $i$, we can reformulate it as a subproblem tailored for the $i$-th client. Therefore, we can have 
\begin{equation}\label{eq_update3_1}
\begin{aligned}
&\operatorname*{min}_{\mathbf{w}_i} \sum_{j=1}^{M} (\Vert \mathbf{R} \mathbf{p}_i - \mathbf{R} \mathbf{p}_j \Vert_2^{2}w_{ij}-\Vert \mathbf{S} \mathbf{q}_i - \mathbf{S} \mathbf{q}_j \Vert_2^{2}w_{ij} \\
&\quad \quad \quad \quad \quad + \gamma w_{ij}^2),\\
&\mathrm{s.t.} \quad \quad \mathbf{w}_i^{\top} \mathbf{1} = 1, \quad w_{ij} \geq 0.
\end{aligned}
\end{equation}
Then, inspired by~\cite{nie2020self}, we employ the Lagrange multiplier method and the Newton method to solve the above problem. Due to space limitations, the detailed optimization process can be found in the Appendix~III-C. Finally, we can have
\begin{equation}\label{eq_update3_12}
w_{ij}^* = (h_{ij} - \hat{b}_i^*)_+,
\end{equation}
\begin{equation}\label{eq_update3_14}
\hat{b}_i^* = \frac{1}{M}\sum_{j=1}^{M}(\hat{b}_i^* - h_{ij})_+, 
\end{equation}
where $h_{ij}=\frac{1}{M} - \frac{t_{ij}}{2\gamma} + \frac{\mathbf{1}^{\top}\mathbf{t}_i}{2M\gamma}$, $t_{ij} = \Vert \mathbf{R} \mathbf{p}_i - \mathbf{R} \mathbf{p}_j \Vert_2^{2}-\Vert \mathbf{S} \mathbf{q}_i - \mathbf{S} \mathbf{q}_j \Vert_2^{2} $, and the optimal $\hat{b}_i^*$ can be found by the Newton method.

\subsection{Federated Aggregation for Polynomial Bases}
Based on the optimized collaboration strengths, we can perform the federated aggregation. However, existing aggregation methods usually work on model weights or gradients, rarely on polynomial bases. Therefore, here we explore the aggregation method for polynomial bases. First, since polynomial bases are fundamental components in the spectral GNNs, we have to take into account their spectral properties when performing the federated aggregation. In Eq.~\eqref{eq1}, different orders of polynomial bases mean different propagation hops. For example, if $K=3$, the graph signal is propagated to 3-hop neighbors. Therefore, we have to allow learnable coefficients $\mathbf{W}_{m}^{k}$ to be federated separately according to each order $k$. Second, since polynomial bases may consist of information related to the graph data, we perform federated aggregation on coefficients instead of bases. Specifically, with the optimized collaboration strengths matrix $\mathbf{W}_\mathrm{c_k}$ corresponding to the $k$-th order of bases, we conduct the weighted averaging of coefficients across different clients. Our proposed separate federation can be defined as
\begin{equation}\label{eq13}
\overline {\mathbf{W}}_m^k\leftarrow \sum_{j=1}^{M} w_{mj}^{k} \cdot \mathbf{W}_j^k,
\end{equation}
where $w_{mj}^{k}$ denotes the $m,j$-th element of $\mathbf{W}_\mathrm{c_k}$ and $\overline {\mathbf{W}}_m^k$ represents the aggregated coefficient.

Meanwhile, we consider the aggregation of $\mathbf{W}_m^{\mathrm{mlp}}$. First, we have to obtain the collaboration strength matrix corresponding to the parameters of MLP (\textit{i.e.}, $\mathbf{W}_\mathrm{c_{MLP}}$), which can be written as
\begin{equation}\label{eq_opt_objective_mlp}
\begin{aligned}
&\operatorname*{min}_{\mathbf{W}_\mathrm{c_{MLP}}, \mathbf{R}_\mathrm{MLP}, \mathbf{S}_\mathrm{MLP}} \sum_{i=1}^{M} \sum_{j=1}^{M}(\Vert \mathbf{R}_\mathrm{MLP} \hat{\mathbf{p}}_i - \mathbf{R}_\mathrm{MLP} \hat{\mathbf{p}}_j \Vert_2^{2}w_{ij}\\
& \quad \quad \quad \quad \quad \quad -\Vert \mathbf{S}_\mathrm{MLP}\hat{\mathbf{q}}_i - \mathbf{S}_\mathrm{MLP} \hat{\mathbf{q}}_j \Vert_2^{2}w_{ij} \\
&\quad \quad \quad \quad \quad \quad \quad + \gamma w_{ij}^2),\\
&\mathrm{s.t.} \quad  \forall i, \quad \mathbf{w}_i^{\top} \mathbf{1} = 1, \quad w_{ij} \geq 0,\\
& \quad \quad \mathbf{r}^{\top} \mathbf{1} = 1, r_l \geq 0, l=1,2, \cdots, d\\
& \quad \quad \mathbf{s}^{\top} \mathbf{1} = 1, s_l \geq 0, l=1,2, \cdots, d\\
& \quad \quad \mathbf{R}_\mathrm{MLP} = \mathrm{diag}(\mathbf{r}),\mathbf{S}_\mathrm{MLP} = \mathrm{diag}(\mathbf{s}),
\end{aligned}
\end{equation}
where $\hat{\mathbf{p}}_i = [\mathbf{p}_i^{0}, \mathbf{p}_i^{1}, \cdots, \mathbf{p}_i^{K}]$, $\hat{\mathbf{p}}_j = [\mathbf{p}_j^{0}, \mathbf{p}_j^{1}, \cdots, \mathbf{p}_j^{K}]$, $\hat{\mathbf{q}}_i = [\mathbf{q}_i^{0}, \mathbf{q}_i^{1}, \cdots, \mathbf{q}_i^{K}]$, $\hat{\mathbf{q}}_j = [\mathbf{q}_j^{0}, \mathbf{q}_j^{1}, \cdots, \mathbf{q}_j^{K}]$, $\mathbf{w}_i \in \mathbb{R}^{M \times 1}$ is the $i$-th column of $\mathbf{W}_\mathrm{c_{MLP}}$,  $\mathbf{R}_\mathrm{MLP} \in \mathbb{R}^{d \times d}$ and $\mathbf{S}_\mathrm{MLP} \in \mathbb{R}^{d \times d}$ are self-attention matrices, and $w_{ij}$ is the element of $\mathbf{W}_\mathrm{c_{MLP}}$. Then, the federation of $\mathbf{W}_m^{\mathrm{mlp}}$ can be defined as
\begin{equation}\label{eq14}
\overline {\mathbf{W}}_m^{\mathrm{mlp}}\leftarrow \sum_{j=1}^{M} w_{mj}^{\mathrm{mlp}} \cdot \mathbf{W}_j^{\mathrm{mlp}},
\end{equation}
where $w_{mj}^{\mathrm{mlp}}$ denotes the $m,j$-th element of $\mathbf{W}_\mathrm{c_{MLP}}$, and $\overline {\mathbf{W}}_m^{\mathrm{mlp}}$ represents the aggregated parameter. Eq.~\eqref{eq14} can be similarly optimized by using the methods in the Section~\ref{Optimization_of_Collaboration}. Finally, Algorithm~\ref{algorithm_client} and Algorithm~\ref{algorithm_server} summarize the main steps of our proposed FedGSP for the clients and the server, respectively.

\begin{algorithm}[t]
\caption{\textbf{FedGSP} Client Algorithm}\label{algorithm_client}

{\textbf{Input:}} Number of local epochs $E$; the order of bases $K$; number of selected columns $t$; $\mathcal{G}_m$ on client $m$; node feature matrix $\mathbf{X}_m$ on client $m$; label matrix $\mathbf{Y}_{m}$ on client $m$; coefficients of bases $\mathbf{W}_m^{0:K}$ on client $m$; parameters of MLP $\mathbf{W}_m^{\mathrm{mlp}}$ on client $m$; federated parameters $\overline{\mathbf{W}}_m^{\mathrm{mlp}}$ and $\overline{\mathbf{W}}_{m}^{0:K}$ from the server.

{\textbf{Output:}} Predicted label $\hat{\mathbf{Y}}_{m}$.
\begin{algorithmic}[1]
\STATE Download federated parameters $\overline{\mathbf{W}}_m^{\mathrm{mlp}}$ and $\overline{\mathbf{W}}_{m}^{0:K}$ from the server;
\STATE $\mathbf{W}_m^{\mathrm{mlp}}\leftarrow \overline{\mathbf{W}}_m^{\mathrm{mlp}}$, $\mathbf{W}_{m}^{0:K}\leftarrow \overline{\mathbf{W}}_{m}^{0:K}$;
\STATE Construct the homophily bases $\mathbf{H}_m$ and heterophily bases $\mathbf{U}_m$ for $\mathcal{G}_m$, respectively;
\STATE \textbf{for} each local epoch $e$ from 1 to $E$ \textbf{do}
\STATE \hspace{0.3cm} Obtain the ﬁltered spectral signal $\mathbf{Z}_{m}$ via Eq.~\eqref{eq1};
\STATE \hspace{0.3cm} Obtain the predicted results $\hat{\mathbf{Y}}_{m}$ via Eq.~\eqref{eq_mlp};
\STATE \hspace{0.3cm} Use cross-entropy loss to update $\mathbf{W}_m^{\mathrm{mlp}}$ and $\mathbf{W}_{m}^{0:K}$;
\STATE \hspace{0.3cm} \textbf{for} $k$ from 0 to $K$ \textbf{do}
\STATE \hspace{0.6cm} Operate the SVD of $\mathbf{H}_m$ and $\mathbf{U}_m$ via Eq.~\eqref{eq2_1} and Eq.~\eqref{eq2_2} to obtain $\mathbf{Q}_m^{k,p}$ and $\mathbf{Q}_m^{k,q}$, respectively;
\STATE \hspace{0.6cm} Select the first $t$ columns of $\mathbf{Q}_m^{k,p}$ and $\mathbf{Q}_m^{k,q}$ and then flatten them to obtain $\mathbf{p}_m^{k}$ and $\mathbf{q}_m^{k}$, respectively;
\STATE \hspace{0.3cm} \textbf{end}
\STATE \textbf{end}
\STATE Upload $\mathbf{W}_m^{\mathrm{mlp}}$, $\mathbf{W}_{m}^{0:K}$, $\mathbf{p}_m^{0:K}$, and $\mathbf{q}_m^{0:K}$ to the server.
\end{algorithmic}
\end{algorithm}

\begin{algorithm}[t]
\caption{\textbf{FedGSP} Server Algorithm}\label{algorithm_server}

{\textbf{Input:}} Number of rounds $R$; number of clients $M$; the order of bases $K$; coefficients of bases $\mathbf{W}_m^{0:K}$ from client $m$; parameters of MLP $\mathbf{W}_m^{\mathrm{mlp}}$ from client $m$; pre-processed vectors $\mathbf{p}_m^{0:K}$ and $\mathbf{q}_m^{0:K}$ from client $m$.

{\textbf{Output:}} Federated parameters for clients.

\begin{algorithmic}[1]
\STATE Initialize parameters $({\overline{\mathbf{W}}^{\mathrm{mlp}}})^{(1)}$ and $({\overline{\mathbf{W}}^{0:K}})^{(1)}$;
\STATE \textbf{for} each round $r$ from 1 to $R$ \textbf{do}
\STATE \hspace{0.3cm} \textbf{for} client $m \in \{1, 2, \cdots, M\}$ \textbf{in parallel do}
\STATE \hspace{0.6cm} \textbf{if} $r=1$ \textbf{then}
\STATE \hspace{0.9cm} Send $({\overline{\mathbf{W}}^{\mathrm{mlp}}})^{(r)}$ and $({\overline{\mathbf{W}}^{0:K}})^{(r)}$ to client $m$;
\STATE \hspace{0.6cm} \textbf{end}
\STATE \hspace{0.6cm} \textbf{else}
\STATE \hspace{0.9cm} Receive $\mathbf{W}_m^{\mathrm{mlp}}$ and $\mathbf{W}_{m}^{0:K}$ from client $m$;
\STATE \hspace{0.9cm} Receive $\mathbf{p}_m^{0:K}$ and $\mathbf{q}_m^{0:K}$ from client $m$;
\STATE \hspace{0.9cm} \textbf{for} $k$ from 0 to $K$ \textbf{do}
\STATE \hspace{1.2cm} Optimize $\mathbf{W}_\mathrm{c_k}$ via Eq.~\eqref{eq_opt_objective} to Eq.~\eqref{eq_update3_14};
\STATE \hspace{0.9cm} \textbf{end}
\STATE \hspace{0.9cm} Obtain $({\overline{\mathbf{W}}_{m}^{0:K}})^{(r)}$ via Eq.~\eqref{eq13};
\STATE \hspace{0.9cm} Optimize $\mathbf{W}_\mathrm{c_{MLP}}$ by solving Eq.~\eqref{eq_opt_objective_mlp};
\STATE \hspace{0.9cm} Obtain  $({\overline{\mathbf{W}}^{\mathrm{mlp}}_m})^{(r)}$ via Eq.~\eqref{eq14};
\STATE \hspace{0.9cm} Send $({\overline{\mathbf{W}}^{\mathrm{mlp}}_m})^{(r)}$ and $({\overline{\mathbf{W}}_{m}^{0:K}})^{(r)}$ to client $m$;
\STATE \hspace{0.6cm} \textbf{end}
\STATE \hspace{0.6cm} Perform Algorithm~\ref{algorithm_client} on client $m$;
\STATE \hspace{0.3cm} \textbf{end}
\STATE \textbf{end}
\end{algorithmic}
\end{algorithm}

\subsection{Efficiency Analysis}
Here we present the spatial and temporal complexities of different methods on the client and server sides in Tab.~\ref{table_comple}, respectively. Due to space limitations, the detailed efficiency analysis of our proposed FedGSP is provided in the Appendix~IV. According to Tab.~\ref{table_comple}, we can find that the temporal complexity of our proposed FedGSP is mainly related to the optimization process (\textit{i.e.}, $\mathcal{O}\big(M \times d \times (M + d + 1)\big)$), which is independent of the local model training. Therefore, in the practical implementation, we implement the optimization process and the local model training in parallel, so that our proposed FedGSP is efficient in real-world applications, which is also validated by our experiments (see Section~\ref{time_communication}).

\begin{table*}[t]
  \centering
  \caption{The spatial and temporal complexities of different methods on the client and server sides, respectively. Here $M$, $K$, $d$, $n_m$, $e_m$, and $c_m$ denote the number of clients, orders, dimensions, nodes, edges, and classes, respectively.}
    \label{table_comple}
    \renewcommand{\arraystretch}{0.8} 
    \scalebox{0.82}{
    \begin{tabular}{c|cc|cc}
      \hline
      \rowcolor{gray!50}
      Method                               & Client Spa. Comp. & Server Spa. Comp. & Client Temp. Comp. & Server Temp. Comp. \\ \hline
      FedAvg~\cite{mcmahan2017communication}        &  $\mathcal{O}(d  + d^2)$                 & $\mathcal{O}\big(M \times (1 + d^2)\big)$              & $\mathcal{O}(e_m \times d + n_m \times d^2)$                   &  $\mathcal{O}(M)$                  \\
      FedProx~\cite{MLSYS2020_1f5fe839}             &  $\mathcal{O}(d  + 2d^2)$                & $\mathcal{O}\big(M \times (1 + d^2)\big)$             & $\mathcal{O}(e_m \times d + n_m \times d^2  + d^2)$                   &  $\mathcal{O}(M)$                   \\
      FedPer~\cite{Arivazhagan2019} &  $\mathcal{O}(d  + 2d^2)$                & $\mathcal{O}\big(M \times (1 + d^2)\big)$             & $\mathcal{O}(e_m \times d + n_m \times d^2  + d^2)$                   &  $\mathcal{O}(M)$                   \\
      GCFL~\cite{NEURIPS2021_9c6947bd}              &  $\mathcal{O}(d  + d^2)$                 & $\mathcal{O}\big(M \times (1 + 2d^2)\big)$             & $\mathcal{O}(e_m \times d + n_m \times d^2)$                   & $\mathcal{O}\big(M + M^2 \times (\log M + d^2)\big)$                    \\
      FedGNN~\cite{wu2021fedgnn} &  $\mathcal{O}(d  + 2d^2)$                 & $\mathcal{O}\big(2M \times (1 + 2d^2)\big)$             & $\mathcal{O}(e_m \times d + n_m \times d^2 + d)$                   & $\mathcal{O}(M)$                    \\
      FedSage+\cite{NEURIPS2021_34adeb8e}           &  $\mathcal{O}(n_m \times d  + 3d^2)$     & $\mathcal{O}\big(M \times (1 + 3d^2)\big)$             &  $\mathcal{O}(e_m \times d + n_m \times d^2)$                  &  $\mathcal{O}(M)$                  \\
      FED-PUB~\cite{baek2023personalized} &  $\mathcal{O}(n_m \times d  + d^2)$     & $\mathcal{O}\big(M \times (d^2 + M)\big)$             &  $\mathcal{O}(e_m \times d + n_m \times d^2)$                  &  $\mathcal{O}\big(M \times d \times (M + d)\big)$                  \\
      FedGTA~\cite{li2023fedgta}                    &  $\mathcal{O}(d  + d^2 + K \times c_m)$  & $\mathcal{O}\big(M \times (1 + d^2 + M \times K \times c_m)\big)$            &  $\mathcal{O}\big(e_m \times (d + n_m \times c_m) + n_m \times ( d^2 + c_m)\big)$                  & $\mathcal{O}(M + M \times K \times c_m)$                   \\
      AdaFGL~\cite{li2024adafgl} &  $\mathcal{O}(n_m \times d  + 2d^2)$  & $\mathcal{O}\big(M \times (1 + d^2)\big)$            &  $\mathcal{O}\big(K \times (e_m \times d + e_m \times n_m+ n_m \times d^2)\big)$                  & $\mathcal{O}(M)$                   \\
      FedTAD~\cite{zhu2024fedtad} &  $\mathcal{O}(n_m \times d  + 2d^2)$  & $\mathcal{O}\big(M \times (1 + d^2)\big)$            &  $\mathcal{O}(K \times n_m + e_m)$                  & $\mathcal{O}\big(n_m \times d \times (d + n_m + 2M \times c_m)\big)$                   \\
      FedIIH~\cite{wentao2025fediih} &  $\mathcal{O}(n_m \times d  + d^2)$     & $\mathcal{O}\big( M \times K \times (d^2 + M)\big)$             &  $\mathcal{O}\big(K \times (e_m \times d + n_m \times d^2)\big)$                  &  $\mathcal{O}\big(M \times d \times (K \times M + d)\big)$                  \\
      FedGSP (Ours) & $\mathcal{O}( K \times d  + d^2)$                  &  $\mathcal{O}\big( M \times K \times (d^2 + M)\big)$            &    $\mathcal{O}\big(K \times (e_m \times d + n_m \times d^2)\big)$                &  $\mathcal{O}\big(M \times d \times (M + d + 1) + M \times K\big)$                    \\ \hline
\end{tabular}
    }
\end{table*}

\section{Experiments}
To demonstrate the effectiveness of our proposed \underline{\textbf{Fed}}erated learning method by mining \underline{\textbf{G}}raph \underline{\textbf{S}}pectral \underline{\textbf{P}}roperties (FedGSP), intensive experiments are carried out on eleven well-known graph datasets, including six homophilic datasets and five heterophilic datasets. In particular, we perform experiments on a large-scale graph dataset (\textit{i.e.}, \textit{ogbn-arxiv}), which consists of more than millions of edges. First, we compare our proposed FedGSP with eleven state-of-the-art methods under both the non-overlapping and overlapping subgraph partitioning settings. Second, we conduct ablation studies to evaluate the effectiveness of sharing generic spectral properties and complementing non-generic spectral properties, respectively. Third, we perform case studies to analyze the spectral properties captured by the local model after federation. Fourth, we plot the convergence curves of our proposed FedGSP and the baseline methods. In addition, to reveal the advantage of our proposed FedGSP over the baseline methods in terms of efficiency, we report the communication round time of the proposed FedGSP and the compared baseline methods. Furthermore, we perform a detailed sensitivity analysis on the hyperparameters used in our proposed FedGSP. The implementation details are presented in the Appendix~V.

\subsection{Experimental Settings}
In our experiments, eleven state-of-the-art methods are employed for comparison. Concretely, we adopt one classic Federated Learning (FL) method (\textit{i.e.}, FedAvg~\cite{mcmahan2017communication}), two personalized FL methods (\textit{i.e.}, FedProx~\cite{MLSYS2020_1f5fe839} and FedPer~\cite{Arivazhagan2019}), three general GFL methods (\textit{i.e.}, GCFL~\cite{NEURIPS2021_9c6947bd}, FedGNN~\cite{wu2021fedgnn}, and FedSage+\cite{NEURIPS2021_34adeb8e}), and five personalized GFL methods (\textit{i.e.}, FED-PUB~\cite{baek2023personalized}, FedGTA~\cite{li2023fedgta}, AdaFGL~\cite{li2024adafgl}, FedTAD~\cite{zhu2024fedtad}, and FedIIH~\cite{wentao2025fediih}). To ensure the fairness of experiments, according to~\cite{baek2023personalized, wentao2025fediih}, all methods are repeated three times, where the mean accuracies and standard deviations are both reported.

\subsection{Experimental Results}
Our proposed FedGSP is compared with the above-mentioned baseline methods on both homophilic and heterophilic datasets.

\subsubsection{Homophilic Datasets}
Tab.~\ref{table1} and Tab.~\ref{table2} show the comparison results on the homophilic datasets in two partitioning settings, respectively. Our proposed FedGSP achieves the best performance among most of the baseline methods. Moreover, the standard deviations of FedGSP are relatively small, indicating that FedGSP is more stable than the compared baseline methods. Although the average accuracies of
FedGSP for all six datasets in both non-overlapping and overlapping scenarios are only slightly higher than the second-best method (\textit{i.e.}, FedIIH~\cite{wentao2025fediih}), our proposed FedGSP is more efficient than FedIIH (see the Section~\ref{time_communication}), achieving more than three times speed improvement. Since some conventional baseline methods can not well deal with the heterogeneity, they have large standard deviations. For example, the standard deviations of FedAvg, FedProx, and FedSage+ are 5.64, 4.56, and 5.94 on the \textit{Cora} datasets with 5 clients, respectively. Although some recent methods alleviate the node feature heterogeneity and structure heterogeneity to some extent, FedGSP still performs better than them. This is because FedGSP not only shares generic spectral properties but also complements non-generic spectral properties, so that it effectively learns graphs with varying homophily levels across different clients.

\begin{table*}[]
  \centering
  \scriptsize
  \caption{Accuracy comparison of various methods on six \textbf{homophilic} graph datasets in the \textbf{non-overlapping} subgraph partitioning setting. The best and second-best records on each dataset are highlighted in \textbf{bold} and \underline{underlined}, respectively.}
    \label{table1}
    \renewcommand{\arraystretch}{0.8} 
       \scalebox{0.82}{
  \begin{tabular}{lcccccccccc}
  \hline
  \rowcolor{gray!50}
  \textbf{}     & \multicolumn{3}{c}{Cora}                                                    & \multicolumn{3}{c}{CiteSeer}                                                & \multicolumn{3}{c}{PubMed}                                                  & -              \\ \cline{2-11} 
  Methods       & 5 Clients               & 10 Clients              & 20 Clients              & 5 Clients               & 10 Clients              & 20 Clients              & 5 Clients               & 10 Clients              & 20 Clients              & -              \\ \hline
  Local         & 81.30$\pm$0.21          & 79.94$\pm$0.24          & 80.30$\pm$0.25          & 69.02$\pm$0.05          & 67.82$\pm$0.13          & 65.98$\pm$0.17          & 84.04$\pm$0.18          & 82.81$\pm$0.39          & 82.65$\pm$0.03          & -              \\ \hline
  FedAvg~\cite{mcmahan2017communication}        & 74.45$\pm$5.64          & 69.19$\pm$0.67          & 69.50$\pm$3.58          & 71.06$\pm$0.60          & 63.61$\pm$3.59          & 64.68$\pm$1.83          & 79.40$\pm$0.11          & 82.71$\pm$0.29          & 80.97$\pm$0.26          & -              \\
  FedProx~\cite{MLSYS2020_1f5fe839}       & 72.03$\pm$4.56          & 60.18$\pm$7.04          & 48.22$\pm$6.18          & 71.73$\pm$1.11          & 63.33$\pm$3.25          & 64.85$\pm$1.35          & 79.45$\pm$0.25          & 82.55$\pm$0.24          & 80.50$\pm$0.25          & -              \\
  FedPer~\cite{Arivazhagan2019}        & 81.68$\pm$0.40          & 79.35$\pm$0.04          & 78.01$\pm$0.32          & 70.41$\pm$0.32          & 70.53$\pm$0.28          & 66.64$\pm$0.27          & 85.80$\pm$0.21          & 84.20$\pm$0.28          & 84.72$\pm$0.31          & -              \\
  GCFL~\cite{NEURIPS2021_9c6947bd}          & 81.47$\pm$0.65          & 78.66$\pm$0.27          & 79.21$\pm$0.70          & 70.34$\pm$0.57          & 69.01$\pm$0.12          & 66.33$\pm$0.05          & 85.14$\pm$0.33          & 84.18$\pm$0.19          & 83.94$\pm$0.36          & -              \\
  FedGNN~\cite{wu2021fedgnn}        & 81.51$\pm$0.68          & 70.12$\pm$0.99          & 70.10$\pm$3.52          & 69.06$\pm$0.92          & 55.52$\pm$3.17          & 52.23$\pm$6.00          & 79.52$\pm$0.23          & 83.25$\pm$0.45          & 81.61$\pm$0.59          & -              \\
  FedSage+\cite{NEURIPS2021_34adeb8e}      & 72.97$\pm$5.94          & 69.05$\pm$1.59          & 57.97$\pm$12.6          & 70.74$\pm$0.69          & 65.63$\pm$3.10          & 65.46$\pm$0.74          & 79.57$\pm$0.24          & 82.62$\pm$0.31          & 80.82$\pm$0.25          & -              \\
  FED-PUB~\cite{baek2023personalized}       & 83.70$\pm$0.19          & 81.54$\pm$0.12          & 81.75$\pm$0.56          & 72.68$\pm$0.44          & 72.35$\pm$0.53          & 67.62$\pm$0.12          & 86.79$\pm$0.09          & 86.28$\pm$0.18          & 85.53$\pm$0.30          & -              \\
  FedGTA~\cite{li2023fedgta}        & 80.06$\pm$0.63          & 80.59$\pm$0.38          & 79.01$\pm$0.31          & 70.12$\pm$0.10          & 71.57$\pm$0.34          & 69.94$\pm$0.14          & 87.75$\pm$0.01          & 86.80$\pm$0.01          & 87.12$\pm$0.05          & -              \\
  AdaFGL~\cite{li2024adafgl}        & 82.01$\pm$0.51          & 80.09$\pm$0.00          & 79.74$\pm$0.05          & 71.44$\pm$0.27          & 72.34$\pm$0.00          & 70.95$\pm$0.45          & 86.91$\pm$0.28          & 86.97$\pm$0.10          & 86.59$\pm$0.21          & -              \\ 
  FedTAD~\cite{zhu2024fedtad}       & 80.31$\pm$0.26          & 80.87$\pm$0.11          & 80.07$\pm$0.15          & 70.34$\pm$0.37          & 69.43$\pm$0.75          & 68.09$\pm$0.69          & 84.00$\pm$0.13          & 84.61$\pm$0.17          & 84.33$\pm$0.18          & -              \\ 
  FedIIH~\cite{wentao2025fediih}    & \underline{84.11$\pm$0.17}          & \underline{81.85$\pm$0.09}          & \underline{83.01$\pm$0.15}          & \underline{72.86$\pm$0.25}          & \underline{76.50$\pm$0.06}          & \textbf{73.36$\pm$0.41}          & \underline{87.80$\pm$0.18}          & \underline{87.65$\pm$0.18}          & \underline{87.19$\pm$0.25}         & -              \\ \hline
  FedGSP (Ours)                 & \textbf{84.72$\pm$0.06} & \textbf{83.00$\pm$0.10} & \textbf{83.66$\pm$0.08} & \textbf{73.60$\pm$0.11} & \textbf{77.35$\pm$0.09} & \underline{72.28$\pm$0.33} & \textbf{88.20$\pm$0.09} & \textbf{87.89$\pm$0.13} & \textbf{87.52$\pm$0.10} & -              \\ \hline
  \rowcolor{gray!50}
  & \multicolumn{3}{c}{Amazon-Computer}                                         & \multicolumn{3}{c}{Amazon-Photo}                                            & \multicolumn{3}{c}{ogbn-arxiv}                                              & Avg.            \\ \cline{2-11} 
  Methods       & 5 Clients               & 10 Clients              & 20 Clients              & 5 Clients               & 10 Clients              & 20 Clients              & 5 Clients               & 10 Clients              & 20 Clients              & All           \\ \hline
  Local         & 89.22$\pm$0.13          & 88.91$\pm$0.17          & 89.52$\pm$0.20          & 91.67$\pm$0.09          & 91.80$\pm$0.02          & 90.47$\pm$0.15          & 66.76$\pm$0.07          & 64.92$\pm$0.09          & 65.06$\pm$0.05          & 79.57          \\ \hline
  FedAvg~\cite{mcmahan2017communication}        & 84.88$\pm$1.96          & 79.54$\pm$0.23          & 74.79$\pm$0.24          & 89.89$\pm$0.83          & 83.15$\pm$3.71          & 81.35$\pm$1.04          & 65.54$\pm$0.07          & 64.44$\pm$0.10          & 63.24$\pm$0.13          & 74.58          \\
  FedProx~\cite{MLSYS2020_1f5fe839}       & 85.25$\pm$1.27          & 83.81$\pm$1.09          & 73.05$\pm$1.30          & 90.38$\pm$0.48          & 80.92$\pm$4.64          & 82.32$\pm$0.29          & 65.21$\pm$0.20          & 64.37$\pm$0.18          & 63.03$\pm$0.04          & 72.84          \\
  FedPer~\cite{Arivazhagan2019}        & 89.67$\pm$0.34          & 89.73$\pm$0.04          & 87.86$\pm$0.43          & 91.44$\pm$0.37          & 91.76$\pm$0.23          & 90.59$\pm$0.06          & 66.87$\pm$0.05          & 64.99$\pm$0.18          & 64.66$\pm$0.11          & 79.94          \\
  GCFL~\cite{NEURIPS2021_9c6947bd}          & 89.07$\pm$0.91          & 90.03$\pm$0.16          & 89.08$\pm$0.25          & 91.99$\pm$0.29          & 92.06$\pm$0.25          & 90.79$\pm$0.17          & 66.80$\pm$0.12          & 65.09$\pm$0.08          & 65.08$\pm$0.04          & 79.90          \\
  FedGNN~\cite{wu2021fedgnn}        & 88.08$\pm$0.15          & 88.18$\pm$0.41          & 83.16$\pm$0.13          & 90.25$\pm$0.70          & 87.12$\pm$2.01          & 81.00$\pm$4.48          & 65.47$\pm$0.22          & 64.21$\pm$0.32          & 63.80$\pm$0.05          & 75.23          \\
  FedSage+\cite{NEURIPS2021_34adeb8e}      & 85.04$\pm$0.61          & 80.50$\pm$1.13          & 70.42$\pm$0.85          & 90.77$\pm$0.44          & 76.81$\pm$8.24          & 80.58$\pm$1.15          & 65.69$\pm$0.09          & 64.52$\pm$0.14          & 63.31$\pm$0.20          & 73.47          \\
  FED-PUB~\cite{baek2023personalized}       & \underline{90.74$\pm$0.05} & 90.55$\pm$0.13          & 90.12$\pm$0.09          & 93.29$\pm$0.19          & 92.73$\pm$0.18          & 91.92$\pm$0.12          & 67.77$\pm$0.09          & 66.58$\pm$0.08          & 66.64$\pm$0.12          & 81.59          \\
  FedGTA~\cite{li2023fedgta}        & 86.69$\pm$0.18          & 86.66$\pm$0.23          & 85.01$\pm$0.87          & 93.33$\pm$0.12          & 93.50$\pm$0.21          & 92.61$\pm$0.15          & 60.32$\pm$0.04                   & 60.22$\pm$0.09                   & 58.74$\pm$0.14                         & 79.45          \\ 
  AdaFGL~\cite{li2024adafgl}        & 80.20$\pm$0.05          & 83.62$\pm$0.26          & 84.53$\pm$0.23          & 86.69$\pm$0.19          & 89.85$\pm$0.83          & 88.11$\pm$0.05          & 52.73$\pm$0.19          & 51.77$\pm$0.36          & 50.94$\pm$0.08          & 76.97 \\     
  FedTAD~\cite{zhu2024fedtad}       & 82.20$\pm$1.20          & 85.50$\pm$0.33          & 83.91$\pm$1.54          & 92.29$\pm$0.39          & 90.59$\pm$0.09          & 89.18$\pm$0.84          & 65.35$\pm$0.14          & 64.06$\pm$0.25          & 64.45$\pm$0.13          & 78.87 \\    
  FedIIH~\cite{wentao2025fediih} & \underline{90.74$\pm$0.13}          & \underline{90.86$\pm$0.23} & \textbf{90.44$\pm$0.05} & \underline{93.42$\pm$0.02} & \underline{94.22$\pm$0.08} & \underline{93.55$\pm$0.09} & \underline{70.30$\pm$0.06} & \underline{69.34$\pm$0.02} & \underline{68.65$\pm$0.04} & \underline{83.10} \\ \hline
  FedGSP (Ours)                  & \textbf{91.08$\pm$0.09}          & \textbf{91.08$\pm$0.12} & \underline{90.38$\pm$0.06} & \textbf{93.63$\pm$0.04} & \textbf{94.28$\pm$0.10} & \textbf{93.72$\pm$0.07} & \textbf{70.57$\pm$0.06} & \textbf{69.40$\pm$0.01} & \textbf{68.72$\pm$0.09} & \textbf{83.39} \\ \hline  
  \end{tabular}
  }
  \end{table*}

  \begin{table*}[]
      \centering
          \scriptsize
          \caption{Accuracy comparison of various methods on six \textbf{homophilic} graph datasets in the \textbf{overlapping} subgraph partitioning setting. The best and second-best records on each dataset are highlighted in \textbf{bold} and \underline{underlined}, respectively.}
          \label{table2}
    \renewcommand{\arraystretch}{0.8} 
           \scalebox{0.82}{
      \begin{tabular}{lcccccccccc}
      \hline
      \rowcolor{gray!50}
      \multicolumn{1}{c}{} & \multicolumn{3}{c}{Cora}                                                    & \multicolumn{3}{c}{CiteSeer}                                                      & \multicolumn{3}{c}{PubMed}                                                  & -              \\ \cline{2-11} 
      Methods              & 10 Clients              & 30 Clients              & 50 Clients              & 10 Clients                    & 30 Clients              & 50 Clients              & 10 Clients              & 30 Clients              & 50 Clients              & -              \\ \hline
      Local                & 73.98$\pm$0.25          & 71.65$\pm$0.12          & 76.63$\pm$0.10          & 65.12$\pm$0.08                & 64.54$\pm$0.42          & 66.68$\pm$0.44          & 82.32$\pm$0.07          & 80.72$\pm$0.16          & 80.54$\pm$0.11          & -              \\ \hline
      FedAvg~\cite{mcmahan2017communication}               & 76.48$\pm$0.36          & 53.99$\pm$0.98          & 53.99$\pm$4.53          & 69.48$\pm$0.15                & 66.15$\pm$0.64          & 66.51$\pm$1.00          & 82.67$\pm$0.11          & 82.05$\pm$0.12          & 80.24$\pm$0.35          & -              \\
      FedProx~\cite{MLSYS2020_1f5fe839}              & 77.85$\pm$0.50          & 51.38$\pm$1.74          & 56.27$\pm$9.04          & 69.39$\pm$0.35                & 66.11$\pm$0.75          & 66.53$\pm$0.43          & 82.63$\pm$0.17          & 82.13$\pm$0.13          & 80.50$\pm$0.46          & -              \\
      FedPer~\cite{Arivazhagan2019}               & 78.73$\pm$0.31          & 74.18$\pm$0.24          & 74.42$\pm$0.37          & 69.81$\pm$0.28                & 65.19$\pm$0.81          & 67.64$\pm$0.44          & 85.31$\pm$0.06          & 84.35$\pm$0.38          & 83.94$\pm$0.10          & -              \\
      GCFL~\cite{NEURIPS2021_9c6947bd}                 & 78.84$\pm$0.26          & 73.41$\pm$0.27          & 76.63$\pm$0.16          & 69.48$\pm$0.39                & 64.92$\pm$0.18          & 65.98$\pm$0.30          & 83.59$\pm$0.25          & 80.77$\pm$0.12          & 81.36$\pm$0.11          & -              \\
      FedGNN~\cite{wu2021fedgnn}               & 70.63$\pm$0.83          & 61.38$\pm$2.33          & 56.91$\pm$0.82          & 68.72$\pm$0.39                & 59.98$\pm$1.52          & 58.98$\pm$0.98          & 84.25$\pm$0.07          & 82.02$\pm$0.22          & 81.85$\pm$0.10          & -              \\
      FedSage+\cite{NEURIPS2021_34adeb8e}             & 77.52$\pm$0.46          & 51.99$\pm$0.42          & 55.48$\pm$11.5          & 68.75$\pm$0.48                & 65.97$\pm$0.02          & 65.93$\pm$0.30          & 82.77$\pm$0.08          & 82.14$\pm$0.11          & 80.31$\pm$0.68          & -              \\
      FED-PUB~\cite{baek2023personalized}              & 79.60$\pm$0.12          & 75.40$\pm$0.54          & \underline{77.84$\pm$0.23}          & 70.58$\pm$0.20                & 68.33$\pm$0.45          & 69.21$\pm$0.30          & 85.70$\pm$0.08          & 85.16$\pm$0.10          & 84.84$\pm$0.12          & -              \\
      FedGTA~\cite{li2023fedgta}               & 76.42$\pm$0.62          & 75.63$\pm$0.33          & 77.69$\pm$0.14          & 70.43$\pm$0.08 & 71.71$\pm$0.33          & 69.19$\pm$0.32          & 85.34$\pm$0.42          & 84.99$\pm$0.05          & 84.47$\pm$0.06          & -              \\
      AdaFGL~\cite{li2024adafgl}               & 78.50$\pm$0.19          & 75.80$\pm$0.23          & 74.41$\pm$0.00          & 72.63$\pm$0.15 & 68.18$\pm$0.31          & 62.90$\pm$0.75          & 85.58$\pm$0.23          & 85.85$\pm$0.41          & 84.45$\pm$0.07          & -              \\
	 FedTAD~\cite{zhu2024fedtad}              & 79.29$\pm$0.78          & 60.92$\pm$2.17          & 68.08$\pm$0.44          & \textbf{73.47$\pm$0.16}            & 67.74$\pm$0.57          & 63.51$\pm$0.68          & 82.98$\pm$0.20          & 82.11$\pm$0.15          & 81.63$\pm$0.19          & -              \\ 
      FedIIH~\cite{wentao2025fediih}        & \underline{80.57$\pm$0.23}  & \underline{76.82$\pm$0.24} & \textbf{78.58$\pm$0.25} & \underline{73.16$\pm$0.18}       & \underline{72.27$\pm$0.21}    & \underline{69.56$\pm$0.11} & \underline{85.87$\pm$0.03} & \underline{86.65$\pm$0.11} & \textbf{85.65$\pm$0.12} & -              \\ \hline
      FedGSP (Ours)                         & \textbf{80.99$\pm$0.15}    & \textbf{76.86$\pm$0.12} & 77.82$\pm$0.09          & 72.28$\pm$0.05                & \textbf{72.55$\pm$0.13}    & \textbf{69.79$\pm$0.13} & \textbf{87.21$\pm$0.06} & \textbf{86.92$\pm$0.15} & \underline{85.31$\pm$0.15} & -              \\ \hline
      \rowcolor{gray!50}
                           & \multicolumn{3}{c}{Amazon-Computer}                                         & \multicolumn{3}{c}{Amazon-Photo}                                                  & \multicolumn{3}{c}{ogbn-arxiv}                                              & Avg.            \\ \cline{2-11} 
      Methods              & 10 Clients              & 30 Clients              & 50 Clients              & 10 Clients                    & 30 Clients              & 50 Clients              & 10 Clients              & 30 Clients              & 50 Clients              & All              \\ \hline
      Local                & 88.50$\pm$0.20          & 86.66$\pm$0.00          & 87.04$\pm$0.02          & 92.17$\pm$0.12                & 90.16$\pm$0.12          & 90.42$\pm$0.15          & 62.52$\pm$0.07          & 61.32$\pm$0.04          & 60.04$\pm$0.04          & 76.72          \\ \hline
      FedAvg~\cite{mcmahan2017communication}               & 88.99$\pm$0.19          & 83.37$\pm$0.47          & 76.34$\pm$0.12          & 92.91$\pm$0.07                & 89.30$\pm$0.22          & 74.19$\pm$0.57          & 63.56$\pm$0.02          & 59.72$\pm$0.06          & 60.94$\pm$0.24          & 73.38          \\
      FedProx~\cite{MLSYS2020_1f5fe839}              & 88.84$\pm$0.20          & 83.84$\pm$0.89          & 76.60$\pm$0.47          & 92.67$\pm$0.19                & 89.17$\pm$0.40          & 72.36$\pm$2.06          & 63.52$\pm$0.11          & 59.86$\pm$0.16          & 61.12$\pm$0.04          & 73.38          \\
      FedPer~\cite{Arivazhagan2019}               & 89.30$\pm$0.04          & 87.99$\pm$0.23          & 88.22$\pm$0.27          & 92.88$\pm$0.24                & 91.23$\pm$0.16          & 90.92$\pm$0.38          & 63.97$\pm$0.08          & 62.29$\pm$0.04          & 61.24$\pm$0.11          & 78.42          \\
      GCFL~\cite{NEURIPS2021_9c6947bd}                 & 89.01$\pm$0.22          & 87.24$\pm$0.09          & 87.02$\pm$0.22          & 92.45$\pm$0.10                & 90.58$\pm$0.11          & 90.54$\pm$0.08          & 63.24$\pm$0.02          & 61.66$\pm$0.10          & 60.32$\pm$0.01          & 77.61          \\
      FedGNN~\cite{wu2021fedgnn}               & 88.15$\pm$0.09          & 87.00$\pm$0.10          & 83.96$\pm$0.88          & 91.47$\pm$0.11                & 87.91$\pm$1.34          & 78.90$\pm$6.46          & 63.08$\pm$0.19          & 60.09$\pm$0.04          & 60.51$\pm$0.11          & 73.66          \\
      FedSage+\cite{NEURIPS2021_34adeb8e}             & 89.24$\pm$0.15          & 81.33$\pm$1.20          & 76.72$\pm$0.39          & 92.76$\pm$0.05                & 88.69$\pm$0.99          & 72.41$\pm$1.36          & 63.24$\pm$0.02          & 59.90$\pm$0.12          & 60.95$\pm$0.09          & 73.12          \\
      FED-PUB~\cite{baek2023personalized}              & 89.98$\pm$0.08          & 89.15$\pm$0.06          & 88.76$\pm$0.14          & 93.22$\pm$0.07                & 92.01$\pm$0.07          & 91.71$\pm$0.11          & 64.18$\pm$0.04          & 63.34$\pm$0.12          & 62.55$\pm$0.12          & 79.53          \\
      FedGTA~\cite{li2023fedgta}               & 90.10$\pm$0.18          & 88.79$\pm$0.27          & 88.15$\pm$0.21          & 93.13$\pm$0.14                & 92.49$\pm$0.06          & 91.77$\pm$0.06          & 55.98$\pm$0.09          & 56.76$\pm$0.07          & 57.89$\pm$0.09          & 74.40          \\
      AdaFGL~\cite{li2024adafgl}        & 80.49$\pm$0.00          & 80.42$\pm$0.00          & 82.12$\pm$0.00          & 89.24$\pm$0.00          & 88.34$\pm$0.00          & 87.68$\pm$0.00          & 56.81$\pm$0.06                   & 55.17$\pm$0.00                   & 54.82$\pm$0.00                         & 75.74          \\ 
      FedTAD~\cite{zhu2024fedtad}        & 79.09$\pm$5.63          & 79.48$\pm$0.85          & 77.05$\pm$0.07          & 81.94$\pm$3.09          & 86.58$\pm$1.75          & 84.38$\pm$1.33          & 58.45$\pm$0.15                   & 57.75$\pm$0.54                   & 56.52$\pm$0.14                         & 73.39          \\
      FedIIH~\cite{wentao2025fediih}        & \underline{90.15$\pm$0.04} & \underline{89.56$\pm$0.19}                 & \textbf{89.99$\pm$0.00} & \underline{93.38$\pm$0.00}                & \underline{94.17$\pm$0.04} & \underline{93.25$\pm$0.16} & \underline{66.69$\pm$0.09}          & \underline{66.10$\pm$0.03}          & \textbf{65.67$\pm$0.06} & \underline{81.01} \\ \hline
      FedGSP (Ours)                       & \textbf{90.27$\pm$0.06} & \textbf{89.67$\pm$0.11} & \underline{89.31$\pm$0.00} & \textbf{93.69$\pm$0.03}       & \textbf{94.33$\pm$0.05} & \textbf{93.28$\pm$0.07} & \textbf{67.34$\pm$0.04} & \textbf{66.18$\pm$0.02} & \underline{65.64$\pm$0.08} & \textbf{81.08} \\ \hline
      \end{tabular}
      }
  \end{table*}

\subsubsection{Heterophilic Datasets}
Tab.~\ref{table3} and Tab.~\ref{table4} show the comparison results on the heterophilic datasets in two partitioning settings, respectively. Our proposed FedGSP not only obtains the best performance among most of the baseline methods, but also outperforms the second-best method (\textit{i.e.}, FedIIH~\cite{wentao2025fediih}) by an average margin of 3.28\% and 1.61\% in the non-overlapping and overlapping scenarios, respectively. Furthermore, since the heterogeneity of heterophilic graph datasets is generally stronger than that of the homophilic graph datasets~\cite{li2024adafgl,wentao2025fediih}, heterophilic graph datasets are usually more challenging than homophilic graph datasets. However, our proposed FedGSP still outperforms the second-best method (\textit{i.e.}, FedIIH) in both kinds of graph datasets and achieves a larger performance gain on the heterophilic datasets than on the homophilic datasets, which validates the effectiveness of mining graph spectral properties.
 
\begin{table*}[]
\centering
\scriptsize
\caption{Comparison of various methods on five \textbf{heterophilic} graph datasets in the \textbf{non-overlapping} subgraph partitioning setting. Accuracy (\%) is reported for \textit{Roman-empire} and \textit{Amazon-ratings}, and AUC (\%) is reported for \textit{Minesweeper}, \textit{Tolokers}, and \textit{Questions}. The best and second-best records on each dataset are highlighted in \textbf{bold} and \underline{underlined}, respectively.}
\label{table3}
\renewcommand{\arraystretch}{0.8} 
  \scalebox{0.82}{
  \begin{tabular}{lcccccccccc}
  \hline
  \rowcolor{gray!50}
  \textbf{}     & \multicolumn{3}{c}{Roman-empire}                                            & \multicolumn{3}{c}{Amazon-ratings}                                          & \multicolumn{3}{c}{Minesweeper}                                             & -              \\ \cline{2-11} 
  Methods       & 5 Clients               & 10 Clients              & 20 Clients              & 5 Clients               & 10 Clients              & 20 Clients              & 5 Clients               & 10 Clients              & 20 Clients              & -              \\ \hline
  Local         & 33.65$\pm$0.13          & 28.42$\pm$0.26          & 23.89$\pm$0.32          & \underline{45.03$\pm$0.31} & \textbf{45.89$\pm$0.19} & \underline{46.02$\pm$0.02} & 71.35$\pm$0.17          & 69.96$\pm$0.16          & 69.31$\pm$0.09          & -              \\ \hline
  FedAvg~\cite{mcmahan2017communication}        & 38.93$\pm$0.32          & 35.43$\pm$0.32          & 32.00$\pm$0.39          & 41.26$\pm$0.53          & 41.66$\pm$0.14          & 42.20$\pm$0.21          & 72.60$\pm$0.08          & 71.84$\pm$0.02          & 71.36$\pm$0.16          & -              \\
  FedProx~\cite{MLSYS2020_1f5fe839}       & 27.95$\pm$0.59          & 26.43$\pm$1.41          & 23.12$\pm$0.49          & 36.92$\pm$0.00          & 36.86$\pm$0.14          & 36.96$\pm$0.05          & 71.91$\pm$0.27          & 70.66$\pm$0.20          & 71.50$\pm$0.37          & -              \\
  FedPer~\cite{Arivazhagan2019}        & 20.75$\pm$1.75          & 15.51$\pm$1.13          & 15.45$\pm$2.76          & 36.62$\pm$0.30          & 32.34$\pm$1.01          & 36.96$\pm$0.03          & 58.73$\pm$10.45         & 65.35$\pm$7.02          & 53.80$\pm$11.40         & -              \\
  GCFL~\cite{NEURIPS2021_9c6947bd}          & 30.40$\pm$0.16          & 29.44$\pm$0.49          & 26.73$\pm$0.19          & 36.92$\pm$0.00          & 36.86$\pm$0.14          & 36.96$\pm$0.02          & 72.04$\pm$0.13          & 71.14$\pm$0.09          & 47.77$\pm$0.14          & -              \\
  FedGNN~\cite{wu2021fedgnn}        & 30.26$\pm$0.11          & 29.09$\pm$0.01          & 26.60$\pm$0.02          & 36.80$\pm$0.06          & 36.72$\pm$0.00          & 36.45$\pm$0.09          & 72.15$\pm$0.13          & 71.08$\pm$0.07          &  71.71$\pm$0.27    & -              \\
  FedSage+\cite{NEURIPS2021_34adeb8e}      & 57.26$\pm$0.00          & 49.07$\pm$0.00          & 38.36$\pm$0.00          & 36.82$\pm$0.00          & 36.71$\pm$0.00          & 37.03$\pm$0.02          & \underline{77.74$\pm$0.00} & 72.80$\pm$0.00    & 69.70$\pm$0.00          & -              \\
  FED-PUB~\cite{baek2023personalized}       & 40.80$\pm$0.26          & 36.77$\pm$0.30          & 32.67$\pm$0.39          & 44.41$\pm$0.41    &  44.85$\pm$0.17    & 45.39$\pm$0.50    & 72.18$\pm$0.02          & 71.56$\pm$0.05          & 70.72$\pm$0.40          & -              \\
  FedGTA~\cite{li2023fedgta}        & 61.56$\pm$0.27    & 60.94$\pm$0.19    & 59.65$\pm$0.28    & 41.22$\pm$0.66          & 39.40$\pm$0.44          & 39.24$\pm$0.12          & 45.60$\pm$1.41          & 64.97$\pm$0.35          & 49.63$\pm$8.64          & -              \\
  AdaFGL~\cite{li2024adafgl}        & 67.64$\pm$0.18          & 64.55$\pm$0.00          & 62.42$\pm$0.26          & 41.70$\pm$0.06          & 42.30$\pm$0.00          & 42.59$\pm$0.14          & 47.45$\pm$2.10                  & 65.59$\pm$0.56                   & 51.48$\pm$7.14                         & -          \\ 
  FedTAD~\cite{zhu2024fedtad} & 45.26$\pm$0.19          & 44.71$\pm$0.38          & 42.04$\pm$0.13          & 43.59$\pm$0.33          & 43.35$\pm$0.29          & 44.50$\pm$0.26          & 69.52$\pm$0.06                  & 70.74$\pm$0.09                   & 72.74$\pm$0.03                         & -          \\ 
  FedIIH~\cite{wentao2025fediih} & \underline{68.32$\pm$0.05} & \underline{66.44$\pm$0.28} & \underline{64.61$\pm$0.13} & 44.26$\pm$0.24          & 44.24$\pm$0.10          & 45.19$\pm$0.04          & 74.29$\pm$0.02    & \underline{73.23$\pm$0.04} & \underline{72.81$\pm$0.02} &   -             \\ \hline
FedGSP (Ours) & \textbf{68.80$\pm$0.04} & \textbf{67.72$\pm$0.09} & \textbf{66.02$\pm$0.07} & \textbf{45.96$\pm$0.06}          & \underline{45.35$\pm$0.06}          & \textbf{46.04$\pm$0.05}          & \textbf{85.56$\pm$0.04}    & \textbf{85.28$\pm$0.06} & \textbf{84.22$\pm$0.06} & -               \\ \hline
\rowcolor{gray!50}
                & \multicolumn{3}{c}{Tolokers}                                                & \multicolumn{3}{c}{Questions}                                               & \multicolumn{4}{c}{Avg.}                                                                     \\ \cline{2-11} 
  Methods       & 5 Clients               & 10 Clients              & 20 Clients              & 5 Clients               & 10 Clients              & 20 Clients              & 5 Clients               & 10 Clients              & 20 Clients              & All            \\ \hline
  Local         & 67.81$\pm$0.17          & 70.04$\pm$0.23          & 62.34$\pm$0.67          & 66.73$\pm$0.57          & 57.96$\pm$0.10          & 60.00$\pm$0.21          & 56.91                   & 54.45                   & 52.31                   & 54.56          \\ \hline
  FedAvg~\cite{mcmahan2017communication}        & 60.74$\pm$0.31          & 54.73$\pm$0.50          & 56.36$\pm$0.39          & 65.68$\pm$0.23          & 58.91$\pm$0.22          & 60.33$\pm$0.15          & 55.84                   & 52.51                   & 52.45                   & 53.60          \\
  FedProx~\cite{MLSYS2020_1f5fe839}       & 42.90$\pm$0.24          & 41.15$\pm$0.22          & 40.42$\pm$0.62          & 47.36$\pm$0.38          & 45.46$\pm$0.34          & 46.83$\pm$0.11          & 45.41                   & 44.11                   & 43.77                   & 44.43          \\
  FedPer~\cite{Arivazhagan2019}        & 46.61$\pm$9.88          & 54.97$\pm$13.23         & 44.82$\pm$11.61         & 58.38$\pm$9.39          & 59.40$\pm$9.71          & 62.32$\pm$1.56          & 44.22                   & 45.51                   & 42.67                   & 44.13          \\
  GCFL~\cite{NEURIPS2021_9c6947bd}          & 35.54$\pm$1.42          & 38.86$\pm$0.65          & 36.69$\pm$0.57          & 47.94$\pm$0.41          & 45.71$\pm$0.25          & 47.47$\pm$0.21          & 42.98                   & 40.59                   & 35.29                   & 39.62          \\
  FedGNN~\cite{wu2021fedgnn}        & 43.10$\pm$0.27          & 41.57$\pm$0.07          & 40.70$\pm$0.74          & 47.55$\pm$0.02          & 45.65$\pm$0.12          & 47.39$\pm$0.13          & 45.99                   & 44.82                   & 44.57                   & 45.13          \\
  FedSage+\cite{NEURIPS2021_34adeb8e}      & \textbf{75.06$\pm$0.00} & 71.31$\pm$0.00          &  69.73$\pm$0.00    & 64.95$\pm$0.00          &  65.06$\pm$0.00    & 59.33$\pm$0.00          & 62.37            &  58.99             & 54.83                   &  58.73    \\
  FED-PUB~\cite{baek2023personalized}       & 70.88$\pm$0.58          & \underline{72.46$\pm$0.68} & 65.26$\pm$0.59          &  67.71$\pm$3.99    & 54.91$\pm$0.42          & 62.48$\pm$2.92    & 59.20                   & 56.11                   &  55.30             & 56.87          \\
  FedGTA~\cite{li2023fedgta}        & 33.33$\pm$0.51          & 49.97$\pm$2.68          & 50.68$\pm$3.94          & 53.61$\pm$0.36          & 53.79$\pm$0.41          & 61.70$\pm$0.35          & 47.06                   & 53.81                   & 52.18                   & 51.02          \\
  AdaFGL~\cite{li2024adafgl}        & 34.41$\pm$0.63          & 49.82$\pm$2.17          & 50.62$\pm$4.19          & 54.18$\pm$0.45          & 54.87$\pm$0.52          & 62.84$\pm$0.49          & 49.08                   & 55.43                   & 53.99                         & 52.83          \\ 
FedTAD~\cite{zhu2024fedtad} & 60.91$\pm$0.25          & 53.39$\pm$1.73          & 56.47$\pm$1.58          & 58.76$\pm$0.17          & 58.22$\pm$0.11          & 57.46$\pm$0.24          & 55.61                  & 54.08                   & 54.64                         & 54.78          \\
  FedIIH~\cite{wentao2025fediih} &  71.09$\pm$0.26    & 71.32$\pm$0.09    & \underline{70.30$\pm$0.10} & \underline{68.32$\pm$0.03} & \underline{67.99$\pm$0.09} & \underline{65.40$\pm$0.07} & \underline{65.26}          & \underline{64.64}          & \underline{63.66}          & \underline{64.52} \\ \hline
FedGSP (Ours) & \underline{73.98$\pm$0.07} & \textbf{73.72$\pm$0.10} & \textbf{70.78$\pm$0.07} & \textbf{69.42$\pm$0.11}          & \textbf{68.52$\pm$0.13}          & \textbf{65.63$\pm$0.04}          & \textbf{68.74}    & \textbf{68.12} & \textbf{66.54} & \textbf{67.80}               \\ \hline
      \end{tabular}
       }
      \end{table*}

\begin{table*}[]
      \centering
      \scriptsize
      \caption{Comparison of various methods on five \textbf{heterophilic} graph datasets in the \textbf{overlapping} subgraph partitioning setting. Accuracy (\%) is reported for \textit{Roman-empire} and \textit{Amazon-ratings}, and AUC (\%) is reported for \textit{Minesweeper}, \textit{Tolokers}, and \textit{Questions}. The best and second-best records on each dataset are highlighted in \textbf{bold} and \underline{underlined}, respectively.}
      \label{table4}
    \renewcommand{\arraystretch}{0.82} 
       \scalebox{0.83}{
  \begin{tabular}{lcccccccccc}
  \hline
  \rowcolor{gray!50}
  \textbf{}     & \multicolumn{3}{c}{Roman-empire}                                            & \multicolumn{3}{c}{Amazon-ratings}                                          & \multicolumn{3}{c}{Minesweeper}                                                   & -              \\ \cline{2-11} 
  Methods       & 10 Clients              & 30 Clients              & 50 Clients              & 10 Clients              & 30 Clients              & 50 Clients              & 10 Clients              & 30 Clients              & 50 Clients                    & -              \\ \hline
  Local         & 39.47$\pm$0.03          & 34.43$\pm$0.14          & 31.28$\pm$0.18          & 41.43$\pm$0.04          & 41.81$\pm$0.14          & 42.57$\pm$0.12          & 67.98$\pm$0.07          & 64.39$\pm$0.10          & 62.73$\pm$0.23                & -              \\ \hline
  FedAvg~\cite{mcmahan2017communication}        & 40.89$\pm$0.25          & 38.66$\pm$0.08          & 36.71$\pm$0.20          & 39.86$\pm$0.06          & 41.40$\pm$0.02          & 41.02$\pm$0.16          & 69.06$\pm$0.07          & 67.95$\pm$0.04          & 66.89$\pm$0.08                & -              \\
  FedProx~\cite{MLSYS2020_1f5fe839}       & 36.63$\pm$0.14          & 35.31$\pm$0.17          & 33.61$\pm$0.59          & 37.00$\pm$0.00          & 36.60$\pm$0.00          & 36.89$\pm$0.00          & 68.27$\pm$0.05          & 66.75$\pm$0.19          & 66.03$\pm$0.16                & -              \\
  FedPer~\cite{Arivazhagan2019}        & 23.66$\pm$3.27          & 23.27$\pm$3.09          & 22.23$\pm$3.58          & 32.33$\pm$4.23          & 31.58$\pm$0.54          & 34.48$\pm$2.25          & 61.85$\pm$1.02          & 60.13$\pm$1.38          & 60.06$\pm$3.61                & -              \\
  GCFL~\cite{NEURIPS2021_9c6947bd}          & 37.65$\pm$0.27          & 36.32$\pm$0.19          & 34.80$\pm$0.09          & 37.00$\pm$0.00          & 36.60$\pm$0.00          & 36.89$\pm$0.00          & 68.47$\pm$0.06          & 67.13$\pm$0.10          & 57.41$\pm$12.56               & -              \\
  FedGNN~\cite{wu2021fedgnn}        & 37.46$\pm$0.12          & 36.47$\pm$0.24          & 34.92$\pm$0.26          & 36.58$\pm$0.16          & 36.77$\pm$0.12          & 36.95$\pm$0.15          & 68.59$\pm$0.21          & 67.30$\pm$0.17          & 66.41$\pm$0.23                & -              \\
  FedSage+\cite{NEURIPS2021_34adeb8e}      & 57.48$\pm$0.00          & 42.55$\pm$0.00          & 33.99$\pm$0.00          & 36.86$\pm$0.00          & 36.71$\pm$0.00          & 37.03$\pm$0.00          & \textbf{76.64$\pm$0.00} & \underline{70.56$\pm$0.00} &  \underline{70.34$\pm$0.00} & -              \\
  FED-PUB~\cite{baek2023personalized}       & 43.80$\pm$0.25          & 40.46$\pm$0.16          & 37.73$\pm$0.09          & 42.25$\pm$0.25    &  42.25$\pm$0.06    & \underline{42.88$\pm$0.34} & 68.80$\pm$0.09          & 67.43$\pm$0.25          & 65.98$\pm$0.15                & -              \\
  FedGTA~\cite{li2023fedgta}        & 59.86$\pm$0.04    & 58.32$\pm$0.09    & 57.57$\pm$0.21    & 40.81$\pm$0.24          & 39.44$\pm$0.06          & 39.37$\pm$0.04          & 54.35$\pm$0.73          & 48.20$\pm$1.28          & 52.94$\pm$1.77                & -              \\
  AdaFGL~\cite{li2024adafgl}        & 64.44$\pm$0.03          & 61.77$\pm$0.02          & 59.55$\pm$0.01          & 39.39$\pm$0.05          & 41.19$\pm$0.15          & 40.71$\pm$0.25          & 55.15$\pm$0.84                   & 50.15$\pm$1.63                   & 54.18$\pm$2.15                         & -          \\
  FedTAD~\cite{zhu2024fedtad}        & 44.14$\pm$0.13          & 41.94$\pm$0.18          & 40.82$\pm$0.01          & 39.53$\pm$0.17          & 40.69$\pm$0.13          & 40.58$\pm$0.26          & 68.69$\pm$0.08                   & 68.43$\pm$0.05                   & 66.66$\pm$0.05                         & -          \\
  FedIIH~\cite{wentao2025fediih} & \underline{65.48$\pm$0.12} & \underline{63.32$\pm$0.06} & \underline{62.42$\pm$0.10}          & \underline{42.63$\pm$0.02}           & \underline{42.40$\pm$0.05}          & 42.65$\pm$0.21              & 69.35$\pm$0.25    &  68.09$\pm$0.26    &  67.37$\pm$0.14 & -         \\ \hline
 FedGSP (Ours) & \textbf{65.81$\pm$0.13} & \textbf{63.96$\pm$0.19} & \textbf{62.92$\pm$0.07} & \textbf{42.79$\pm$0.04} & \textbf{42.60$\pm$0.06} & \textbf{42.98$\pm$0.12}    & \underline{ 75.70$\pm$0.17}    & \textbf{ 75.08$\pm$0.15}    & \textbf{ 71.52$\pm$0.09} & -              \\ \hline
 \rowcolor{gray!50}
                & \multicolumn{3}{c}{Tolokers}                                                & \multicolumn{3}{c}{Questions}                                               & \multicolumn{4}{c}{Avg.}                                                                           \\ \cline{2-11} 
  Methods       & 10 Clients              & 30 Clients              & 50 Clients              & 10 Clients              & 30 Clients              & 50 Clients              & 10 Clients              & 30 Clients              & 50 Clients                    & All            \\ \hline
  Local         & 73.83$\pm$0.03          & 69.01$\pm$0.31          & 66.63$\pm$0.20          & 63.17$\pm$0.02          & 57.17$\pm$0.08          & 56.13$\pm$0.02          & 57.18                   & 53.36                   & 51.87                         & 54.14          \\ \hline
  FedAvg~\cite{mcmahan2017communication}        & 72.99$\pm$0.40          & 58.51$\pm$0.27          & 55.47$\pm$0.42          & 62.80$\pm$0.63          & 58.88$\pm$0.18          & 60.78$\pm$0.27          & 57.12                   & 53.08                   & 52.17                         & 54.12          \\
  FedProx~\cite{MLSYS2020_1f5fe839}       & 54.49$\pm$1.69          & 45.59$\pm$0.41          & 41.49$\pm$0.45          & 52.53$\pm$0.34          & 51.54$\pm$0.41          & 50.72$\pm$0.40          & 49.78                   & 47.16                   & 45.75                         & 47.56          \\
  FedPer~\cite{Arivazhagan2019}        & 39.60$\pm$0.11          & 59.44$\pm$0.79          & 41.92$\pm$0.06          & 61.31$\pm$0.29          & 53.41$\pm$1.53          & 50.29$\pm$0.10          & 43.75                   & 45.57                   & 41.80                         & 43.70          \\
  GCFL~\cite{NEURIPS2021_9c6947bd}          & 55.91$\pm$1.13          & 47.91$\pm$0.59          & 18.40$\pm$0.25          & 53.04$\pm$0.47          & 51.84$\pm$0.38          & 51.10$\pm$0.38          & 50.41                   & 43.03                   & 39.72                         & 44.39          \\
  FedGNN~\cite{wu2021fedgnn}        & 56.21$\pm$1.20          & 46.85$\pm$0.31          & 42.18$\pm$0.45          & 53.25$\pm$0.15          & 51.90$\pm$0.15          & 51.22$\pm$0.14          & 50.42                   & 47.86                   & 46.34                         & 48.97          \\
  FedSage+\cite{NEURIPS2021_34adeb8e}      & \textbf{74.54$\pm$0.00} &  70.88$\pm$0.00    &  69.61$\pm$0.00    & 64.22$\pm$0.00          &  65.34$\pm$0.00    &  62.76$\pm$0.00    & 61.95             & 57.21             &  54.75              & 57.97    \\
  FED-PUB~\cite{baek2023personalized}       & 74.17$\pm$0.29    & 70.35$\pm$0.54          & 66.80$\pm$0.85          &  65.39$\pm$2.44    & 58.38$\pm$1.19          & 58.76$\pm$0.16          & 58.88                   & 55.77                   & 54.43                         & 56.36          \\
  FedGTA~\cite{li2023fedgta}        & 40.02$\pm$1.70          & 47.34$\pm$0.75          & 45.81$\pm$1.96          & 35.56$\pm$5.46          & 50.43$\pm$1.05          & 53.33$\pm$0.40          & 46.12                   & 48.75                   & 49.80                         & 48.22          \\
  AdaFGL~\cite{li2024adafgl}        & 45.15$\pm$2.15          & 49.18$\pm$0.84          & 47.54$\pm$2.48          & 41.05$\pm$6.49          & 52.18$\pm$2.16          & 56.46$\pm$0.92          & 49.04                   & 50.89                   & 51.69                         & 50.54          \\
  FedTAD~\cite{zhu2024fedtad}        & 69.27$\pm$1.26          & 62.11$\pm$0.27          & 56.39$\pm$0.52          & 59.28$\pm$0.28         & 59.24$\pm$0.36          & 57.81$\pm$0.24          & 56.18                   & 54.48                   & 52.45                         & 54.37          \\ 
  FedIIH~\cite{wentao2025fediih}     & 71.67$\pm$0.02          & \underline{71.69$\pm$0.12}          & \underline{69.99$\pm$0.03}      & \underline{68.79$\pm$0.09}   & \textbf{66.98$\pm$0.04} & \underline{64.73$\pm$0.35}          & \underline{63.58}          & \underline{62.50}          & \underline{61.43}                & \underline{62.50} \\ \hline
  FedGSP (Ours)        & \underline{74.41$\pm$0.03}          & \textbf{72.21$\pm$0.12} & \textbf{70.52$\pm$0.09}         & \textbf{69.38$\pm$0.07}    & \underline{66.27$\pm$0.03}  & \textbf{65.54$\pm$0.14} & \textbf{65.62}          & \textbf{64.02}          & \textbf{62.70}                & \textbf{64.11} \\ \hline
      \end{tabular}
       }
\end{table*}

\subsection{Ablation Study}
Since our proposed FedGSP not only shares generic spectral properties but also complements non-generic spectral properties, we carry out the ablation experiments to shed light on the contributions of these two strategies. Specifically, we employ the `Sharing' and `Complementing' to represent sharing generic spectral properties and complementing non-generic spectral properties, respectively. Depending on whether these two strategies are used, there are a total of four combinations, as shown in Tab.~\ref{table5}. We can clearly observe that the performance decreases when any strategy is removed, indicating that each strategy contributes a lot to the final performance. For example, the accuracies on the \textit{Cora} dataset are obviously reduced when both two strategies are removed.

\begin{table*}[]
  \centering
  \scriptsize
  \caption{Ablation studies in both non-overlapping and overlapping partitioning settings on the \textit{Cora} and \textit{Roman-empire} datasets.}
  \label{table5}
  \renewcommand{\arraystretch}{0.82} 
  \scalebox{0.83}{
  \begin{tabular}{cccccccc}
  \hline
  \rowcolor{gray!50}
           &               & \multicolumn{6}{c}{Cora}                                                                                                                                                                                                                                                                                                                                                                                                       \\ \hline
  Sharing  & Complementing & \begin{tabular}[c]{@{}c@{}}non-overlapping\\ 5 clients\end{tabular} & \begin{tabular}[c]{@{}c@{}}non-overlapping\\ 10 clients\end{tabular} & \begin{tabular}[c]{@{}c@{}}non-overlapping\\ 20 clients\end{tabular} & \begin{tabular}[c]{@{}c@{}}overlapping\\ 10 clients\end{tabular} & \begin{tabular}[c]{@{}c@{}}overlapping\\ 30 clients\end{tabular} & \begin{tabular}[c]{@{}c@{}}overlapping\\ 50 clients\end{tabular} \\ \hline
  \textcolor{teal}{\CheckmarkBold}  & \textcolor{red}{\XSolidBrush}      & 82.52$\pm$0.21 ($\downarrow$ 2.20)                                  & 80.26$\pm$0.15 ($\downarrow$ 2.74)                                   & 81.55$\pm$0.22 ($\downarrow$ 2.11)                                   & 78.84$\pm$0.21 ($\downarrow$ 2.15)                               & 74.51$\pm$0.14 ($\downarrow$ 2.35)                                   & 75.76$\pm$0.16 ($\downarrow$ 2.06)                               \\ \hline
  \textcolor{red}{\XSolidBrush} & \textcolor{teal}{\CheckmarkBold}       & 71.52$\pm$0.43 ($\downarrow$ 13.20)                                 & 74.14$\pm$0.21 ($\downarrow$ 8.86)                                   & 78.32$\pm$0.16 ($\downarrow$ 5.34)                                   & 77.32$\pm$0.19 ($\downarrow$ 3.67)                               & 73.98$\pm$0.33 ($\downarrow$ 2.88)                                   & 75.00$\pm$0.22 ($\downarrow$ 2.82)                               \\ \hline
  \textcolor{red}{\XSolidBrush} & \textcolor{red}{\XSolidBrush}      & 64.61$\pm$0.34 ($\downarrow$ 20.11)                                 & 76.38$\pm$0.23 ($\downarrow$ 6.62)                                   & 77.30$\pm$0.18 ($\downarrow$ 6.36)                                   & 78.32$\pm$0.35 ($\downarrow$ 2.67)                               & 74.72$\pm$0.31 ($\downarrow$ 2.14)                                   & 71.49$\pm$0.18 ($\downarrow$ 6.33)                               \\ \hline
  \textcolor{teal}{\CheckmarkBold}  & \textcolor{teal}{\CheckmarkBold}       & \textbf{84.72$\pm$0.06}                                             & \textbf{83.00$\pm$0.10}                                              & \textbf{83.66$\pm$0.08}                                              & \textbf{80.99$\pm$0.15}                                          & \textbf{76.86$\pm$0.12}                                              & \textbf{77.82$\pm$0.09}                                          \\ \hline
  \rowcolor{gray!50}
           &               & \multicolumn{6}{c}{Roman-empire}                                                                                                                                                                                                                                                                                                                                                                                               \\ \hline
  Sharing  & Complementing & \begin{tabular}[c]{@{}c@{}}non-overlapping\\ 5 clients\end{tabular} & \begin{tabular}[c]{@{}c@{}}non-overlapping\\ 10 clients\end{tabular} & \begin{tabular}[c]{@{}c@{}}non-overlapping\\ 20 clients\end{tabular} & \begin{tabular}[c]{@{}c@{}}overlapping\\ 10 clients\end{tabular} & \begin{tabular}[c]{@{}c@{}}overlapping\\ 30 clients\end{tabular} & \begin{tabular}[c]{@{}c@{}}overlapping\\ 50 clients\end{tabular} \\ \hline
  \textcolor{teal}{\CheckmarkBold}  & \textcolor{red}{\XSolidBrush}      & 65.40$\pm$0.18 ($\downarrow$ 3.40)                                  & 65.57$\pm$0.16 ($\downarrow$ 2.15)                                   & 62.70$\pm$0.12 ($\downarrow$ 2.72)                                   & 62.84$\pm$0.17 ($\downarrow$ 2.97)                               & 62.33$\pm$0.24 ($\downarrow$ 1.63)                                   & 60.89$\pm$0.16 ($\downarrow$ 2.03)                               \\ \hline
  \textcolor{red}{\XSolidBrush} & \textcolor{teal}{\CheckmarkBold}       & 58.69$\pm$0.41 ($\downarrow$ 10.11)                                 & 65.46$\pm$0.18 ($\downarrow$ 2.26)                                   & 63.14$\pm$0.18 ($\downarrow$ 2.28)                                   & 62.27$\pm$0.15 ($\downarrow$ 3.54)                               & 61.00$\pm$0.21 ($\downarrow$ 2.96)                                   & 60.30$\pm$0.15 ($\downarrow$ 2.62)                               \\ \hline
  \textcolor{red}{\XSolidBrush} & \textcolor{red}{\XSolidBrush}      & 56.30$\pm$0.15 ($\downarrow$ 12.50)                                 & 63.11$\pm$0.17 ($\downarrow$ 4.61)                                   & 61.03$\pm$0.14 ($\downarrow$ 4.39)                                   & 63.06$\pm$0.18 ($\downarrow$ 2.75)                               & 61.02$\pm$0.26 ($\downarrow$ 2.94)                                   & 60.64$\pm$0.14 ($\downarrow$ 2.28)                               \\ \hline
  \textcolor{teal}{\CheckmarkBold}  & \textcolor{teal}{\CheckmarkBold}       & \textbf{68.80$\pm$0.04}                                             & \textbf{67.72$\pm$0.09}                                              & \textbf{65.42$\pm$0.07}                                              & \textbf{65.81$\pm$0.13}                                          & \textbf{63.96$\pm$0.19}                                              & \textbf{62.92$\pm$0.07}                                          \\ \hline
  \end{tabular}
  }
\end{table*}

\subsection{Case Study}
\begin{figure}[!t]
  \centering
  \subfloat[\footnotesize{Client 34 of \textit{CiteSeer} dataset\\(homophily level=0.891)}]{\includegraphics[width=0.5\columnwidth]{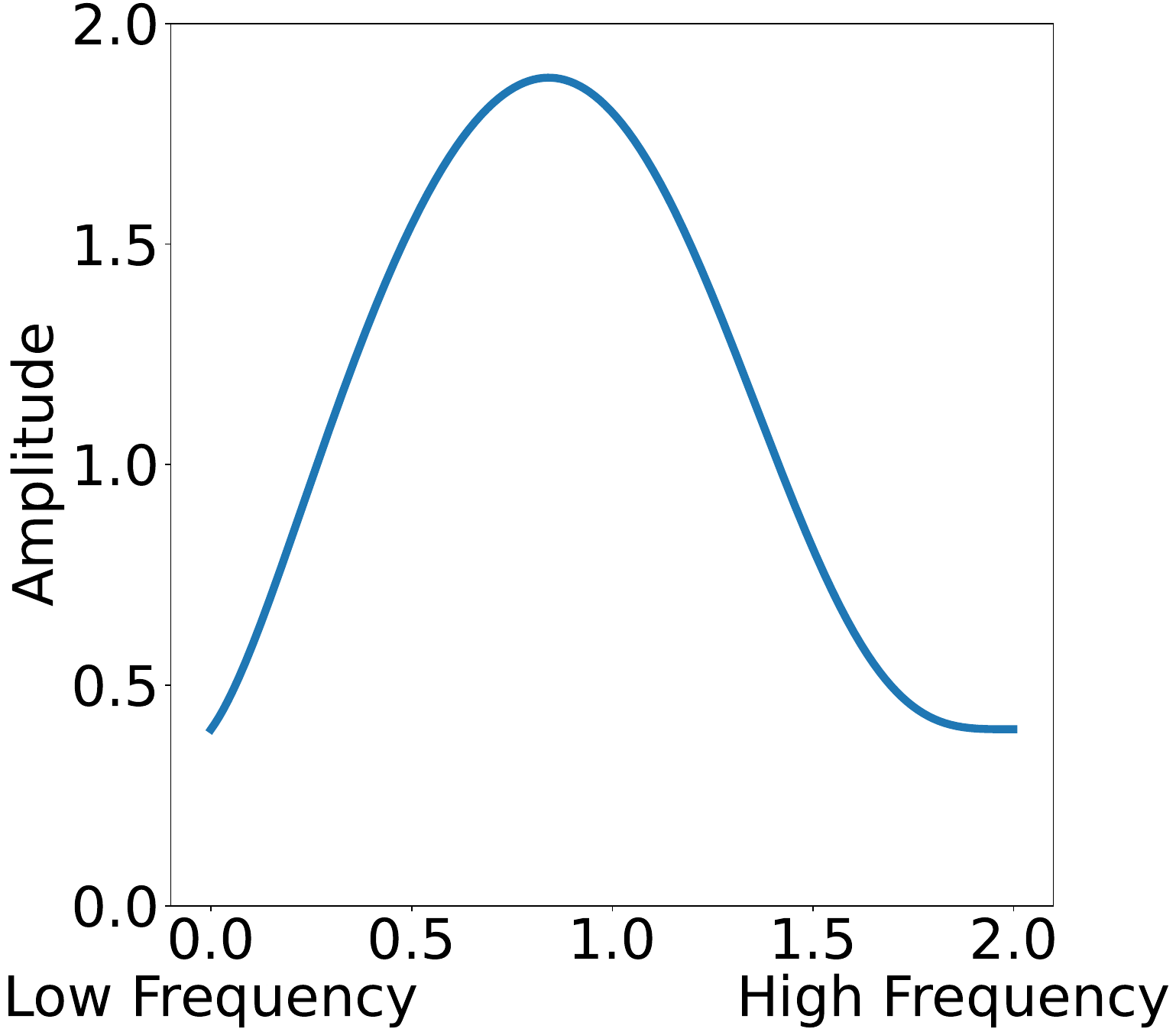}\label{fig_case_1}}
  \hfill
  \subfloat[\footnotesize{Client 16 of \textit{CiteSeer} dataset\\(homophily level=0.071)}]{\includegraphics[width=0.5\columnwidth]{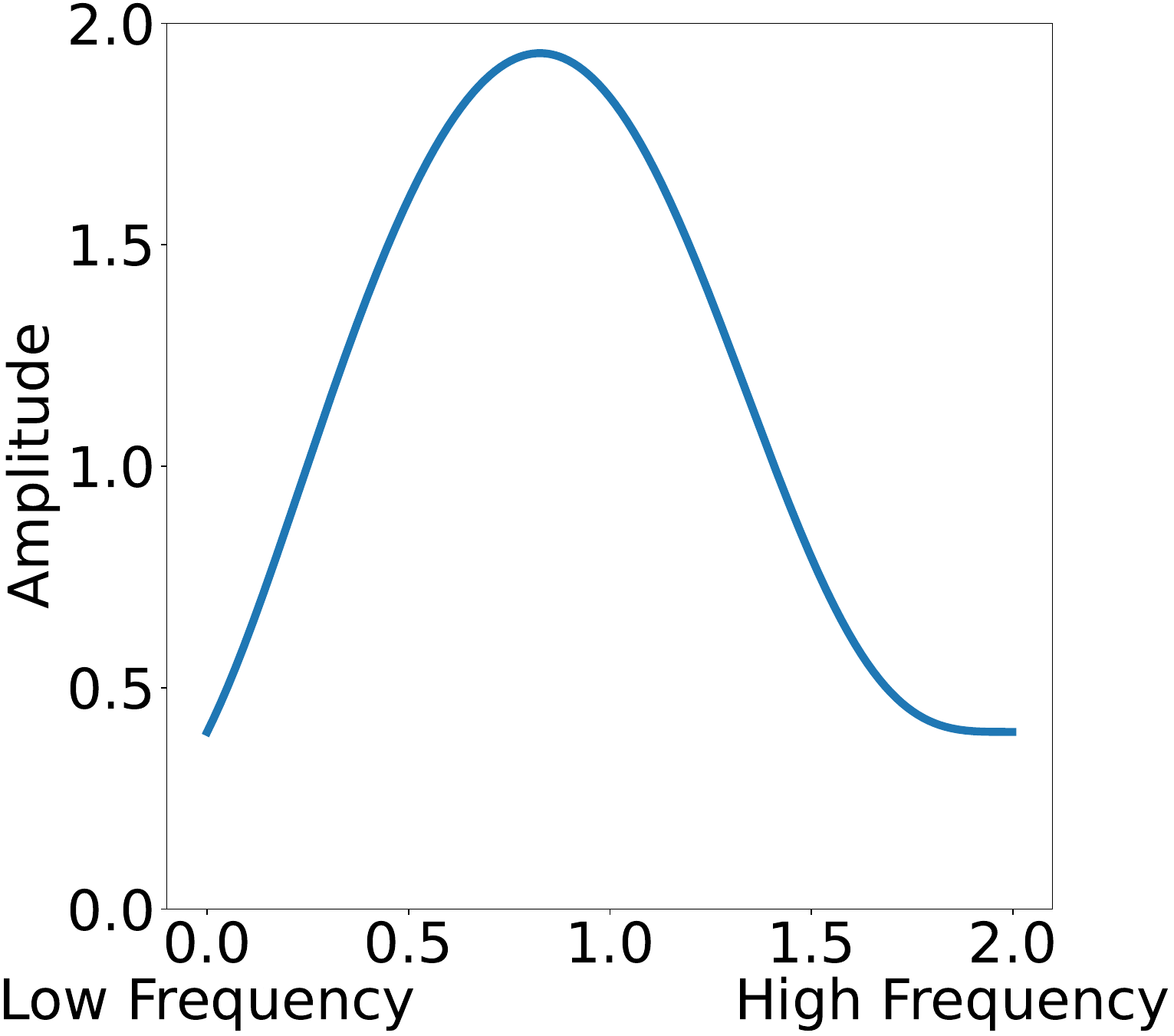}\label{fig_case_2}}
  \hfill
  \subfloat[\footnotesize{Client 7 of \textit{Questions} dataset\\(homophily level=0.212)}]{\includegraphics[width=0.5\columnwidth]{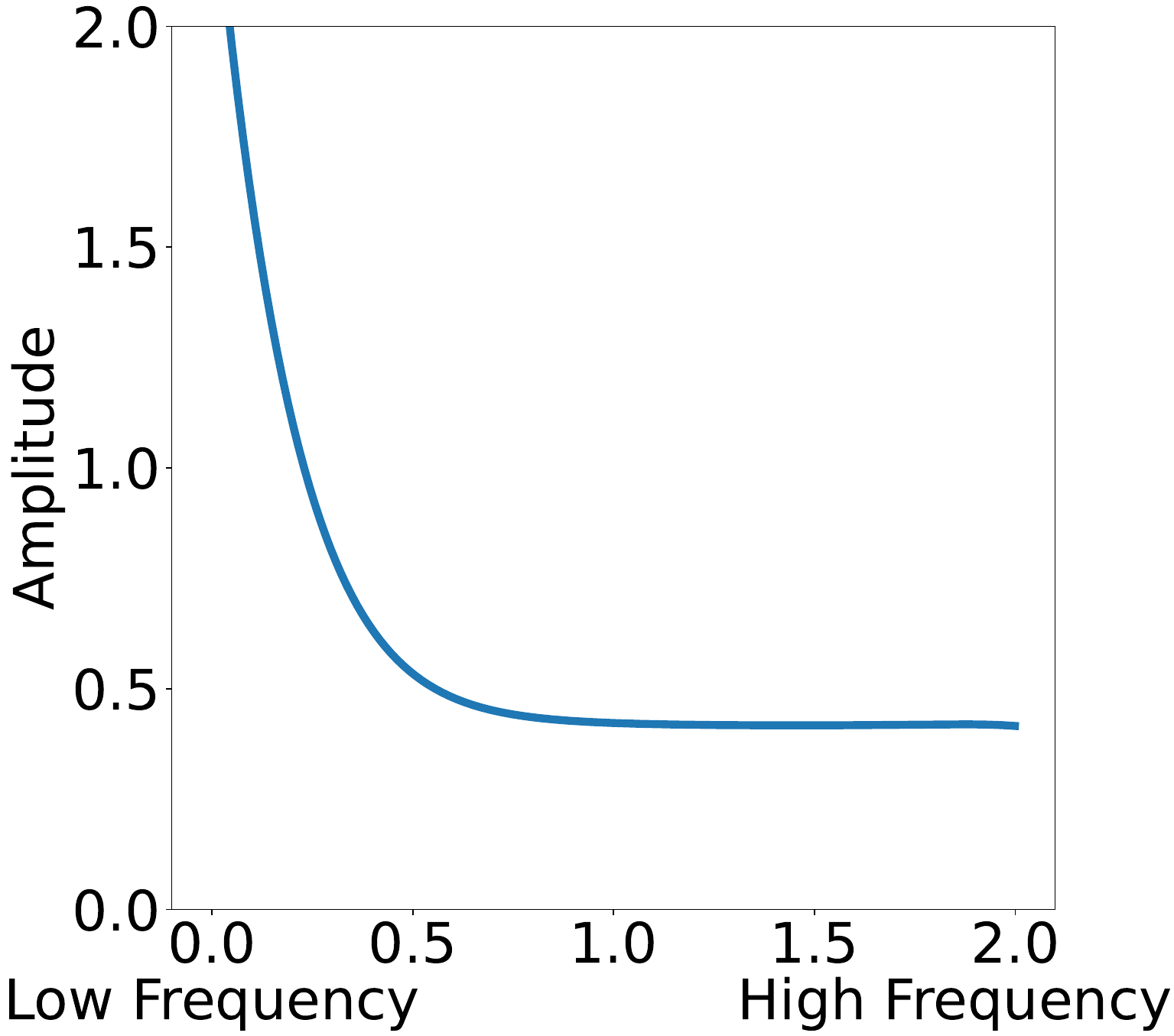}\label{fig_case_3}}
  \hfill
  \subfloat[\footnotesize{Client 6 of \textit{Questions} dataset\\(homophily level=-0.081)}]{\includegraphics[width=0.5\columnwidth]{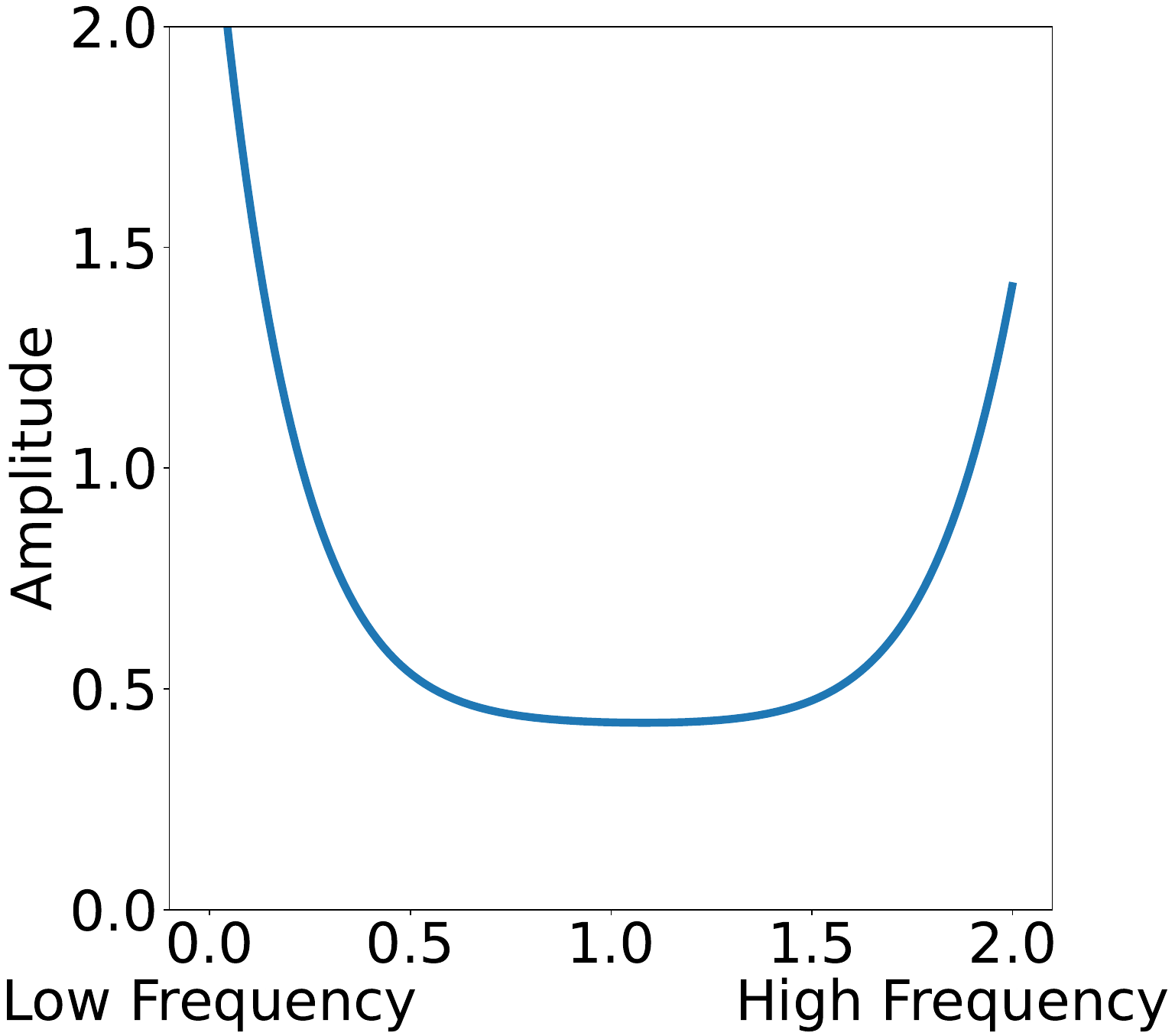}\label{fig_case_4}}
  \caption{Spectral properties captured by local models after the federated aggregation.}
  \label{fig_case}
\end{figure}

Since our proposed FedGSP not only shares generic spectral properties but also complements non-generic spectral properties among different clients, one might wonder what spectral properties are captured by local models after federation. Therefore, we carry out the case studies to analyze the spectral properties captured by local models after the federated aggregation. Comparing Fig.~\ref{fig2} and Fig.~\ref{fig_case}, we can find that our proposed FedGSP allows the local model on each client not only to preserve its local graph spectral properties, but also to obtain the additional spectral properties from collaborations. For example, compared with Fig.~\ref{fig2_1} and Fig.~\ref{fig2_2}, Fig.~\ref{fig_case_1} and Fig.~\ref{fig_case_2} show both the low-frequency and high-frequency properties due to obtaining the additional spectral properties from collaborations. Similarly, compared with Fig.~\ref{fig2_4}, the low-frequency properties are strengthened in Fig.~\ref{fig_case_4}. Furthermore, compared with Fig.~\ref{fig2_3}, the low-frequency properties are strengthened in Fig.~\ref{fig_case_3} due to preserving its local graph spectral properties.

\subsection{Convergence Curves}
\begin{figure*}[!t]
  \centering
  \subfloat[\footnotesize{\textit{Cora}}]{\includegraphics[width=0.51\columnwidth]{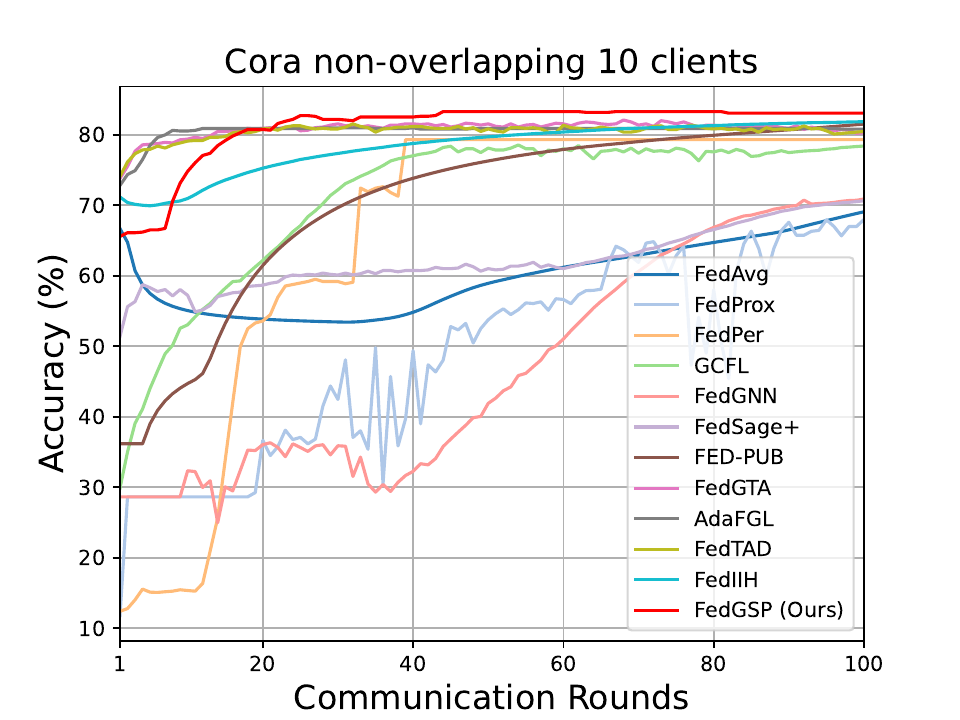}\label{fig5_1}}
  \hfill
  \subfloat[\footnotesize{\textit{CiteSeer}}]{\includegraphics[width=0.51\columnwidth]{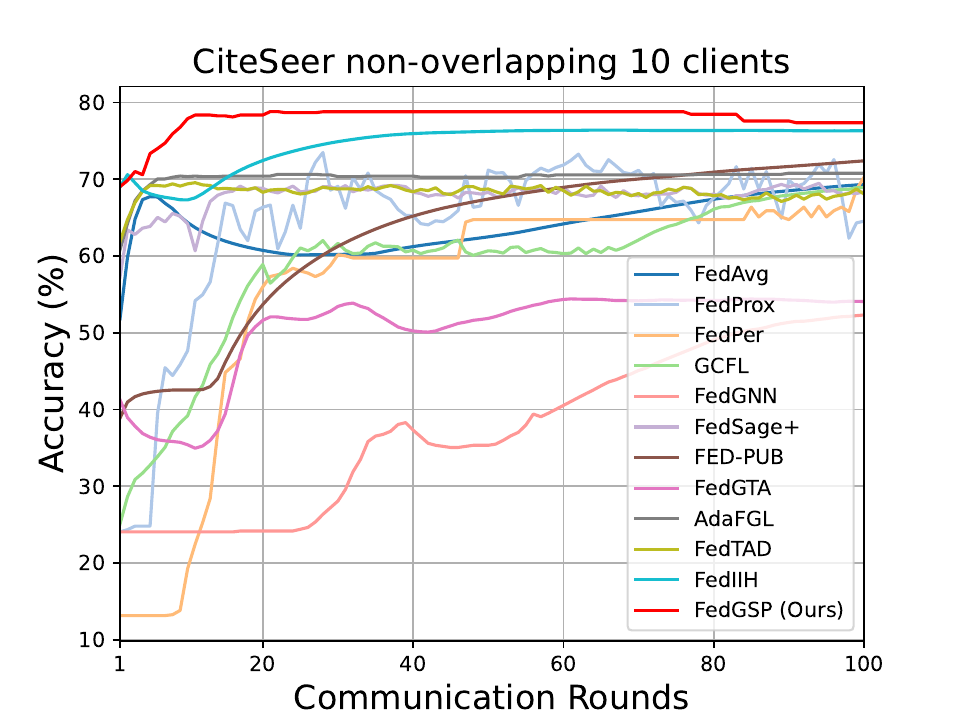}\label{fig5_2}}
  \hfill
  \subfloat[\footnotesize{\textit{PubMed}}]{\includegraphics[width=0.51\columnwidth]{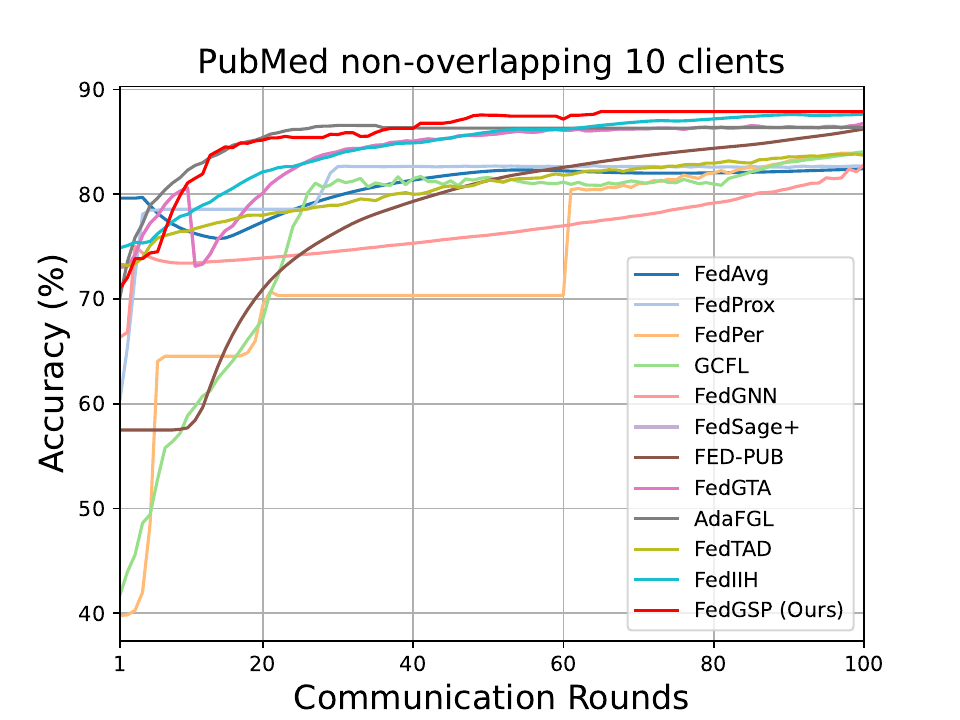}\label{fig5_3}}
  \hfill
  \subfloat[\footnotesize{\textit{ogbn-arxiv}}]{\includegraphics[width=0.51\columnwidth]{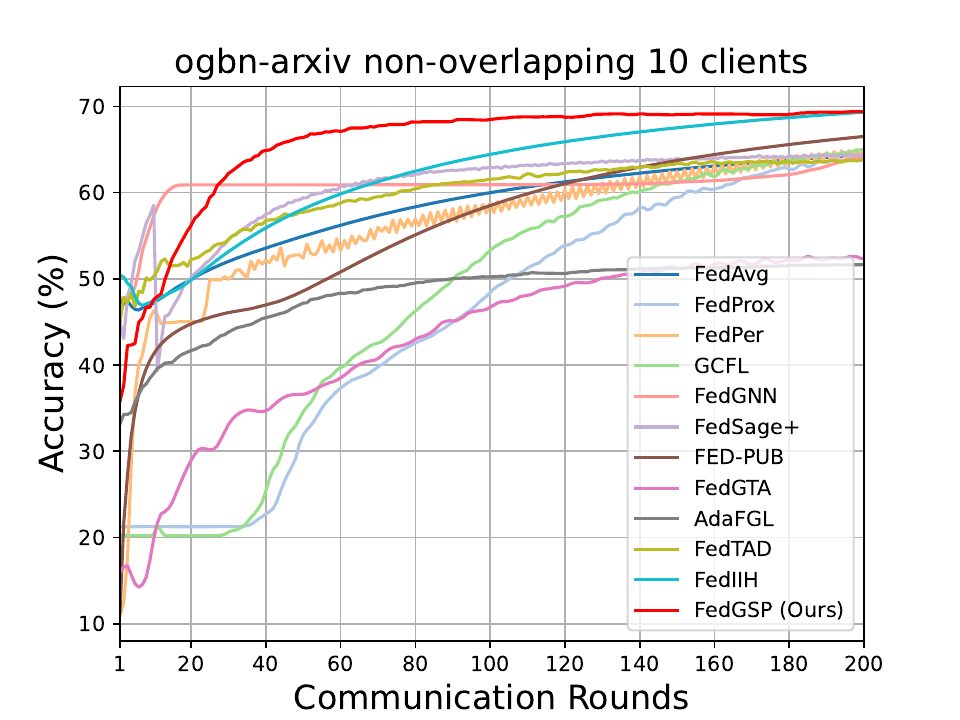}\label{fig5_4}}
  \hfill
  \subfloat[\footnotesize{\textit{Roman-empire}}]{\includegraphics[width=0.51\columnwidth]{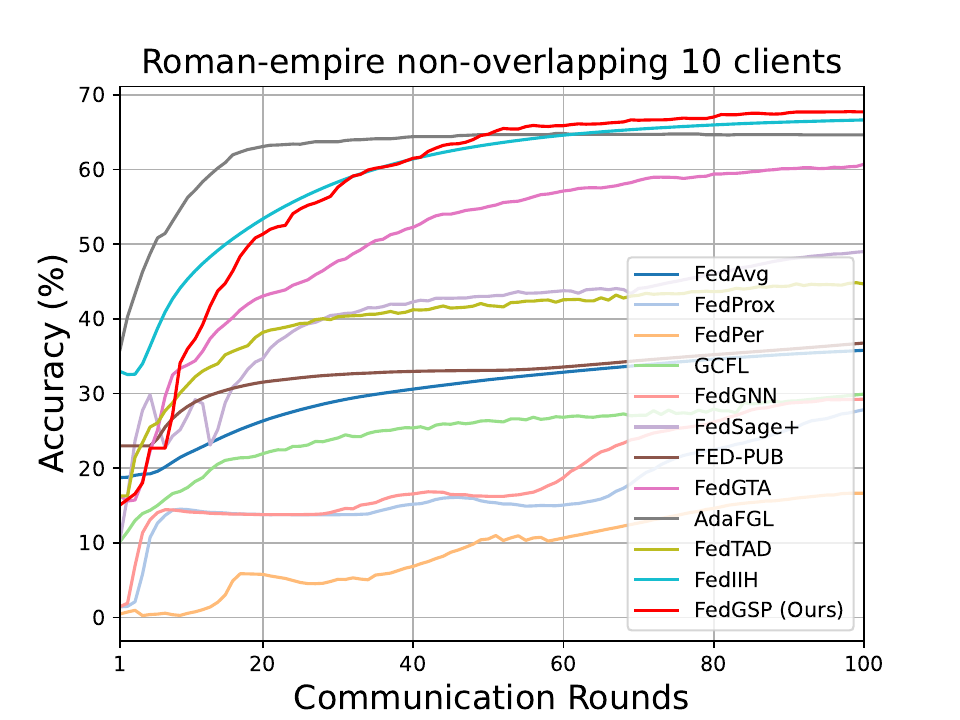}\label{fig5_5}}
  \hfill
  \subfloat[\footnotesize{\textit{Minesweeper}}]{\includegraphics[width=0.51\columnwidth]{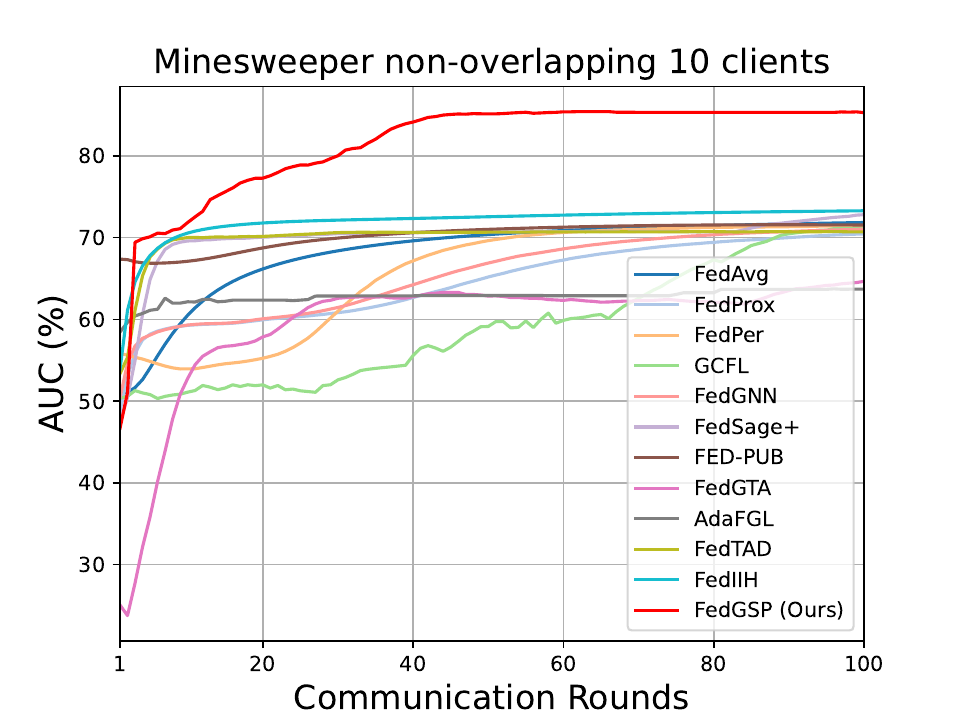}\label{fig5_6}}
  \hfill
  \subfloat[\footnotesize{\textit{Tolokers}}]{\includegraphics[width=0.51\columnwidth]{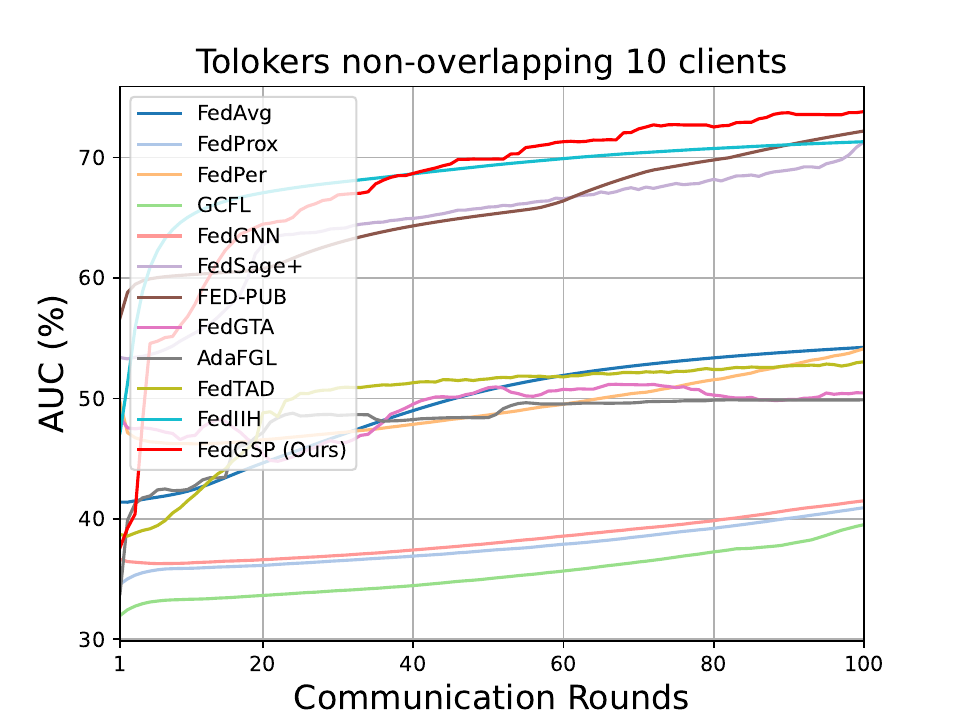}\label{fig5_7}}
  \hfill
  \subfloat[\footnotesize{\textit{Questions}}]{\includegraphics[width=0.51\columnwidth]{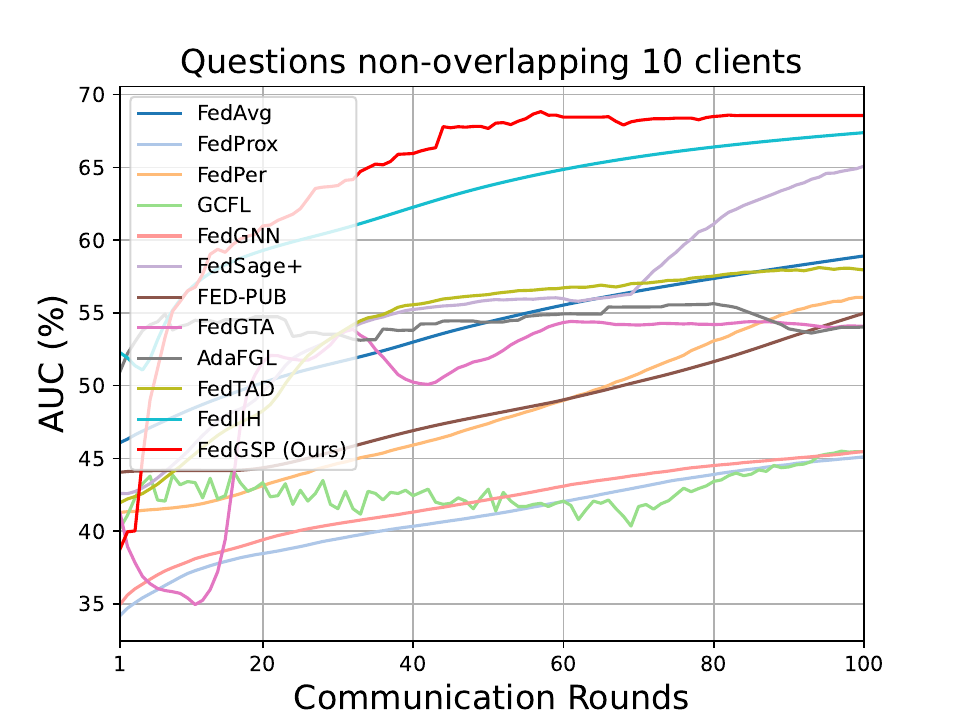}\label{fig5_8}}
  \caption{Convergence curves in the non-overlapping partitioning settings on eight datasets with 10 clients.}
  \label{fig5}
\end{figure*}

\begin{figure*}[!t]
  \centering
  \subfloat[\footnotesize{\textit{Cora}}]{\includegraphics[width=0.51\columnwidth]{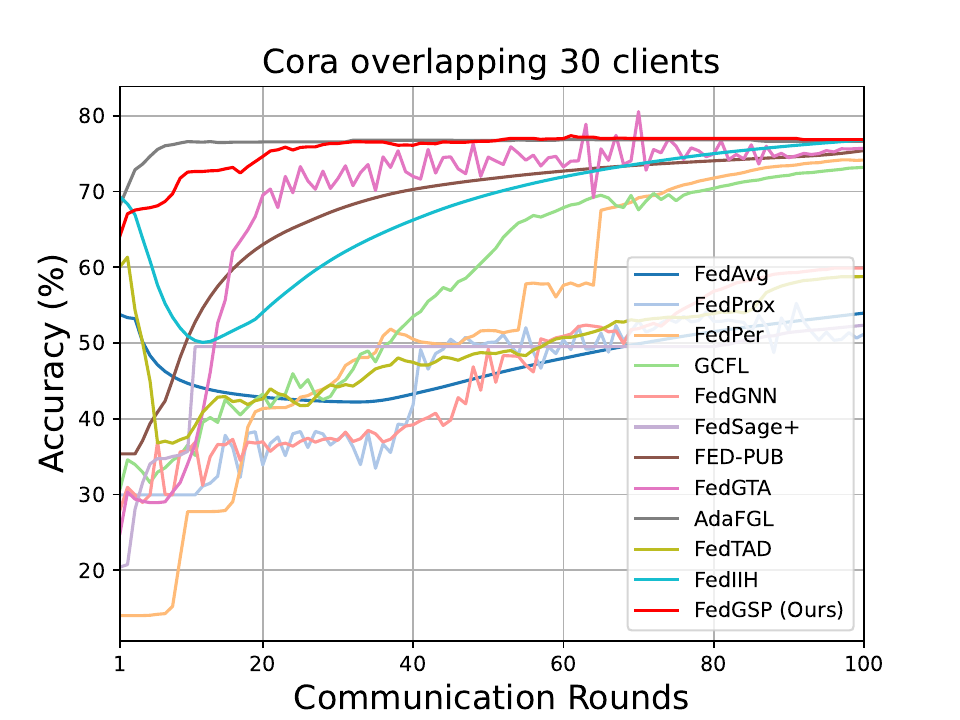}\label{fig6_1}}
  \hfill
  \subfloat[\footnotesize{\textit{CiteSeer}}]{\includegraphics[width=0.51\columnwidth]{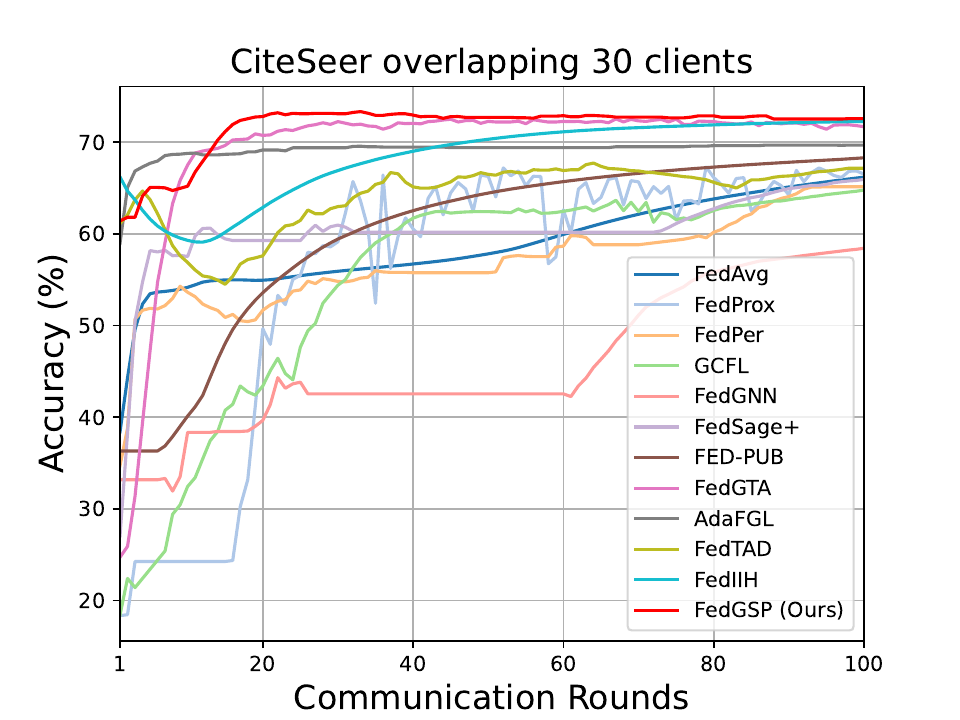}\label{fig6_2}}
  \hfill
  \subfloat[\footnotesize{\textit{PubMed}}]{\includegraphics[width=0.51\columnwidth]{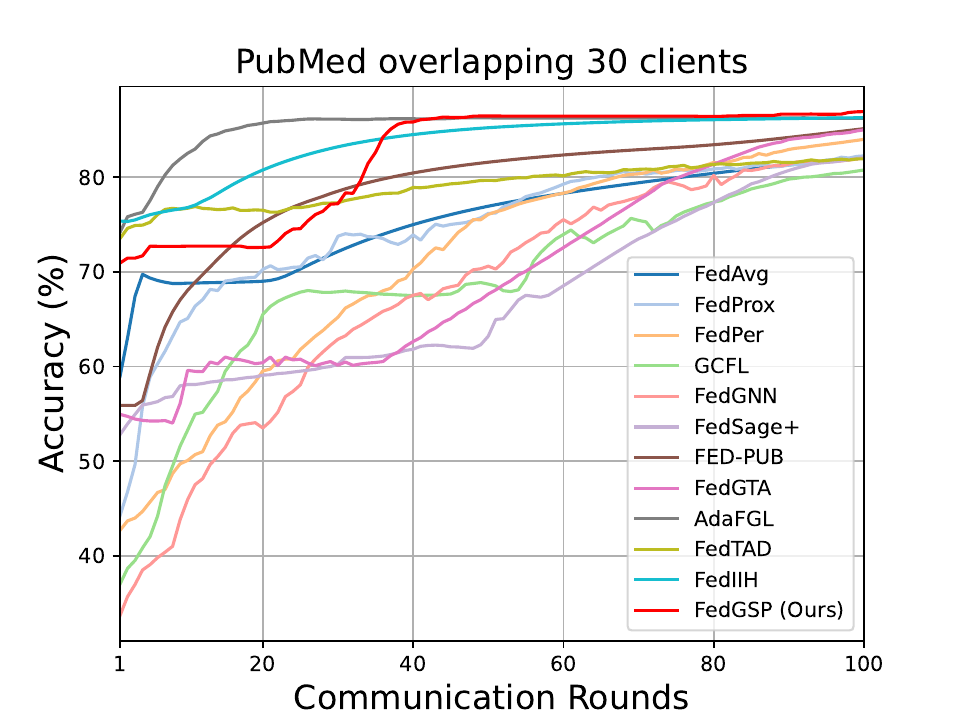}\label{fig6_3}}
  \hfill
  \subfloat[\footnotesize{\textit{ogbn-arxiv}}]{\includegraphics[width=0.51\columnwidth]{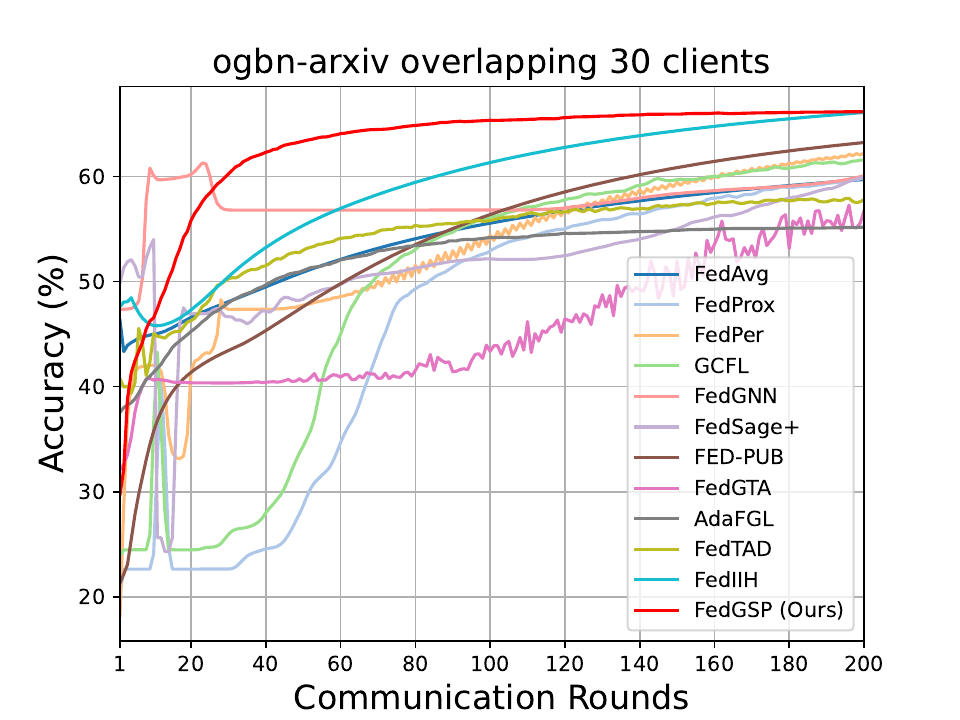}\label{fig6_4}}
  \hfill
  \subfloat[\footnotesize{\textit{Roman-empire}}]{\includegraphics[width=0.51\columnwidth]{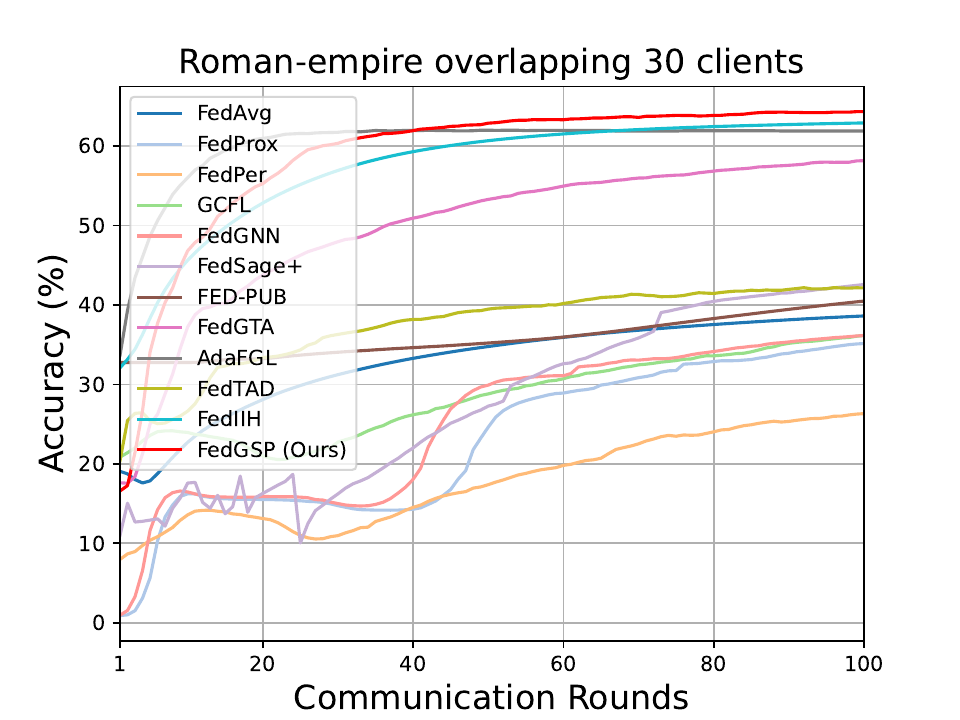}\label{fig6_5}}
  \hfill
  \subfloat[\footnotesize{\textit{Minesweeper}}]{\includegraphics[width=0.51\columnwidth]{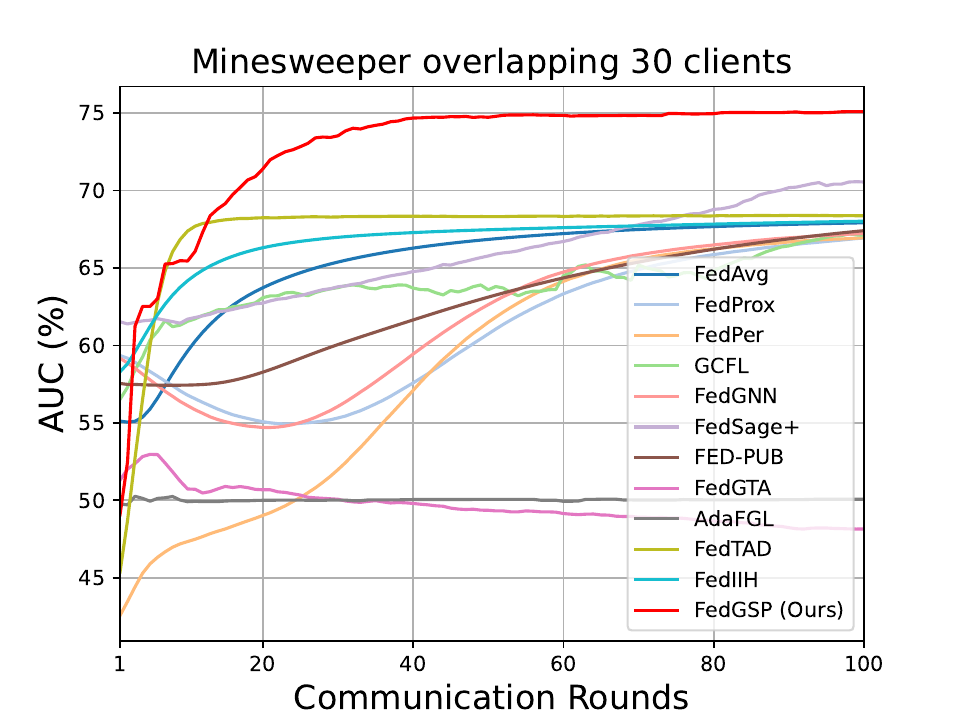}\label{fig6_6}}
  \hfill
  \subfloat[\footnotesize{\textit{Tolokers}}]{\includegraphics[width=0.51\columnwidth]{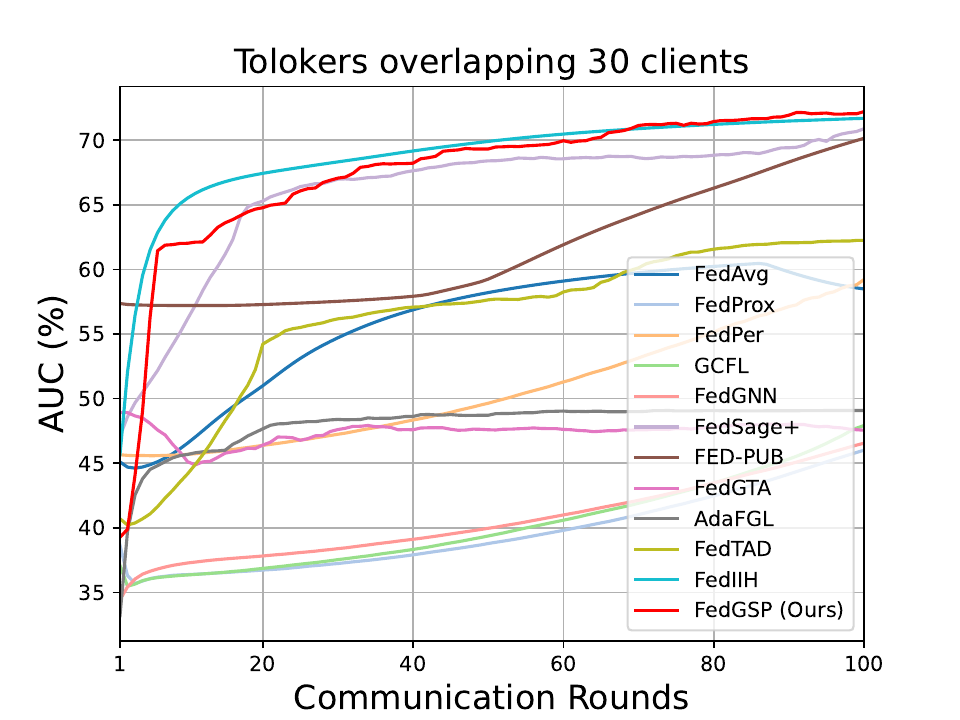}\label{fig6_7}}
  \hfill
  \subfloat[\footnotesize{\textit{Questions}}]{\includegraphics[width=0.51\columnwidth]{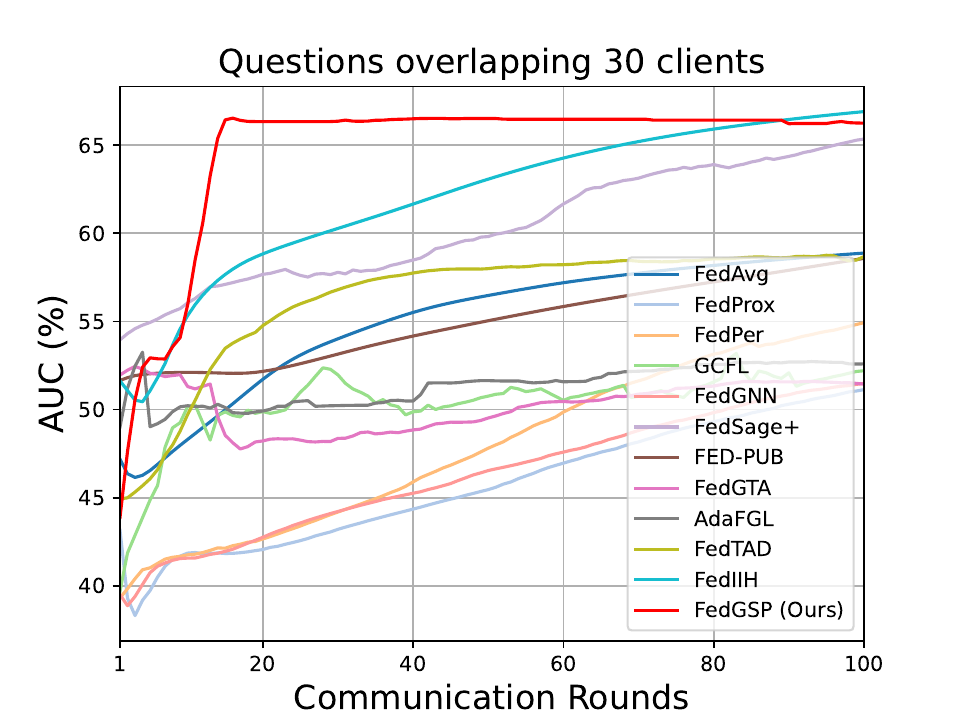}\label{fig6_8}}
  \caption{Convergence curves in the overlapping partitioning settings on eight datasets with 30 clients.}
  \label{fig6}
\end{figure*}

Here we plot the convergence curves of our proposed FedGSP and the baseline methods in Fig.~\ref{fig5} and Fig.~\ref{fig6}. We can find that our proposed FedGSP converges fastly within several communication rounds (\textit{e.g.}, Fig.~\ref{fig5_2} and Fig.~\ref{fig6_8}). Moreover, the fluctuations in the curves of our FedGSP are quite low, which confirms its stability. These results validate that FedGSP can be employed for various practical applications. This can be attributed to our efficient optimization method. In contrast, the convergence curves of GCFL (\textit{e.g.}, Fig.~\ref{fig5_8}) and FedGTA (\textit{e.g.}, Fig.~\ref{fig6_1}) are not stable enough. This is because GCFL and FedGTA perform the federated aggregation based on the clustering result and the similarity matrix between clients, respectively. However, both are unstable across communication rounds due to variations in local models caused by homophily heterogeneity.

\subsection{Time of Each Communication Round}
\label{time_communication}
To demonstrate the efficiency of our proposed FedGSP over the baseline methods, we report the time of each communication round for the proposed FedGSP and the compared baseline methods. According to Tab.~\ref{table6}, the time cost of our proposed method is significantly lower than the average in most scenarios. Furthermore, we can find that our proposed FedGSP is more efficient than several promising baseline methods (\textit{i.e.}, FED-PUB and FedIIH). Specifically, compared to the second-best baseline method (\textit{i.e.}, FedIIH), FedGSP achieves more than three times speed improvement. For example, on the \textit{Cora} dataset in the overlapping subgraph partitioning setting with 30 clients, the accuracies of FedGSP and FedIIH are 76.86 and 76.82, respectively. However, the time cost of our FedGSP and FedIIH are 7.30 seconds and 65.49 seconds, respectively. This is because, in each communication round, FedIIH employs a hierarchical variation inference model to infer the distribution of the subgraph data on each client, which is computationally expensive. Although our proposed FedGSP utilizes the Lagrange multiplier method and the Newton method to optimize the collaboration graphs, their time complexity is small. Although several baseline methods (\textit{i.e.}, AdaFGL and FedGNN) cost less time than our FedGSP, their performances are not good enough than FedGSP. In addition, the time cost of FedGSP slowly increases as the number of clients increases, demonstrating that our proposed FedGSP is acceptable in real-world applications.

\begin{table*}[t]
  \centering
  \scriptsize
  \caption{The time (seconds) of each communication round for the proposed FedGSP and the compared baseline methods on the \textit{Cora} and \textit{Roman-empire} datasets.}
  \label{table6}
  \renewcommand{\arraystretch}{0.9} 
  \scalebox{0.9}{
  \begin{tabular}{ccccccc}
  \hline
  \rowcolor{gray!50}
                & \multicolumn{6}{c}{Cora}                                                                                                                                                                                                                                                                                                                                                                                                       \\ \hline
  Methods       & \begin{tabular}[c]{@{}c@{}}non-overlapping\\ 5 clients\end{tabular} & \begin{tabular}[c]{@{}c@{}}non-overlapping\\ 10 clients\end{tabular} & \begin{tabular}[c]{@{}c@{}}non-overlapping\\ 20 clients\end{tabular} & \begin{tabular}[c]{@{}c@{}}overlapping\\ 10 clients\end{tabular} & \begin{tabular}[c]{@{}c@{}}overlapping\\ 30 clients\end{tabular} & \begin{tabular}[c]{@{}c@{}}overlapping\\ 50 clients\end{tabular} \\ \hline
  FedAvg~\cite{mcmahan2017communication}        & 5.51                                                                & 2.19                                                                 & 5.63                                                                 & 5.40                                                             & 7.27                                                                 & 12.08                                                            \\ \hline
  FedProx~\cite{MLSYS2020_1f5fe839}       & 4.24                                                                & 4.65                                                                 & 8.86                                                                 & 5.49                                                             & 13.71                                                                & 22.43                                                            \\ \hline
  FedPer~\cite{Arivazhagan2019}        & 4.06                                                                & 4.13                                                                 & 8.17                                                                 & 4.16                                                             & 12.38                                                                & 20.24                                                            \\ \hline
  GCFL~\cite{NEURIPS2021_9c6947bd}          & 6.44                                                                & 9.42                                                                 & 18.63                                                                & 9.04                                                             & 27.70                                                                & 46.77                                                            \\ \hline
  FedGNN~\cite{wu2021fedgnn}        & 2.28                                                                & 4.42                                                                 & 8.76                                                                 & 5.40                                                             & 13.07                                                                & 23.04                                                            \\ \hline
  FedSage+\cite{NEURIPS2021_34adeb8e}      & 6.88                                                                & 8.55                                                                 & 17.88                                                                & 9.37                                                             & 16.97                                                                & 23.35                                                            \\ \hline
  FED-PUB~\cite{baek2023personalized}       & 22.04                                                               & 27.34                                                                & 60.31                                                                & 33.46                                                            & 80.54                                                                & 147.84                                                           \\ \hline
  FedGTA~\cite{li2023fedgta}        & 3.36                                                                & 2.24                                                                 & 5.89                                                                 & 4.30                                                             & 5.00                                                                 & 7.18                                                             \\ \hline
  AdaFGL~\cite{li2024adafgl}        & 1.99                                                                & 2.49                                                                 & 4.63                                                                 & 4.44                                                             & 6.16                                                                 & 7.83                                                             \\ \hline
  FedTAD~\cite{zhu2024fedtad}        & 4.91                                                                & 5.25                                                                 & 9.13                                                                 & 5.22                                                             & 12.84                                                                & 19.71                                                            \\ \hline
  FedIIH~\cite{wentao2025fediih}        & 19.57                                                               & 22.76                                                                & 56.09                                                                & 19.67                                                            & 65.49                                                                & 139.03                                                           \\ \hline
  FedGSP (Ours) & 6.73                                                                & 2.26                                                                 & 5.86                                                                 & 5.18                                                             & 7.30                                                                 & 11.90                                                            \\ \hline
  \rowcolor{green!5} Average       & 7.33                                                                & 7.98                                                                 & 17.49                                                                & 9.26                                                             & 22.37                                                            & 40.12                                                            \\ \hline
  \rowcolor{gray!50}
                & \multicolumn{6}{c}
                {Roman-empire}                                                                                                                                                                                                                                                                                                                                                                                               \\ \hline
  Methods       & \begin{tabular}[c]{@{}c@{}}non-overlapping\\ 5 clients\end{tabular} & \begin{tabular}[c]{@{}c@{}}non-overlapping\\ 10 clients\end{tabular} & \begin{tabular}[c]{@{}c@{}}non-overlapping\\ 20 clients\end{tabular} & \begin{tabular}[c]{@{}c@{}}overlapping\\ 10 clients\end{tabular} & \begin{tabular}[c]{@{}c@{}}overlapping\\ 30 clients\end{tabular} & \begin{tabular}[c]{@{}c@{}}overlapping\\ 50 clients\end{tabular} \\ \hline
  FedAvg~\cite{mcmahan2017communication}        & 8.20                                                                & 5.76                                                                 & 8.46                                                                 & 10.25                                                            & 18.47                                                                & 19.12                                                            \\ \hline
  FedProx~\cite{MLSYS2020_1f5fe839}       & 4.31                                                                & 6.73                                                                 & 9.04                                                                 & 6.90                                                             & 16.25                                                                & 23.66                                                            \\ \hline
  FedPer~\cite{Arivazhagan2019}        & 5.35                                                                & 6.19                                                                 & 9.19                                                                 & 6.47                                                             & 15.58                                                                & 20.22                                                            \\ \hline
  GCFL~\cite{NEURIPS2021_9c6947bd}          & 6.58                                                                & 9.39                                                                 & 18.32                                                                & 10.78                                                            & 29.05                                                                & 50.58                                                            \\ \hline
  FedGNN~\cite{wu2021fedgnn}        & 3.33                                                                & 6.60                                                                 & 9.43                                                                 & 6.45                                                             & 15.29                                                                & 22.82                                                            \\ \hline
  FedSage+\cite{NEURIPS2021_34adeb8e}      & 10.06                                                               & 14.82                                                                & 26.27                                                                & 23.09                                                            & 47.42                                                                & 62.72                                                            \\ \hline
  FED-PUB~\cite{baek2023personalized}       & 18.81                                                               & 28.03                                                                & 61.75                                                                & 28.45                                                            & 83.07                                                                & 133.05                                                           \\ \hline
  FedGTA~\cite{li2023fedgta}        & 2.17                                                                & 3.58                                                                 & 5.32                                                                 & 3.42                                                             & 6.12                                                                 & 9.92                                                             \\ \hline
  AdaFGL~\cite{li2024adafgl}        & 4.34                                                                & 4.14                                                                 & 5.55                                                                 & 6.49                                                             & 7.80                                                                 & 10.69                                                            \\ \hline
  FedTAD~\cite{zhu2024fedtad}        & 4.88                                                                & 8.55                                                                 & 15.03                                                                & 10.98                                                            & 22.42                                                                & 37.01                                                            \\ \hline
  FedIIH~\cite{wentao2025fediih}        & 17.45                                                               & 24.90                                                                & 47.01                                                                & 28.19                                                            & 61.17                                                                & 100.10                                                           \\ \hline
  FedGSP (Ours) & 9.09                                                                & 5.80                                                                 & 8.70                                                                 & 10.39                                                            & 18.36                                                                & 19.13                                                            \\ \hline
  \rowcolor{green!5} Average       & 7.88                                                                & 10.37                                                                & 18.67                                                                & 12.66                                                            & 28.42                                                            & 42.42                                                            \\ \hline
  \end{tabular}
  }
\end{table*}

\subsection{Sensitivity Analysis on Hyperparameters}
\begin{figure}[!t]
  \centering
  \subfloat[\footnotesize{\textit{Cora} of hyperparameters $K$}]{\includegraphics[width=0.5\columnwidth]{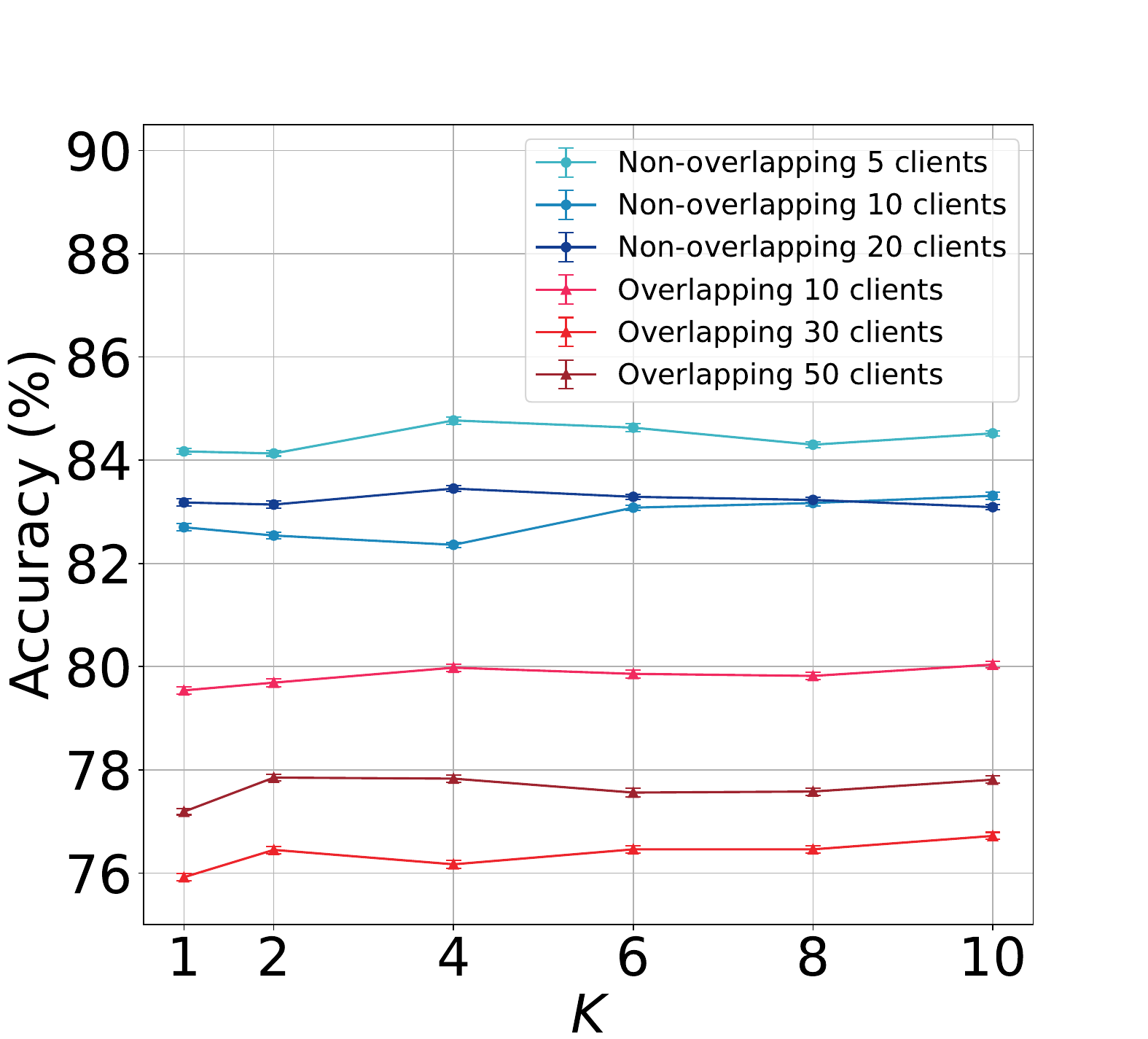}\label{fig7_1}}
  \hfill
  \subfloat[\footnotesize{\textit{Roman-empire} of hyperparameters $K$}]{\includegraphics[width=0.5\columnwidth]{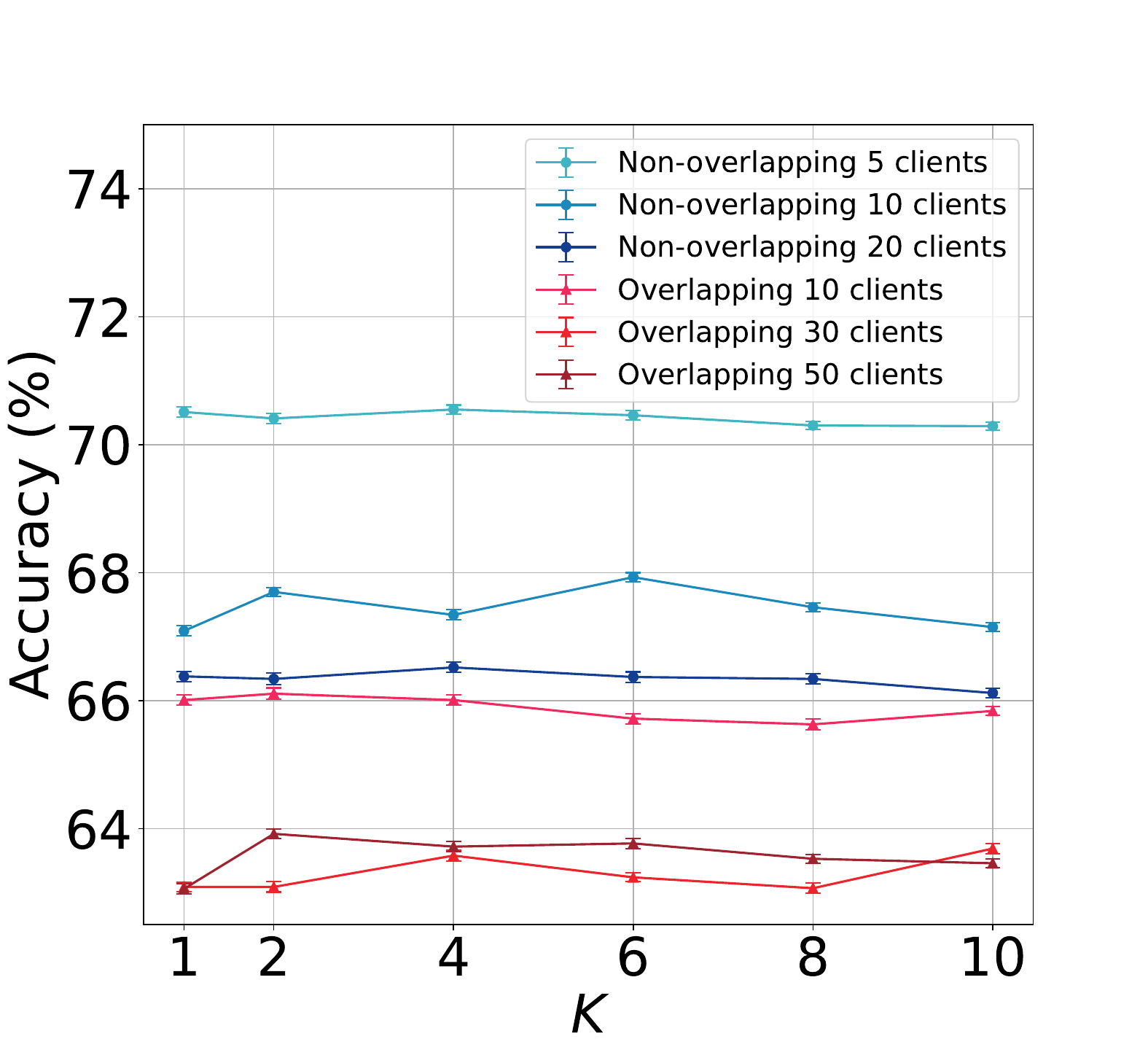}\label{fig7_2}}
  \hfill
  \subfloat[\footnotesize{\textit{Cora}} of hyperparameters $t$]{\includegraphics[width=0.5\columnwidth]{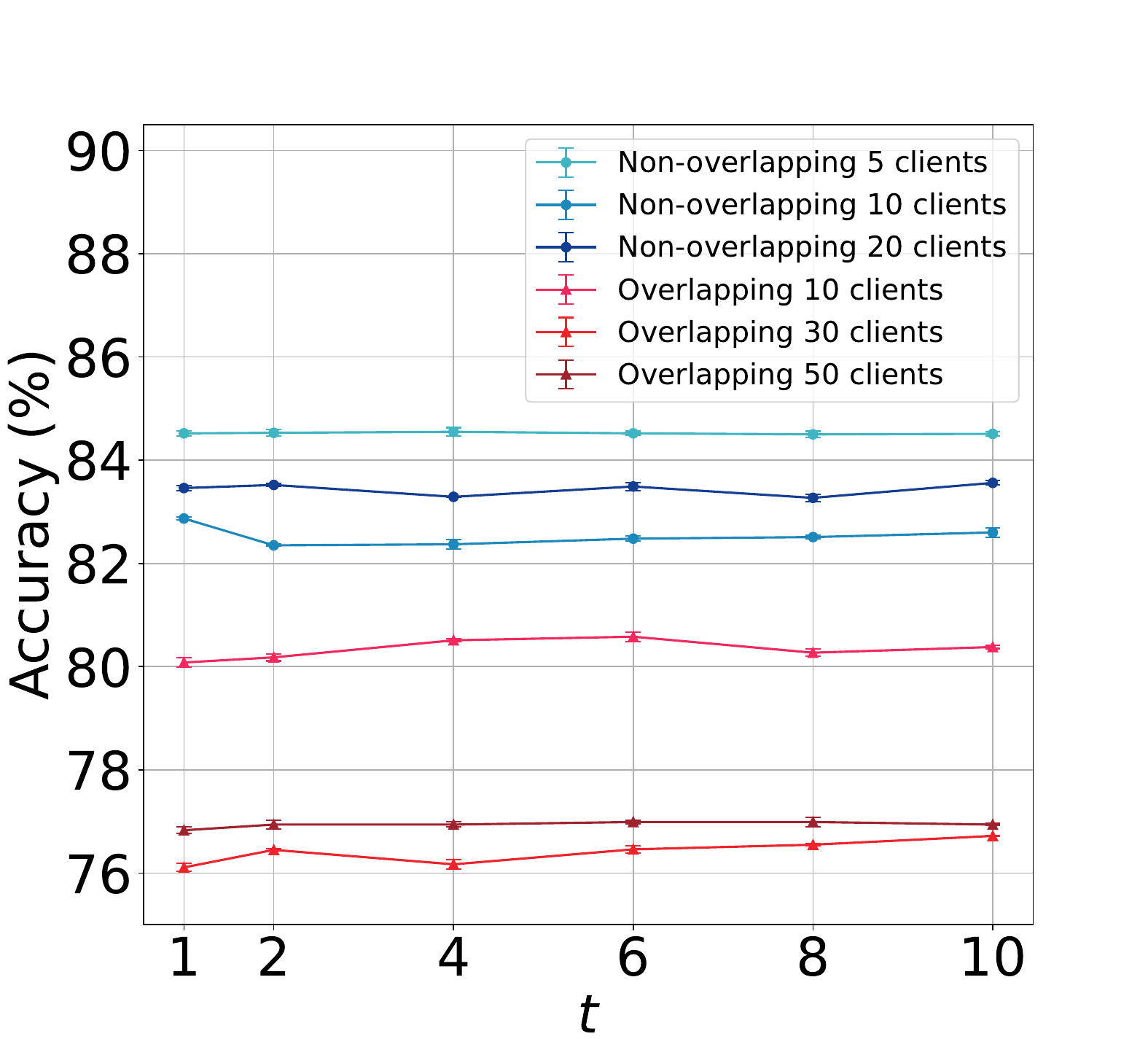}\label{fig7_3}}
  \hfill
  \subfloat[\footnotesize{\textit{Roman-empire}} of hyperparameters $t$]{\includegraphics[width=0.5\columnwidth]{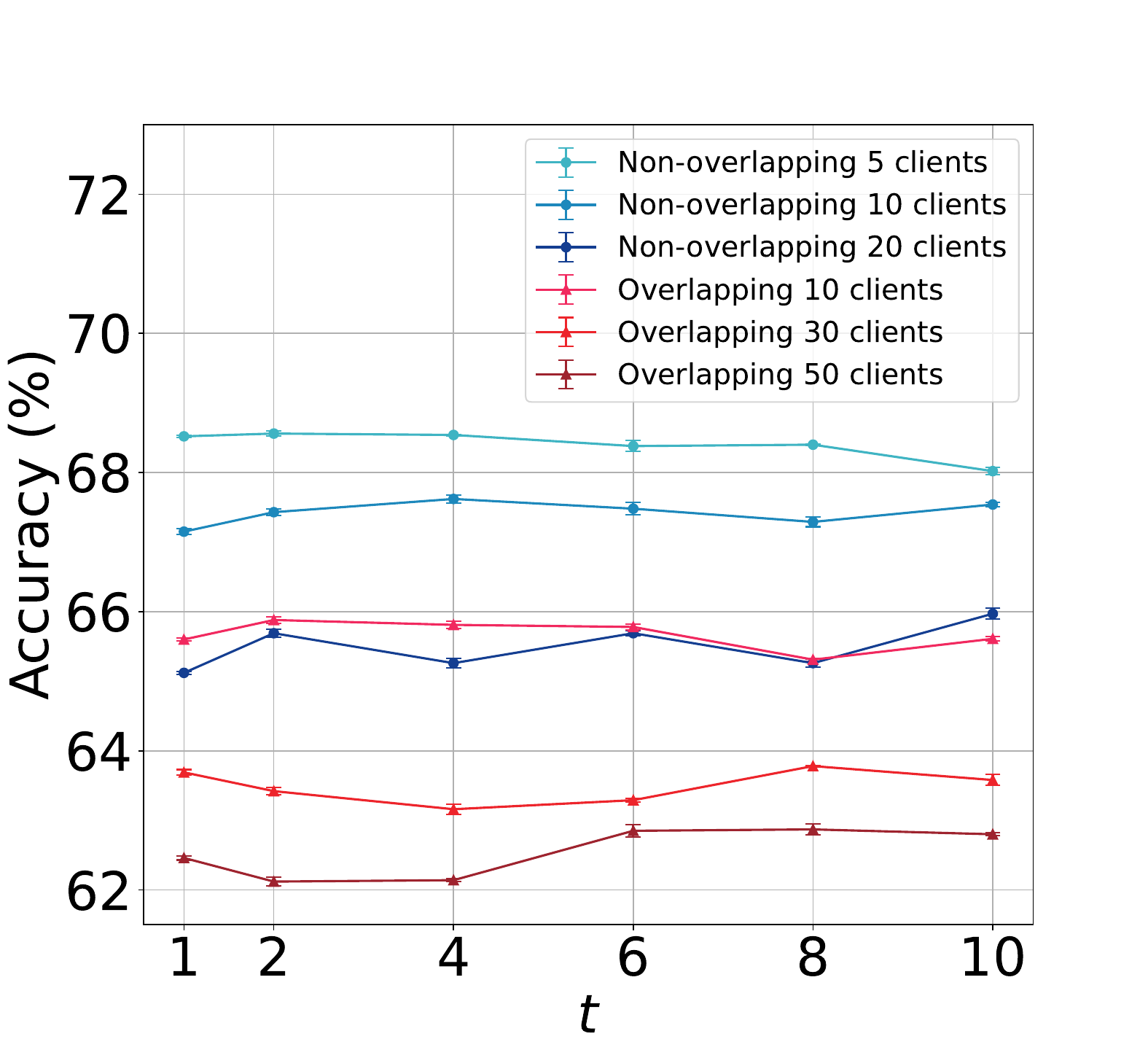}\label{fig7_4}}
  \caption{The sensitivity analyses of hyperparameters $K$ and $t$, respectively.}
  \label{fig7}
\end{figure}

\begin{figure}[!t]
  \centering
  \subfloat[\footnotesize{\textit{Cora} non-overlapping 10 clients}]{\includegraphics[width=0.465\columnwidth]{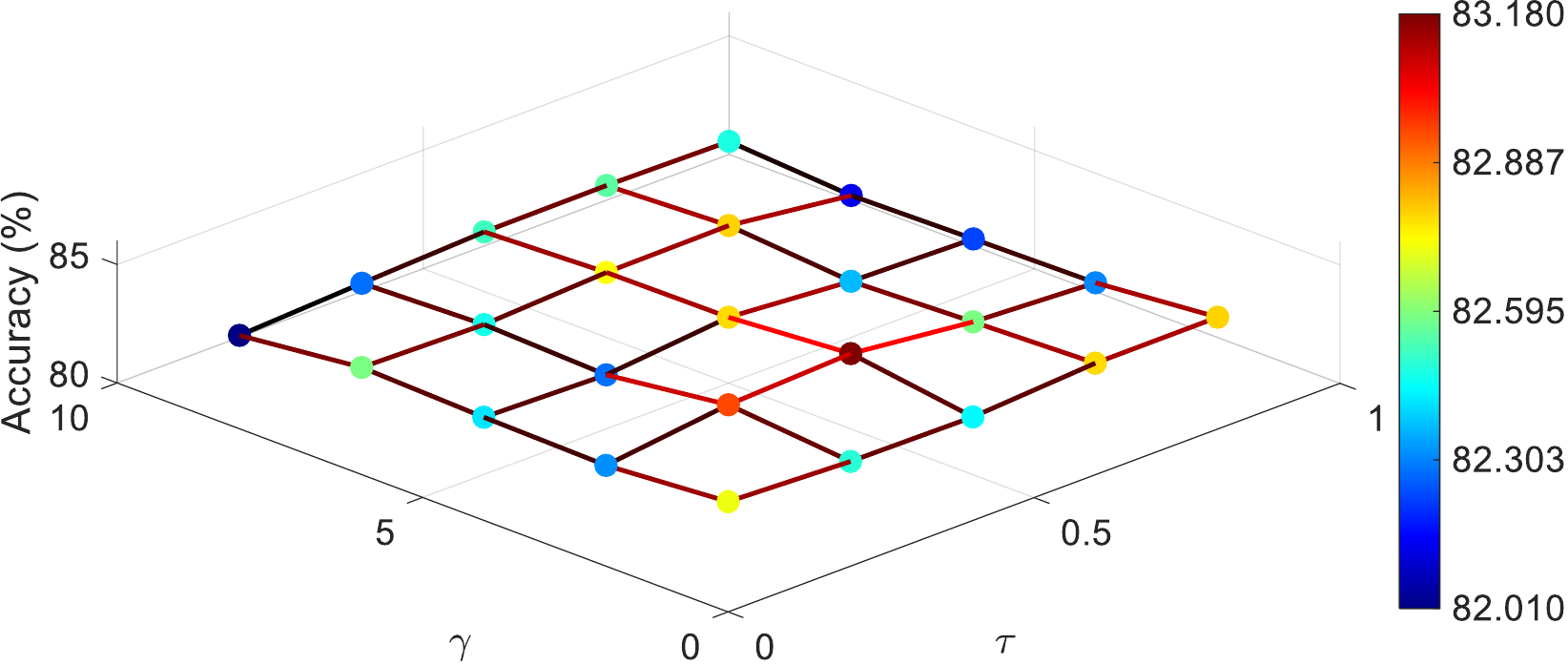}\label{fig8_1}}
  \hfill
  \subfloat[\footnotesize{\textit{Cora} overlapping 30 clients}]{\includegraphics[width=0.465\columnwidth]{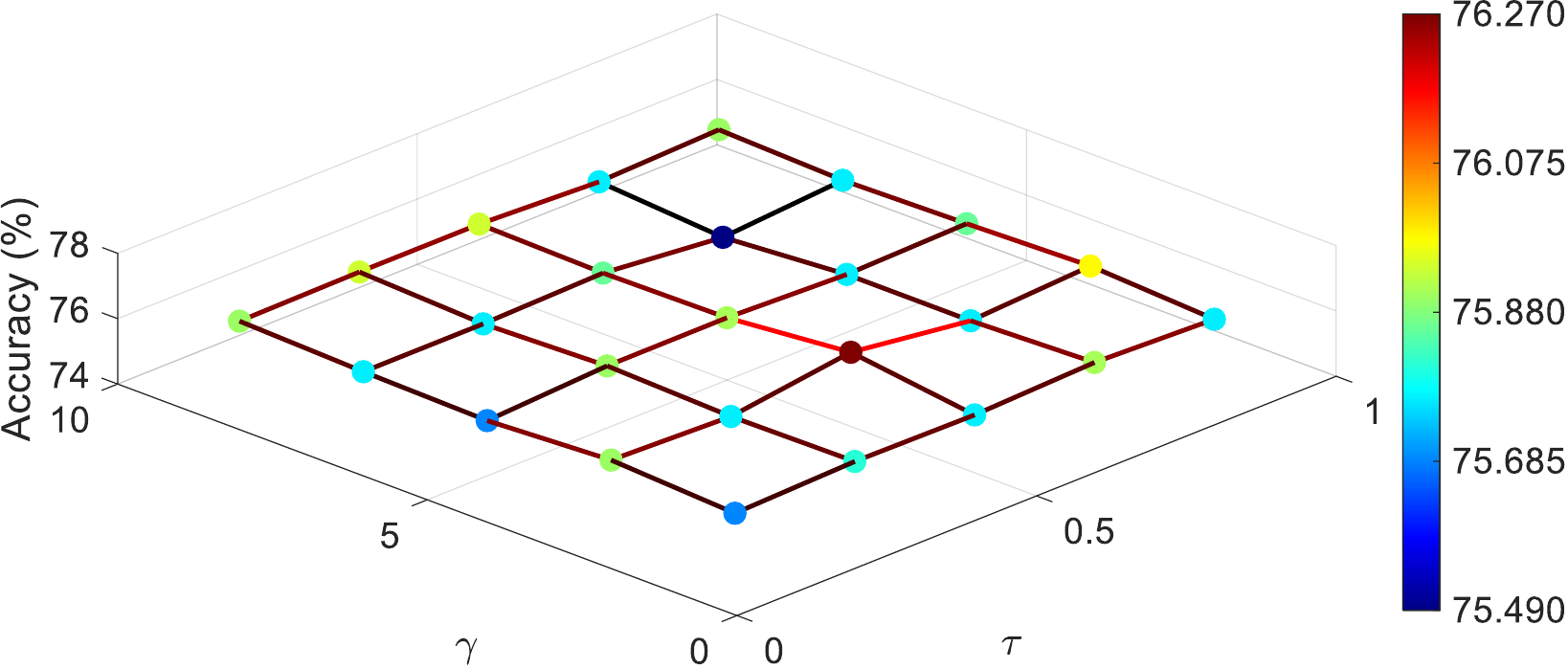}\label{fig8_2}}
  \hfill
  \subfloat[\footnotesize{\textit{Roman-empire} non-overlapping 10 clients}]{\includegraphics[width=0.465\columnwidth]{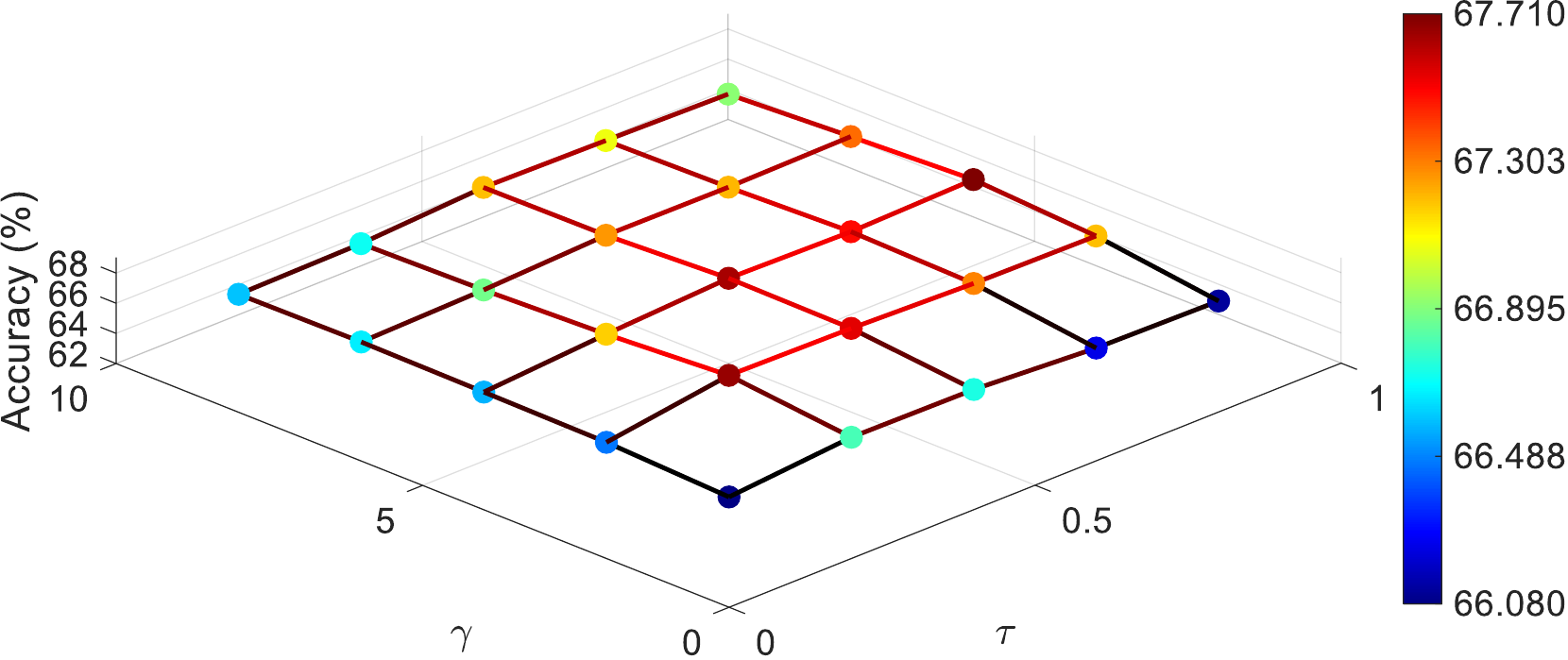}\label{fig8_3}}
  \hfill
  \subfloat[\footnotesize{\textit{Roman-empire} overlapping 30 clients}]{\includegraphics[width=0.465\columnwidth]{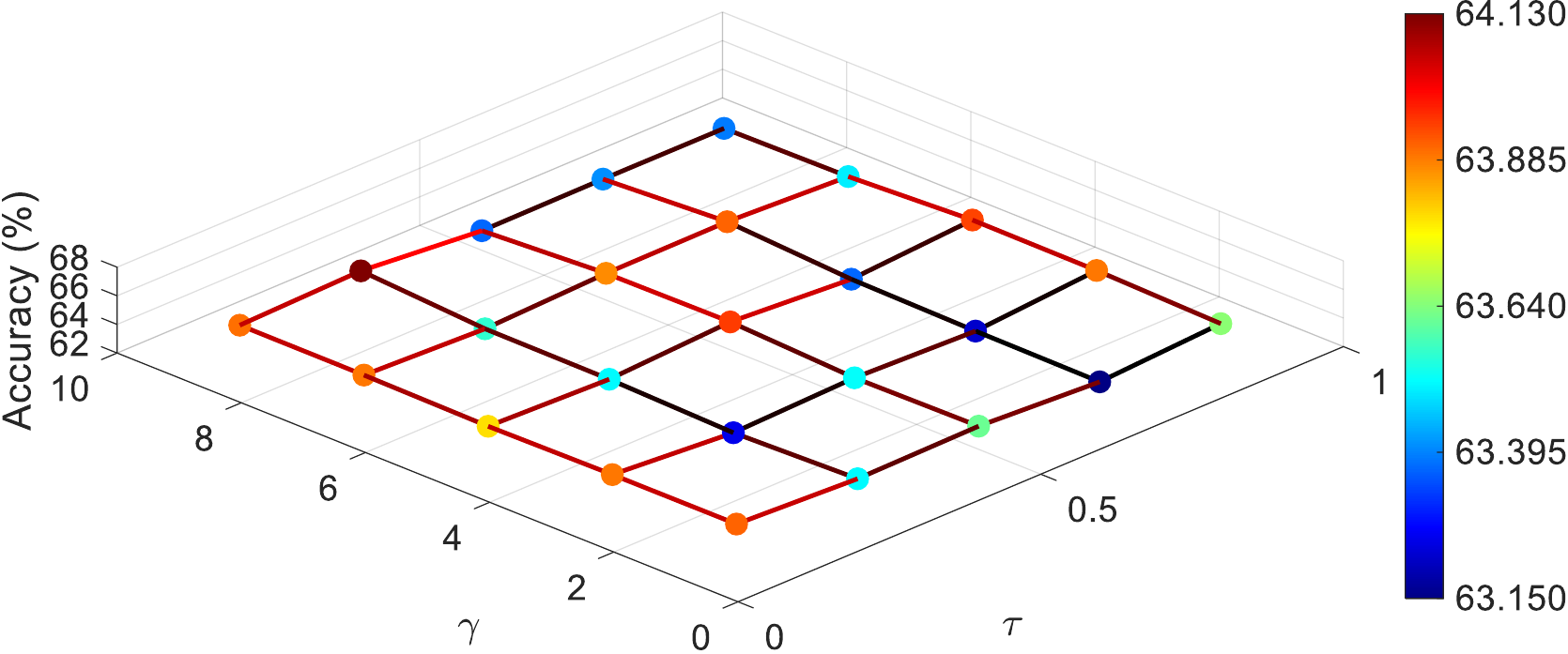}\label{fig8_4}}
  \caption{The sensitivity analyses of hyperparameters $\gamma$ and $\tau$.}
  \label{fig8}
\end{figure}

Here we perform a detailed sensitivity analysis on the hyperparameters used in our proposed FedGSP. There are four hyperparameters (\textit{i.e.}, number of orders $K$, number of selected columns $t$, regularization parameter $\gamma$, and $\tau$) in our FedGSP. First, we plot the accuracy curves along with the variance bar under different values of $K$ and $t$, respectively. As shown in Fig.~\ref{fig7}, the variations in performance under different values of $K$ and $t$ are both small. Second, as shown in Fig.~\ref{fig8}, the variations in performance under different values of $\gamma$ and $\tau$ are both small. These experimental results clearly demonstrate that the performances of FedGSP are very stable within a given range of hyperparameters. In other words, the hyperparameters of our proposed FedGSP can be easily tuned for practical use.

\section{Conclusion}
In this paper, we propose a novel \underline{\textbf{Fed}}erated learning method by mining \underline{\textbf{G}}raph \underline{\textbf{S}}pectral \underline{\textbf{P}}roperties (FedGSP), which effectively mines graph spectral properties to learn graphs with varying homophily levels across different clients. On one hand, FedGSP enables clients to share generic spectral properties, and thus all clients can benefit through collaboration. On the other hand, it allows clients to complement non-generic spectral properties to obtain additional information gain. To the best of our knowledge, this is the \textit{first} time in GFL that alleviates the homophily heterogeneity. Therefore, our method achieves promising performances on eleven datasets and outperforms the second-best method by an average margin of 3.28\% on all heterophilic graph datasets.

\bibliography{mycite}

\end{document}


\title{Homophily Heterogeneity Matters in Graph Federated Learning: A Spectrum Sharing and Complementing Perspective (Appendix)}

\author{Wentao Yu,~\IEEEmembership{Student Member,~IEEE}
}

\markboth{Journal of \LaTeX\ Class Files,~Vol.~14, No.~8, August~2024}%
{Shell \MakeLowercase{\textit{et al.}}: A Sample Article Using IEEEtran.cls for IEEE Journals}


\maketitle

\begin{abstract}
In this appendix, we first provide the proofs for several theorems. After that, we present the details of constructing heterophily bases and the optimization process, respectively. Finally, we illustrate the implementation details of our experiments.
\end{abstract}

\section{Proofs}
\label{sec:Proofs}
In this section, we present the detailed proofs of our proposed theorems.

\subsection{Proof of Theorem 1}
\label{theorem_proportional_proof}
\begin{proof}
First, according to~\cite{huanguniversal}, the Laplacian frequency component of $\mathcal{G}_\mathrm{c}$ can be written as
\begin{equation}
\begin{aligned}
f(\bm{\Theta})&=\mathrm{Tr}(\frac{\bm{\Theta}^{\top} \mathbf{L}_\mathrm{c} \bm{\Theta}}{2})\\
&=\frac{\sum_{\langle i, j\rangle \in \mathcal{E}_\mathrm{c}} \Vert\bm{\theta}_i - \bm{\theta}_j \Vert^2_2 }{2\sum_{i \in \mathcal{V}_\mathrm{c}} \bm{\theta}_{i}^2d_i}.
\end{aligned}
\end{equation}
After several communication rounds of federation, it is reasonable to assume that similar client pairs tend to cluster together, while complementary client pairs exhibit a notable separation.

Second, inspired by~\cite{huanguniversal}, we assume the existence of a constant $\delta$ such that, when $S(i,j) \geq 0.5$, the condition $\Vert\bm{\theta}_i - \bm{\theta}_j\Vert_2 \leq t\delta$ holds for all $i, j \in \mathcal{V}_\mathrm{c}$, where $t \ll 1$ is a small constant. Conversely, when $S(i,j) < 0.5$, the condition $\Vert\bm{\theta}_i - \bm{\theta}_j\Vert_2 = r(\bm{\theta}_i, \bm{\theta}_j)\delta$ holds for all $i, j \in \mathcal{V}_\mathrm{c}$, where $r(\bm{\theta}_i, \bm{\theta}_j) \geq 1$ is a function parameterized by $\bm{\theta}_i$ and $\bm{\theta}_j$. Since $r^2(\bm{\theta}_i, \bm{\theta}_j)$ grows much faster than $t^2$, we can have $t^2 = o(r^2(\bm{\theta}_i, \bm{\theta}_j))$, where ``$o(\cdot)$'' denotes the little-o notation. 

Third, we can have
\begin{equation}
\begin{aligned}
f(\bm{\Theta})&=\frac{\sum_{\langle i, j\rangle \in \mathcal{E}_\mathrm{c}} \Vert\bm{\theta}_i - \bm{\theta}_j \Vert^2_2 }{2\sum_{i \in \mathcal{V}_\mathrm{c}} \bm{\theta}_{i}^2d_i}\\
&=\frac{t^2 \delta^2 r_s |\mathcal{E}_\mathrm{c}| + \sum_{\langle i, j\rangle \in \mathcal{E}_\mathrm{c}, S(i,j) < 0.5}r^2(\bm{\theta}_i,\bm{\theta}_j) \delta^2}{2 \sum_{i \in \mathcal{V}_c}\bm{\theta}_{i}^2d_i}\\
&=\frac{t^2 \delta^2 r_s |\mathcal{E}_\mathrm{c}|}{2 \sum_{i \in \mathcal{V}_\mathrm{c}}\bm{\theta}_{i}^2d_i} + \frac{\sum_{\langle i, j\rangle \in \mathcal{E}_\mathrm{c}, S(i,j) < 0.5}r^2(\bm{\theta}_i, \bm{\theta}_j) \delta^2}{2 \sum_{i \in \mathcal{V}_\mathrm{c}}\bm{\theta}_{i}^2d_i}\\
&=o(r^2(\bm{\theta}_i, \bm{\theta}_j)) + \frac{\sum_{\langle i, j\rangle \in \mathcal{E}_\mathrm{c}, S(i,j) < 0.5}r^2(\bm{\theta}_i, \bm{\theta}_j) \delta^2}{2 \sum_{i \in \mathcal{V}_\mathrm{c}}\bm{\theta}_{i}^2d_i}\\
&= o(r^2(\bm{\theta}_i, \bm{\theta}_j)) + \frac{|\{\langle i, j\rangle \in \mathcal{E}_\mathrm{c}: S(i,j) < 0.5\}|}{2|\mathcal{E}_\mathrm{c}|}\\
&= \frac{1}{2} r_\mathrm{c} + o(r^2(\bm{\theta}_i, \bm{\theta}_j))\\
\end{aligned}
\label{theorem_proportional_proof_eq1}
\end{equation}
\end{proof}

\subsection{Proof of Theorem 2}
\label{theorem_equivalent_proof}
\begin{proof}
According to the theorem 4 in the main paper, the Laplacian Frequency Component $f(\bm{\Theta})$ can be written as
\begin{equation}
\begin{aligned}
f(\bm{\Theta})&=\mathrm{Tr}(\frac{\bm{\Theta}^{\top} \mathbf{L}_\mathrm{c} \bm{\Theta}}{2})\\
&=\frac{1}{4} \sum_{i=1}^M \sum_{j=1}^M \mathbf{W}_\mathrm{c}^{ij} \Vert \bm{\theta}_i - \bm{\theta}_j\Vert^2_2\\
&=\frac{1}{4} \sum_{\langle i, j\rangle \in \mathcal{E}_\mathrm{c}} \mathbf{W}_\mathrm{c}^{ij} \Vert \bm{\theta}_i - \bm{\theta}_j\Vert^2_2 .
\end{aligned}
\label{theorem_equivalent_proof_eq1}
\end{equation}
Since the measure of heterogeneity $H(\mathcal{G}_\mathrm{c})=\sum_{\langle i, j\rangle \in \mathcal{E}_\mathrm{c}} \mathbf{W}_\mathrm{c}^{ij} \Vert\bm{\theta}_i - \bm{\theta}_j\Vert^2_2 $, we can have $f(\bm{\Theta}) \propto H(\mathcal{G}_\mathrm{c})$. Therefore, we can conclude that the Laplacian frequency component of the federated collaboration graph $\mathcal{G}_\mathrm{c}$ is equivalent to the measure of heterogeneity in GFL.

\end{proof}

\section{Details of Constructing Heterophily Bases}
Recall that a fixed angle $\theta$ is formed between any pairs of heterophily bases to ensure the desired spectral property. To determine the value of $\theta$, Huang~\textit{et al.}~\cite{huanguniversal} empirically set $\theta=\frac{\pi}{2}(1-\hat{h}_m)$, where $\hat{h}_m$ is is the estimated homophily ratio on $\mathcal{G}_\mathrm{m}$. Therefore, the procedure of constructing heterophily bases is to manipulate the fixed angle $\theta$ between any pair of bases. First, we normalize the node feature matrix $\mathbf{X}_m$ by the Frobenius norm and treat it as the zeroth order of heterophily bases (\textit{i.e.}, $\mathbf{U}_m^0$). Second, we employ the orthonormal bases $[\mathbf{V}_m^0,\mathbf{V}_m^1,\cdots,\mathbf{V}_m^K]$ to assist in constructing the heterophily bases. Third, we use the three-term recurrence theorem~\cite{guo2023graph} to compute the orthonormal bases $[\mathbf{V}_m^0,\mathbf{V}_m^1,\cdots,\mathbf{V}_m^K]$. Finally, we update the heterophily bases by $\mathbf{U}_m^k\leftarrow\frac{\mathbf{U}_m^k+\mathbf{T}_m^k\mathbf{V}_m^k}{\Vert\mathbf{U}_m^k+\mathbf{T}_m^k\mathbf{V}_m^k\Vert_F}$, where $\mathbf{T}_m^k\leftarrow\sqrt{\left(\frac{(\mathbf{S}_m^{k-1})^{\top}\mathbf{U}_m^{k-1}}{k\cos\theta}\right)^2-\frac{(k-1)\cos\theta+1}{k}}$. According to~\cite{huanguniversal}, the procedure of constructing heterophily bases is exhibited in Algorithm~\ref{algorithm1}, where `$\leftarrow$' denotes the assignment operation. Moreover, the property of the heterophily bases is proved in~\cite{huanguniversal}, which confirms that the fixed angle between any pair of bases is $\theta$.

\begin{algorithm}[t]
\caption{Construct the Heterophily Bases}\label{algorithm1}

{\textbf{Input:}} Graph $\mathcal{G}_\mathrm{m}$; propagation matrix $\mathbf{P}_\mathrm{m}$; node feature matrix $\mathbf{X}_m$; the order of heterophily bases $K$; and the estimated homophily ratio $\hat{h}_m$.

{\textbf{Output:}} Heterophily bases $\mathbf{U}_m=[\mathbf{U}_m^0,\mathbf{U}_m^1,\cdots,\mathbf{U}_m^K]$.

\begin{algorithmic}[1]
\STATE $\mathbf{U}_m^0\leftarrow \frac{\mathbf{X}_m}{\Vert \mathbf{X}_m \Vert_F}$, $\mathbf{V}_m^0\leftarrow \mathbf{U}_m^0$, $\mathbf{V}_m^{-1}\leftarrow \mathbf{0}$, $\mathbf{S}_m^0\leftarrow \mathbf{U}_m^0$, $\theta \leftarrow \frac{\pi}{2}(1-\hat{h}_m)$;
\STATE \textbf{for} $k \leftarrow 1$ \textbf{to} $K$ \textbf{do}
\STATE \hspace{0.5cm} $\mathbf{V}_m^k\leftarrow \mathbf{P}_\mathrm{m}\mathbf{V}_m^{k-1}$;
\STATE \hspace{0.5cm} $\mathbf{V}_m^k\leftarrow \mathbf{V}_m^k - ((\mathbf{V}_m^k)^{\top}\mathbf{V}_m^{k-1})\mathbf{V}_m^{k-1} - ((\mathbf{V}_m^k)^{\top}\mathbf{V}_m^{k-2})\mathbf{V}_m^{k-2}$;
\STATE \hspace{0.5cm} $\mathbf{V}_m^k \leftarrow \frac{\mathbf{V}_m^k}{\Vert \mathbf{V}_m^k \Vert_F}$, $\mathbf{U}_m^k \leftarrow \frac{\mathbf{S}_m^{k-1}}{k}$, $\mathbf{S}_m^{k-1} \leftarrow \sum_{i=0}^{k-1}\mathbf{U}_m^i$;
\STATE \hspace{0.5cm} $\mathbf{T}_m^k\leftarrow\sqrt{\left(\frac{(\mathbf{S}_m^{k-1})^{\top}\mathbf{U}_m^{k-1}}{k\cos\theta}\right)^2-\frac{(k-1)\cos\theta+1}{k}}$;
\STATE \hspace{0.5cm} $\mathbf{U}_m^k\leftarrow\frac{\mathbf{U}_m^k+\mathbf{T}_m^k\mathbf{V}_m^k}{\Vert\mathbf{U}_m^k+\mathbf{T}_m^k\mathbf{V}_m^k\Vert_F}$, $\mathbf{S}_m^k\leftarrow \mathbf{S}_m^{k-1} + \mathbf{U}_m^k$.
\end{algorithmic}
\end{algorithm}

\section{Details of Optimization}
In this section, we first present the detailed optimization processes of three variables, respectively. Second, we analyze the computational complexity of the entire optimization process.

\subsection{Update \texorpdfstring{$\mathbf{R}$}{} with \texorpdfstring{$\mathbf{W}_c$}{} and \texorpdfstring{$\mathbf{S}$}{} Fixed}
Recall that we have to optimize the following problem, which can be written as
\begin{equation}
\begin{aligned}
&\operatorname*{min}_{\mathbf{O}} \mathrm{Tr}(\mathbf{O}^{\top} \mathcal{L}_\mathrm{c} \mathbf{O}) \Rightarrow \operatorname*{min}_{\mathbf{R}} \mathrm{Tr}(\mathbf{R}\mathbf{P}^{\top} \mathcal{L}_\mathrm{c} \mathbf{P}\mathbf{R})\\
&\mathrm{s.t.} \quad \mathbf{r}^{\top} \mathbf{1} = 1, r_l \geq 0, l=1,2, \cdots, d\\
& \quad \quad \mathbf{R} = \mathrm{diag}(\mathbf{r}).
\end{aligned}
\end{equation}
First, inspired by~\cite{nie2020self}, let us denote the matrix $\mathbf{E} \in \mathbb{R}^{d \times d} = \mathbf{P}^{\top} \mathcal{L}_\mathrm{c} \mathbf{P}$ and $e_i^*$ as the $i$-th diagonal element of matrix $\mathbf{E}$, we can obtain
\begin{equation}\label{eq_update1_1}
\begin{aligned}
&\operatorname*{min}_{\mathbf{R}}  \mathrm{Tr}(\mathbf{R}\mathbf{E}\mathbf{R}) \Rightarrow \operatorname*{min}_{r_i} \sum_{i=1}^d r_i^2 e_i^*\\
&\mathrm{s.t.} \quad \mathbf{r}^{\top} \mathbf{1} = 1, r_l \geq 0, l=1,2, \cdots, d\\
& \quad \quad \mathbf{R} = \mathrm{diag}(\mathbf{r}).
\end{aligned}
\end{equation}
Second, we employ the Lagrange multiplier method to solve Eq.~\eqref{eq_update1_1}. In the first step of employing the Lagrange multiplier method, we only consider the constraint $\mathbf{r}^{\top} \mathbf{1} = 1$. Therefore, the Lagrangian function of Eq.~\eqref{eq_update1_1} can be written as
\begin{equation}\label{eq_update1_2}
\mathcal{L}(\mathbf{r},z)= \sum_{i=1}^d r_i^2 e_i^* + z (\sum_{i=1}^d r_i - 1).
\end{equation}
In Eq.~\eqref{eq_update1_2}, we take the derivative with regrad to $r_i$ and set the derivative to 0 for each $i$. Consequently, we can have
\begin{equation}\label{eq_update1_3}
\begin{aligned}
& \frac{ \partial \mathcal{L}(\mathbf{r},z) }{ \partial r_i } = 2r_i e_i^* + z = 0 \Rightarrow \\
& 2r_i e_i^* = -z \Rightarrow r_i = - \frac{z}{2e_i^*}.
\end{aligned}
\end{equation}
Third, we combine Eq.~\eqref{eq_update1_3} with the constraint $\mathbf{r}^{\top} \mathbf{1} = 1$. Therefore, we can have
\begin{equation}\label{eq_update1_4}
- \frac{ z }{ 2 }\sum_{i=1}^d \frac{1}{e_i^*} = 1 \Rightarrow z = - \frac{2}{\sum_{i=1}^d \frac{1}{e_i^*}}.
\end{equation}
After that, we take Eq.~\eqref{eq_update1_4} to Eq.~\eqref{eq_update1_3}. Therefore, we can have
\begin{equation}\label{eq_update1_5}
r_i = \frac{1}{e_i^* \sum_{i=1}^d \frac{1}{e_i^*}}.
\end{equation}
Next, we concentrate on the another constraint $r_l \geq 0$. We can find that the $i$-th diagonal element of matrix $\mathbf{E}$ can be calculated by $e_i^*=\mathbf{p}^{\top}_{[:,i]} \mathcal{L}_\mathrm{c} \mathbf{p}_{[:,i]}$, where $\mathbf{p}_{[:,i]} \in \mathbb{R}^{M \times 1}$ is the $i$-th column of $\mathbf{P}$, and $\mathbf{P} = [\mathbf{p}_1, \mathbf{p}_2, \cdots, \mathbf{p}_M]^{\top}$. Let us suppose the $\mathbf{p}_{[:,i]}$ as the matrix $\mathbf{O}$ in the theorem 4 of the main paper, we can have
\begin{equation}\label{eq_update1_6}
\begin{aligned}
&\mathbf{p}^{\top}_{[:,i]} \mathcal{L}_\mathrm{c} \mathbf{p}_{[:,i]}  \Rightarrow \mathrm{Tr}(\mathbf{p}^{\top}_{[:,i]} \mathcal{L}_\mathrm{c} \mathbf{p}_{[:,i]}) \Rightarrow\\
& \frac{1}{2} \sum_{j=1}^M \sum_{u=1}^M \Vert p_{[:,i], j} - p_{[:,i], u} \Vert_2^{2}w_{ju} .
\end{aligned}
\end{equation}
Since $\mathbf{W}_c$ is the fixed adjacency matrix of federated collaboration graph, we can find that $w_{jk} \geq 0$ in Eq.~\eqref{eq_update1_6}. Moreover, since $\frac{1}{2} \sum_{j=1}^M \sum_{u=1}^M \Vert p_{[:,i], j} - p_{[:,i], u} \Vert_2^{2} \geq 0$, we can have that $e_i^*=\mathbf{p}^{\top}_{[:,i]} \mathcal{L}_\mathrm{c} \mathbf{p}_{[:,i]} \geq 0$, which means that the constraint $r_i = \frac{1}{e_i^* \sum_{i=1}^d \frac{1}{e_i^*}} \geq 0$ is satisfied.

\subsection{Update \texorpdfstring{$\mathbf{S}$}{} with \texorpdfstring{$\mathbf{W}_c$}{} and \texorpdfstring{$\mathbf{R}$}{} Fixed}
\label{Appendix_update2}
Since $\mathbf{W}_c$ and $\mathbf{R}$ are fixed, the optimization objective in Eq.~(6) of the main paper can be written as
\begin{equation}\label{eq_update2}
\begin{aligned}
&\operatorname*{min}_{\mathbf{S}} - \sum_{i=1}^{M} \sum_{j=1}^{M}\Vert \mathbf{S} \mathbf{q}_i - \mathbf{S} \mathbf{q}_j \Vert_2^{2}w_{ij}\\
&\mathrm{s.t.} \quad \mathbf{s}^{\top} \mathbf{1} = 1, s_l \geq 0, l=1,2, \cdots, d\\
& \quad \quad \mathbf{S} = \mathrm{diag}(\mathbf{s}).
\end{aligned}
\end{equation}
According to the theorem 4 of the main paper, we set the $\mathbf{o}_i = \mathbf{S} \mathbf{q}_i \in \mathbb{R}^{d \times 1} \Rightarrow \mathbf{O} = \mathbf{Q}\mathbf{S} $. Therefore, Eq.~\eqref{eq_update2} becomes
\begin{equation}
\begin{aligned}
&\operatorname*{min}_{\mathbf{S}} -\mathrm{Tr}(\mathbf{O}^{\top} \mathcal{L}_\mathrm{c} \mathbf{O}) \Rightarrow \operatorname*{min}_{\mathbf{S}} -\mathrm{Tr}(\mathbf{S}\mathbf{Q}^{\top} \mathcal{L}_\mathrm{c} \mathbf{Q}\mathbf{S})\\
&\mathrm{s.t.} \quad \mathbf{s}^{\top} \mathbf{1} = 1, s_l \geq 0, l=1,2, \cdots, d\\
& \quad \quad \mathbf{S} = \mathrm{diag}(\mathbf{s}).
\end{aligned}
\end{equation}
First, inspired by~\cite{nie2020self}, let us denote the matrix $\mathbf{N} \in \mathbb{R}^{d \times d} = \mathbf{Q}^{\top} \mathcal{L}_\mathrm{c} \mathbf{Q}$ and $n_i^*$ as the $i$-th diagonal element of matrix $\mathbf{N}$, we obtain
\begin{equation}\label{eq_update2_1}
\begin{aligned}
&\operatorname*{min}_{\mathbf{S}} - \mathrm{Tr}(\mathbf{S}\mathbf{N}\mathbf{S}) \Rightarrow \operatorname*{min}_{s_i} -\sum_{i=1}^d s_i^2 n_i^*\\
&\mathrm{s.t.} \quad \mathbf{s}^{\top} \mathbf{1} = 1, s_l \geq 0, l=1,2, \cdots, d\\
& \quad \quad \mathbf{S} = \mathrm{diag}(\mathbf{s}).
\end{aligned}
\end{equation}
Second, we employ the Lagrange multiplier method to solve Eq.~\eqref{eq_update2_1}. In the first step of employing the Lagrange multiplier method, we only consider the constraint $\mathbf{s}^{\top} \mathbf{1} = 1$. Therefore, the Lagrangian function of Eq.~\eqref{eq_update2_1} can be written as
\begin{equation}\label{eq_update2_2}
\mathcal{L}(\mathbf{s},z)= -\sum_{i=1}^d s_i^2 n_i^* + z (\sum_{i=1}^d s_i - 1).
\end{equation}
In Eq.~\eqref{eq_update2_2}, we take the derivative with regrad to $s_i$ and set the derivative to 0 for each $i$. Consequently, we can have
\begin{equation}\label{eq_update2_3}
\begin{aligned}
& \frac{ \partial \mathcal{L}(\mathbf{s},z) }{ \partial s_i } = -2s_i n_i^* + z = 0 \Rightarrow \\
& 2s_i n_i^* = z \Rightarrow s_i = \frac{z}{2n_i^*}.
\end{aligned}
\end{equation}
Third, we combine Eq.~\eqref{eq_update2_3} with the constraint $\mathbf{s}^{\top} \mathbf{1} = 1$. Therefore, we can have
\begin{equation}\label{eq_update2_4}
\frac{ z }{ 2 }\sum_{i=1}^d \frac{1}{n_i^*} = 1 \Rightarrow z = \frac{2}{\sum_{i=1}^d \frac{1}{n_i^*}}.
\end{equation}
After that, we take Eq.~\eqref{eq_update2_4} to Eq.~\eqref{eq_update2_3}. Therefore, we can have
\begin{equation}\label{eq_update2_5}
s_i = \frac{1}{n_i^* \sum_{i=1}^d \frac{1}{n_i^*}}.
\end{equation}
Next, we concentrate on the another constraint $s_l \geq 0$. We can find that the $i$-th diagonal element of matrix $\mathbf{N}$ can be calculated by $n_i^*=\mathbf{q}^{\top}_{[:,i]} \mathcal{L}_\mathrm{c} \mathbf{q}_{[:,i]}$, where $\mathbf{q}_{[:,i]} \in \mathbb{R}^{M \times 1}$ is the $i$-th column of $\mathbf{Q}$, and $\mathbf{Q} = [\mathbf{q}_1, \mathbf{q}_2, \cdots, \mathbf{q}_M]^{\top}$. Let us suppose the $\mathbf{q}_{[:,i]}$ as the matrix $\mathbf{O}$ in the theorem 4 of the main paper, we can have
\begin{equation}\label{eq_update2_6}
\begin{aligned}
&\mathbf{q}^{\top}_{[:,i]} \mathcal{L}_\mathrm{c} \mathbf{q}_{[:,i]}  \Rightarrow \mathrm{Tr}(\mathbf{q}^{\top}_{[:,i]} \mathcal{L}_\mathrm{c} \mathbf{q}_{[:,i]}) \Rightarrow\\
& \frac{1}{2} \sum_{j=1}^M \sum_{u=1}^M \Vert q_{[:,i], j} - q_{[:,i], u} \Vert_2^{2}w_{ju} .
\end{aligned}
\end{equation}
Since $\mathbf{W}_c$ is the fixed adjacency matrix of federated collaboration graph, we can find that $w_{ju} \geq 0$ in Eq.~\eqref{eq_update2_6}. Moreover, since $\frac{1}{2} \sum_{j=1}^M \sum_{u=1}^M \Vert q_{[:,i], j} - q_{[:,i], u} \Vert_2^{2} \geq 0$, we can have that $n_i^*=\mathbf{q}^{\top}_{[:,i]} \mathcal{L}_\mathrm{c} \mathbf{q}_{[:,i]} \geq 0$, which means that the constraint $s_i = \frac{1}{n_i^* \sum_{i=1}^d \frac{1}{n_i^*}} \geq 0$ is satisfied.

\subsection{Update \texorpdfstring{$\mathbf{W}_c$}{} with \texorpdfstring{$\mathbf{S}$}{} and \texorpdfstring{$\mathbf{R}$}{} Fixed}
Recall that we have to optimize the following problem, which can be written as
\begin{equation}\label{eq_update3_1}
\begin{aligned}
&\operatorname*{min}_{\mathbf{w}_i} \sum_{j=1}^{M} (\Vert \mathbf{R} \mathbf{p}_i - \mathbf{R} \mathbf{p}_j \Vert_2^{2}w_{ij}-\Vert \mathbf{S} \mathbf{q}_i - \mathbf{S} \mathbf{q}_j \Vert_2^{2}w_{ij} \\
&\quad \quad \quad \quad \quad \quad + \gamma w_{ij}^2)\\
&\mathrm{s.t.} \quad \quad \mathbf{w}_i^{\top} \mathbf{1} = 1, \quad w_{ij} \geq 0.
\end{aligned}
\end{equation}
First, inspired by~\cite{nie2020self}, we introduce the vector $\mathbf{t}_i$ with the $j$-th element as $t_{ij} = \Vert \mathbf{R} \mathbf{p}_i - \mathbf{R} \mathbf{p}_j \Vert_2^{2}-\Vert \mathbf{S} \mathbf{q}_i - \mathbf{S} \mathbf{q}_j \Vert_2^{2} $. Therefore, we can have
\begin{equation}\label{eq_update3_2}
\begin{aligned}
&\operatorname*{min}_{\mathbf{w}_i} \sum_{j=1}^{M} (t_{ij} w_{ij} + \gamma w_{ij}^2)\\
& \Rightarrow \operatorname*{min}_{\mathbf{w}_i} \sum_{j=1}^{M} ( w_{ij}^2 + \frac{1}{\gamma} t_{ij} w_{ij} +  \frac{1}{4 \gamma^2}t_{ij}^2)\\
& \Rightarrow \operatorname*{min}_{\mathbf{w}_i} \Vert \mathbf{w}_i + \frac{1}{2\gamma}\mathbf{t}_i \Vert_2^2\\
&\mathrm{s.t.} \quad \quad \mathbf{w}_i^{\top} \mathbf{1} = 1, \quad w_{ij} \geq 0.
\end{aligned}
\end{equation}
Second, we employ the Lagrange multiplier method to solve Eq.~\eqref{eq_update3_2}. Therefore, the Lagrangian function of Eq.~\eqref{eq_update3_2} can be written as
\begin{equation}\label{eq_update3_3}
\mathcal{L}(\mathbf{w}_i, \tau, \mathbf{b}_i)= k_l \Vert \mathbf{w}_i + \frac{1}{2\gamma}\mathbf{t}_i \Vert^2_2- \tau(\mathbf{w}_i^{\top} \mathbf{1} - 1) - \mathbf{b}_i^{\top}\mathbf{w}_i,
\end{equation}
where $\tau$ and $\mathbf{b}_i \geq 0$ are the Lagrange multipliers. In Eq.~\eqref{eq_update3_3}, $k_l$ is set to 0.5 according to~\cite{nie2020self}. Third, according to the Karush-Kuhn-Tucker (KKT) condition, we can have
\begin{equation}\label{eq_update3_4}
\begin{cases}
\forall j, \quad w_{ij}^* + \frac{t_{ij}}{2\gamma} - \tau^* - b^*_{ij} = 0 \\
\forall j, \quad w_{ij}^* b^*_{ij} = 0\\
\forall j, \quad b_{ij}^* \geq 0\\
\forall j, \quad w_{ij}^* \geq 0,& 
\end{cases}
\end{equation}
where $w_{ij}^*$ is the $j$-th element of $\mathbf{w}_{i}^*$. According to Eq.~\eqref{eq_update3_4}, we can have
\begin{equation}\label{eq_update3_5}
w_{ij}^* = -\frac{t_{ij}}{2\gamma} + \tau^* + b^*_{ij}.
\end{equation}
Third, we combine Eq.~\eqref{eq_update3_5} with the constraint $\mathbf{w}_i^{\top} \mathbf{1} = 1$. Therefore, we can have
\begin{equation}\label{eq_update3_6}
-\frac{1}{2\gamma} \mathbf{1}^{\top} \mathbf{t}_i + M \tau^* + \mathbf{1}^{\top} \mathbf{b}^*_i = 1
\end{equation}
and then
\begin{equation}\label{eq_update3_7}
\tau^* = \frac{1}{2M\gamma} \mathbf{1}^{\top} \mathbf{t}_i - \frac{1}{M}\mathbf{1}^{\top} \mathbf{b}^*_i +  \frac{1}{M}.
\end{equation}
Combining Eq.~\eqref{eq_update3_7} with the first term of Eq.~\eqref{eq_update3_4}, we can have
\begin{equation}\label{eq_update3_8}
w_{ij}^* = \frac{1}{M} + \frac{1}{2M\gamma} \mathbf{1}^{\top} \mathbf{t}_i - \frac{\mathbf{1}^{\top} \mathbf{b}^*_i}{M} - \frac{t_{ij}}{2 \gamma} + b_{ij}^*
\end{equation}
and then
\begin{equation}\label{eq_update3_9}
\mathbf{w}_{i}^* = \frac{1}{M}(1 - \mathbf{1}^{\top}\mathbf{b}_i^*)\mathbf{1} - \frac{1}{2\gamma}(\mathbf{t}_i - \frac{1}{M}\mathbf{1}^{\top}\mathbf{t}_i\mathbf{1}) + \mathbf{b}_i^*,
\end{equation}
where $\mathbf{1}^{\top} \mathbf{t}_i$ is a constant. After that, we denote $\hat{b}_i^* = \frac{\mathbf{1}^{\top}\mathbf{b}^*_i}{M}$ and $h_{ij}=\frac{1}{M} - \frac{t_{ij}}{2\gamma} + \frac{\mathbf{1}^{\top}\mathbf{t}_i}{2M\gamma}$. Therefore, according to Eq.~\eqref{eq_update3_8}, $w_{ij}^*$ can be written as
\begin{equation}\label{eq_update3_10}
w_{ij}^* = h_{ij} + b_{ij}^* - \hat{b}_i^*.
\end{equation}
Meanwhile, we can have
\begin{equation}\label{eq_update3_11}
b_{ij}^* = w_{ij}^* - h_{ij} + \hat{b}_i^*.
\end{equation}
According to the second, thrid, and fourth terms of Eq.~\eqref{eq_update3_4}, we can infer that $w_{ij}^* = 0$ when $b_{ij}^* \geq 0$, and $b_{ij}^* = 0$ when $w_{ij}^* \geq 0$. Consequently, we can have
\begin{equation}\label{eq_update3_12}
w_{ij}^* = (h_{ij} - \hat{b}_i^*)_+
\end{equation}
and
\begin{equation}\label{eq_update3_13}
b_{ij}^* = (\hat{b}_i^* - h_{ij})_+.
\end{equation}
Based on Eq.~\ref{eq_update3_13}, we can have
\begin{equation}\label{eq_update3_14}
\hat{b}_i^* = \frac{1}{M}\sum_{j=1}^{M}(\hat{b}_i^* - h_{ij})_+.
\end{equation}
Then, we define a loss function and employ the Newton method to find the optimal $\hat{b}_i^*$. The loss function can be defined as
\begin{equation}\label{eq_update3_15}
\mathcal{L}(\hat{b}_i) = \frac{1}{M}\sum_{j=1}^{M}(\hat{b}_i - h_{ij})_+ - \hat{b}_i.
\end{equation}
Here we optimize $\hat{b}_i$ in an iterative manner. In the $t+1$-th iteration, the updated rule is
\begin{equation}\label{eq_update3_16}
\hat{b}_i^{t+1} = \hat{b}_i^{t} - \mathcal{L}(\hat{b}_i^t) \cdot [\frac{ \partial \mathcal{L}(\hat{b}_i^t) }{ \partial \hat{b}_i^{t} }]^{-1}.
\end{equation}
If the loss function $\mathcal{L}(\hat{b}_i) \to 0$, we will obtain the optimal $\hat{b}_i^*$ and then the optimal $\mathbf{w}_{i}^*$. 

\subsection{Temporal Complexity Analysis of the Optimization Processes}
\label{complexity_algorithm}
Here we concentrate on the temporal complexity of the optimization processes. Since the overall optimization problem is made up of three parts, the main temporal complexity comes from the following calculations. In Eq.~\eqref{eq_update1_1}, the temporal complexity of calculating $\mathbf{E} = \mathbf{P}^{\top} \mathcal{L}_\mathrm{c} \mathbf{P}$ is $\mathcal{O}(Md^2+M^2d)$. Similarly, in Eq.~\eqref{eq_update2_1}, the temporal complexity of calculating $\mathbf{N} = \mathbf{Q}^{\top} \mathcal{L}_\mathrm{c} \mathbf{Q}$ is $\mathcal{O}(Md^2+M^2d)$. In addition, in Eq.~\eqref{eq_update3_9}, the temporal complexity of calculating $\mathbf{w}_{i}^*$ is $\mathcal{O}(M \times d)$. In conclusion, the temporal complexity of the optimization processes is $\mathcal{O}\big(M \times d \times (M + d + 1)\big)$. Consequently, we believe that this optimization method is efficient in real-world applications, which is also validated by our experiments.

\section{Efficiency Analysis of Our Proposed FedGSP}
Here we analyze the spatial and temporal complexities of our proposed FedGSP on the client and server sides, respectively.

\subsection{Spatial Complexity}
\subsubsection{Client Side}
Since the local model deployed on each client is actually a spectral GNN (\textit{i.e.}, UniFilter), it means that the spatial complexity of our FedGSP on each client is determined by the UniFilter. In the UniFilter, there are $K$ homophily bases, $K$ heterophily bases, and one MLP layer. Therefore, the spatial complexity of UniFilter consists of two contents: homophily and heterophily bases and the MLP layer. First, the spatial complexity of bases is $\mathcal{O}(K \times d)$. Second, the spatial complexity of the MLP layer is $\mathcal{O}(d^2)$. Consequently, the total spatial complexity of our method on the client side is $\mathcal{O}( K \times d  + d^2)$.

\subsubsection{Server Side}
First, the spatial complexity of storing uploaded parameters of $M$ clients is related with the number of parameters of local models. Suppose that the number of parameters of local models is $K \times d^2$. Therefore, the spatial complexity is $\mathcal{O}(M \times K \times d^2)$. Second, the spatial complexity of optimizing collaboration graphs is $\mathcal{O}(K \times M^2)$ when considering there are $K$ orders of bases. Consequently, the total spatial complexity of our method on the server side is $\mathcal{O}( M \times K \times d^2  + K \times M^2) \rightarrow \mathcal{O}\big( M \times K \times (d^2 + M)\big)$.

\subsection{Temporal Complexity}
\subsubsection{Client Side}
Similarly, the temporal complexity of our method on each client is determined by the UniFilter, which is made up of $K$ bases and one MLP layer. First, the temporal complexity of $K$ homophily bases is $\mathcal{O}(K \times e_m \times d + K \times n_m \times d^2)$. Second, according to~\cite{huanguniversal}, the temporal complexity of the construction of heterophily bases is $\mathcal{O}(K \times (n_m + e_m))$. Notably, since the construction of heterophily bases is performed only once at the beginning, this part of temporal complexity can be ignored. Third, the temporal complexity of the MLP layer is $\mathcal{O}(n_m \times d^2)$. Consequently, the total temporal complexity of our method on the client side is $\mathcal{O}(K \times e_m \times d + K \times n_m \times d^2 + n_m \times d^2)$. After simplification, it can be written as $\mathcal{O}\big(K \times (e_m \times d + n_m \times d^2)\big)$.
\subsubsection{Server Side}
On the server, there are two operations, namely, optimization of collaboration strengths and federated aggregation for polynomial bases. First, according to the Section~\ref{complexity_algorithm}, the temporal complexity of optimizing collaboration strengths is $\mathcal{O}\big(M \times d \times (M + d + 1)\big)$. Second, the temporal complexity of separate aggregation is $\mathcal{O}(M \times K)$. Consequently, the total temporal complexity of our method on the server side is $\mathcal{O}\big(M \times d \times (M + d + 1) + M \times K\big)$.
        
\section{Implementation Details}
\label{implementation_details}
In this section, we provide the implementation details of our experiments, including the experimental platform, the dataset descriptions, the details of subgraph partitioning, the information of baseline methods, and our training details.

\subsection{Experimental Platform}
\label{experimental_platform}
All the experiments in our work are conducted on a Linux server with a 2.90 GHz Intel Xeon Gold 6326 CPU, 64 GB of RAM, and two NVIDIA GeForce RTX 4090 GPUs with 48GB of memory. Our proposed FedGSP is implemented via Python 3.8.8, PyTorch 1.12.0, and PyTorch Geometric (PyG) 2.3.0.

\subsection{Datasets}
\label{dataset_info}
To validate the effectiveness of our proposed FedGSP, we perform extensive experiments on eleven widely used benchmark datasets, namely, six homophilic and five heterophilic graph datasets. For the homophilic graph datasets, we choose four citation graphs: \textit{Cora}, \textit{CiteSeer}, \textit{PubMed}, and \textit{ogbn-arxiv}; two Amazon product graphs: \textit{Amazon-Computer} and \textit{Amazon-Photo}. For the heterophilic graph datasets, we select \textit{Roman-empire}, \textit{Amazon-ratings}, \textit{Minesweeper}, \textit{Tolokers}, and \textit{Questions}~\cite{platonov2023a}. The statistical information of the above benchmark datasets is described in Tab.~\ref{datasets_statistics}. Note that we use the Area Under the ROC curve (AUC) as the evaluation metric (higher values are better) for \textit{Minesweeper}, \textit{Tolokers}, and \textit{Questions} datasets, and use the accuracy as the evaluation metric (higher values are better) for other datasets.

\begin{table*}[t]
    \centering
    \caption{Statistical information of eleven used graph datasets.}
    \label{datasets_statistics}
    \begin{tabular}{cccccc}
    \hline
    Types                               & Datasets                      & \# Nodes & \# Edges  & \# Classes & \# Node Features \\ \hline
    \multirow{6}{*}{homophilic graph}   & \textit{Cora}                 & 2,708    & 5,429     & 7          & 1,433            \\
                                        & \textit{CiteSeer}             & 3,327    & 4,732     & 6          & 3,703            \\
                                        & \textit{PubMed}               & 19,717   & 44,324    & 3          & 500              \\
                                        & \textit{Amazon-Computer}      & 13,752   & 491,722   & 10         & 767              \\
                                        & \textit{Amazon-Photo}         & 7,650    & 238,162   & 8          & 745              \\
                                        & \textit{ogbn-arxiv}           & 169,343  & 1,166,243 & 40         & 128              \\ \hline
    \multirow{6}{*}{heterophilic graph} & \textit{Roman-empire}         & 22,662   & 32,927    & 18         & 300              \\
                                        & \textit{Amazon-ratings}       & 24,492   & 93,050    & 5          & 300              \\
                                        & \textit{Minesweeper}          & 10,000   & 39,402    & 2          & 7                \\
                                        & \textit{Tolokers}             & 11,758   & 519,000   & 2          & 10               \\
                                        & \textit{Questions}            & 48,921   & 153,540   & 2          & 301              \\ \hline
    \end{tabular}
    \end{table*}

In order to facilitate the division of datasets, a random sample of 20\% of nodes is selected for training purposes, 40\% for the purpose of validation, and 40\% for testing, with the exception of the \textit{ogbn-arxiv} dataset. Since \textit{ogbn-arxiv} dataset consists of a relatively large number of nodes in comparison to other datasets, a random sample of 5\% of the nodes is used for training, while the remaining half of the nodes are used for validation and testing, respectively.

\subsection{Subgraph Partitioning}
\label{subgraph_partitioning_detail}
Inspired by real-world requirements and following~\cite{baek2023personalized,wentao2025fediih}, we consider two subgraph partitioning settings: non-overlapping and overlapping. In the non-overlapping setting, $\cup_{m=1}^{M} \mathcal{V}_m=\mathcal{V}$ and $\mathcal{V}_m \cap \mathcal{V}_n=\emptyset $ for $\forall m \neq n \in \{1, 2, \cdots, M\}$, where $\mathcal{V}$ represents the node set of the global graph. Partitioning without this property is called overlapping. Here we present the details of how to partition the original graph into multiple subgraphs.

\subsubsection{Non-overlapping Partitioning}
First, if there are $M$ clients, the number of non-overlapping subgraphs to be generated is specified as $M$. Second, the METIS graph partitioning algorithm~\cite{karypis1997metis} is used to divide the original graph into $M$ subgraphs. In other words, the non-overlapping partitioning subgraph for each client is directly obtained by the output of the METIS algorithm.

\subsubsection{Overlapping Partitioning}
First, if there are $M$ clients, the number of overlapping subgraphs to be generated is specified as $M$. Second, the METIS~\cite{karypis1997metis} graph partitioning algorithm is used to divide the original graph into $\lfloor \frac{M}{5} \rfloor$ subgraphs. Third, in each subgraph generated by METIS, half of the nodes and their associated edges are randomly sampled. This procedure is performed five times to generate five different yet overlapped subgraphs. By doing so, the number of overlapping subgraphs is equal to the number of clients.

\subsection{Baseline Methods}
\label{baseline_methods_info}
We compare our proposed FedGSP with eleven baseline methods, which can be categorized into two groups. Concretely, we adopt one classic Federated Learning (FL) method (\textit{i.e.}, FedAvg~\cite{mcmahan2017communication}), two personalized FL methods (\textit{i.e.}, FedProx~\cite{MLSYS2020_1f5fe839} and FedPer~\cite{Arivazhagan2019}), three general GFL methods (\textit{i.e.}, GCFL~\cite{NEURIPS2021_9c6947bd}, FedGNN~\cite{wu2021fedgnn}, and FedSage+\cite{NEURIPS2021_34adeb8e}), and five personalized GFL methods (\textit{i.e.}, FED-PUB~\cite{baek2023personalized}, FedGTA~\cite{li2023fedgta}, AdaFGL~\cite{li2024adafgl}, FedTAD~\cite{zhu2024fedtad}, and FedIIH~\cite{wentao2025fediih}). Moreover, we perform experiments with local training, that is, training each client without federated aggregation. The detailed descriptions of these baseline methods are provided below.

\textbf{FedAvg} This method~\cite{mcmahan2017communication} represents one of the fundamental baseline methods in the field of FL. First, each client independently trains a model, which is subsequently transmitted to a server. Then, the server aggregates the locally updated models by averaging and transmits the aggregated model back to the clients.

\textbf{FedProx} This method~\cite{MLSYS2020_1f5fe839} is one of the personalized FL baseline methods. It customizes a personalized model for each client by adding a proximal term as a subproblem that minimizes weight differences between local and global models.

\textbf{FedPer} This method~\cite{Arivazhagan2019} is one of the personalized FL baseline methods. It only federates the weights of the backbone while training the personalized classification layer in each client.

\textbf{GCFL} This method~\cite{NEURIPS2021_9c6947bd} is one of the basic GFL methods. Specifically, GCFL is designed for vertical GFL (\textit{e.g.}, GFL for molecular graphs). In particular, it uses the bi-partitioning scheme, which divides a set of clients into two disjoint groups of clients based on the similarity of their gradients. This is similar to the mechanism proposed for image classification in clustered-FL~\cite{9174890}. Then, after partitioning, the model weights are shared only among clustered clients with similar gradients.

\textbf{FedGNN} This method~\cite{wu2021fedgnn} is one of the GFL baseline methods. It extends local subgraphs by exchanging node embeddings from other clients. Specifically, if two nodes in two different clients have exactly the same neighbors, FedGNN transfers the nodes with the same neighbors from other clients and expands them.

\textbf{FedSage+} This method~\cite{NEURIPS2021_34adeb8e} is one of the GFL baseline methods. It generates the missing edges between subgraphs and the corresponding neighbor nodes by using the missing neighbor generator. To train this neighbor generator, each client first receives node representations from other clients, and then computes the gradient of the distances between the local node features and the node representations of the other clients. After that, the gradient is sent back to the other clients, and this gradient is then used to train the neighbor generator.

\textbf{FED-PUB} This method~\cite{baek2023personalized} is one of the personalized GFL baseline methods. It estimates the similarities between the subgraphs based on the outputs of the local models that are given the same test graph. Then, based on the similarities, it performs a weighted averaging of the local models for each client. In addition, it learns a personalized sparse mask at each client in order to select and update only the subgraph-relevant subset of the aggregated parameters.

\textbf{FedGTA} This method~\cite{li2023fedgta} is one of the personalized GFL baseline methods. In this method, each client first computes topology-aware local smoothing confidence and mixed moments of neighbor features. They are then used to compute the inter-subgraph similarities, which are uploaded to the server along with the model parameters. Finally, the server is able to perform a weighted federation for each client.

\textbf{AdaFGL} This method~\cite{li2024adafgl} is one of the personalized GFL baseline methods. Actually, it is a decoupled two-step personalized approach. First, it uses standard multi-client federated collaborative training to acquire the federated knowledge extractor by aggregating uploaded models in the final round at the server. Second, each client performs personalized training based on the local subgraph and the federated knowledge extractor.

\textbf{FedTAD} This method~\cite{zhu2024fedtad} is one of the personalized GFL baseline methods. FedTAD designs a generator to generate a pseudo graph, which is subsequently employed for data-free knowledge distillation. By doing so, FedTAD enhances the reliable knowledge transfer from the local models to the global model, alleviating the negative impact of unreliable knowledge caused by node feature heterogeneity.

\textbf{FedIIH} This method~\cite{wentao2025fediih} is one of the personalized GFL baseline methods. FedIIH integrally models the inter- and intra- heterogeneity in GFL. On one hand, it characterizes the inter-heterogeneity from a multi-level global perspective, and thus it can properly compute the inter-subgraph similarities based on the whole distribution. On the other hand, it disentangles the subgraph into several latent factors, so that it can further consider the critical intra-heterogeneity.

\textbf{Local} This method is the non-FL baseline, where the model is trained only locally for each client, without weight sharing.

\subsection{Training Details}
\label{training_details}
\subsubsection{Training Rounds and Epochs}
For the \textit{Cora}, \textit{CiteSeer}, \textit{PubMed}, \textit{Roman-empire}, \textit{Amazon-ratings}, \textit{Minesweeper}, \textit{Tolokers}, and \textit{Questions} datasets, we set the number of local training epochs and total rounds to 1 and 100, respectively. For several large-scale datasets, namely, \textit{Amazon-Computer}, \textit{Amazon-Photo}, and \textit{ogbn-arxiv}, we set the number of total rounds to 200. Note that the number of local epochs is set to 2 for the \textit{Amazon-Photo} and \textit{ogbn-arxiv} datasets, and to 3 for the \textit{Amazon-Computer} dataset. Finally, we report the test performance of all models at the best validation epoch, and the performance is measured by averaging across all clients in terms of node classification accuracies (or AUCs).

\subsubsection{Network Architectures}
For the experiments of all baseline methods, except FedSage+, FedGTA, FedTAD, FedIIH, and our proposed FedGSP, we use two layers of the Graph Convolutional Network (GCN)~\cite{kipf2017semisupervised} and a linear classifier layer as their network architectures. For the hyperparameter settings of baseline methods, we use the default settings given in their original papers. Because of the inductive and scalability advantages of GraphSAGE~\cite{NIPS2017_5dd9db5e}, FedSage+ uses GraphSAGE as the encoder and then trains a missing neighbor generator to handle missing links across local subgraphs. For FedIIH, we use the node feature projection layer of DisenGCN~\cite{pmlrv97ma19a} to obtain the node representations and a linear classifier layer (\textit{i.e.}, MLP) to perform node classifications. In contrast, FedGTA uses a Graph Attention Multi-Layer Perceptron (GAMLP)~\cite{35346783539121} as its backbone and a linear classifier layer to classify nodes. Note that GAMLP~\cite{35346783539121} is one of the scalable Graph Neural Network (GNN) models, which can capture the underlying correlations between different scales of graph knowledge. For our proposed FedGSP, we use the UniFilter to obtain the node representations and a linear classifier layer (\textit{i.e.}, MLP) to perform node classifications.

\section{Homophily}
\textcolor{blue}{Homophily means that edges in the graph tend to connect nodes of the same class~\cite{chen2024polygcl}. For example, in real-world social networks, people tend to connect to users with similar hobbies. Similarly, in citation networks, papers tend to cite papers in the same field of research. In contrast, in dating networks, edges often connect people of opposite genders. This means that different graphs have different levels of homophily, with some graphs having high homophily and others having low homophily.}

\subsection{Conventional Homophily Measures}
\textcolor{blue}{To evaluate the homophily levels of graphs, lots of homophily measures are proposed. For example, edge homophily~\cite{zhu2020beyond} and node homophily~\cite{peigeom} are two conventional and widely used measures. First, the edge homophily of the graph $\mathcal{G}_m$ can be defined as
\begin{equation}\label{eq_homo_define}
h_\mathrm{edge} = \frac{|\{ \langle u, v \rangle \in \mathcal{E}_m: \mathbf{y}_m^u = \mathbf{y}_m^v \}|}{|\mathcal{E}_m|},
\end{equation}
where $\mathcal{E}_m$ is the edge set, $\mathbf{y}_m^u$ and $\mathbf{y}_m^v$ are labels of nodes, and $h_\mathrm{edge} \in [0, 1]$. If $h_\mathrm{edge}$ is close to 1, the homophily of the graph $\mathcal{G}_m$ is high. Otherwise, the homophily of the graph $\mathcal{G}_m$ is low. Similarly, the node homophily can be defined as
\begin{equation}\label{eq_node_homo_define}
h_\mathrm{node} = \frac{1}{n_m}\sum_{v \in \mathcal{V}_m}\frac{|\{ u \in N(v): \mathbf{y}_m^u = \mathbf{y}_m^v \}|}{d(v)},
\end{equation}
where $\mathcal{V}_m$ is the node set, $n_m$ is the number of nodes, $N(v)$ denotes the neighbors of $v$, and $d(v) = | N(v) |$ represents the degree of $v$. Although these two measures are simple and easy to use, they are sensitive to the number of classes and their balance~\cite{NEURIPS2023_01b68102, platonov2023a}.}

\subsection{Adjusted Homophily}
\textcolor{blue}{To overcome the problems of conventional measures, Platonov \textit{et al.} propose the \textit{adjusted homophily}~\cite{NEURIPS2023_01b68102}, which is based on the edge homophily. Specifically, the adjusted homophily can be defined as
\begin{equation}\label{eq_adj_homo_define}
h_\mathrm{adj}=\frac{h_\mathrm{edge}-\sum_{k=1}^{c_m}\bar{p}(k)^2}{1-\sum_{k=1}^{c_m}\bar{p}(k)^2},
\end{equation}
where $\bar{p}(k)=\frac{D_k}{2|\mathcal{E}_m|}$, $D_k=\sum_{v:\mathbf{y}_m^v=k}d(v)$, and $c_m$ denotes the number of classes. In Eq.~\eqref{eq_adj_homo_define}, $\bar{p}(k)$ computes the degree-weighted distribution of class labels. Since adjusted homophily is comparable across different datasets with varying numbers of classes and class size balance, we employ it as a measure of homophily in this paper. If $h_\mathrm{adj}$ is less than or equal to 0, the homophily of the graph $\mathcal{G}_m$ is low. Otherwise, the homophily of the graph $\mathcal{G}_m$ is high.}

\bibliographystyle{IEEEtran}
\bibliography{mycite}